\pdfoutput=1
\documentclass[10pt]{article}

\usepackage[margin=1in]{geometry}
\usepackage{amsmath,amssymb,amsfonts,amsthm}
\usepackage{graphicx}
\makeatletter
\expandafter\def\csname Gin@rule@.svg\endcsname#1{{pdf}{.pdf}{\Gin@base.pdf}}
\makeatother
\usepackage{booktabs}
\usepackage{float}
\usepackage{array}
\usepackage{tikz}
\usetikzlibrary{arrows.meta,positioning,shapes.geometric,fit}
\usepackage{tikz-cd}
\usepackage{multicol}
\usepackage[T1]{fontenc}
\usepackage[utf8]{inputenc}
\usepackage{newtxtext}
\usepackage{newtxmath}
\usepackage{microtype}
\usepackage{needspace} % reserve vertical space before anchored figures
\usepackage{xcolor}
\usepackage{listings}
\definecolor{lstcomment}{HTML}{6A737D}
\definecolor{lststring}{HTML}{0B6E4F}
\definecolor{lstkeyword}{HTML}{12508C}
\lstdefinestyle{pseudo}{
  language=Python,
  basicstyle=\ttfamily\scriptsize,
  keywordstyle=\color{lstkeyword}\bfseries,
  commentstyle=\color{lstcomment}\itshape,
  stringstyle=\color{lststring},
  showstringspaces=false,
  columns=fixed,
  basewidth=0.5em,
  keepspaces=true,
  frame=single,
  framerule=0.4pt,
  rulecolor=\color{black!30},
  xleftmargin=0.5em,
  xrightmargin=0.5em,
  morekeywords={lambda}
}
\usepackage{authblk}
\usepackage[numbers]{natbib}
\usepackage{hyperref}
\usepackage[nameinlink,capitalise,noabbrev]{cleveref}
\usepackage{tikz}
\usetikzlibrary{arrows.meta}

\newcolumntype{P}[1]{>{\raggedright\arraybackslash}p{#1}}

\newcommand{\degree}{\ensuremath{^\circ}}

\definecolor{dorsalfill}{HTML}{12508C}
\definecolor{dorsalsub}{HTML}{B5D4F4}
\definecolor{ventralfill}{HTML}{6E2A12}
\definecolor{ventralsub}{HTML}{F0997B}
\definecolor{v1fill}{HTML}{444441}
\definecolor{v1sub}{HTML}{C9C7BE}
\definecolor{flowgray}{HTML}{888780}
\definecolor{brainfill}{HTML}{3C3489}
\definecolor{brainsub}{HTML}{CECBF6}
\definecolor{robofill}{HTML}{0F6E56}
\definecolor{robosub}{HTML}{9FE1CB}
\definecolor{flowgray}{HTML}{888780}

\definecolor{brainfill}{HTML}{3C3489}
\definecolor{brainsub}{HTML}{CECBF6}
\definecolor{cerebfill}{HTML}{0F6E56}
\definecolor{cerebsub}{HTML}{9FE1CB}
\definecolor{grayfill}{HTML}{444441}
\definecolor{graysub}{HTML}{C9C7BE}
\definecolor{amberfill}{HTML}{854F0B}
\definecolor{ambersub}{HTML}{FAC775}
\definecolor{flowgray}{HTML}{888780}
\definecolor{tealstroke}{HTML}{0F6E56}
\definecolor{amberstroke}{HTML}{BA7517}
\hypersetup{
  colorlinks=true,
  linkcolor=blue!60!black,
  citecolor=blue!60!black,
  urlcolor=blue!60!black,
  pdftitle={Neuro Inspired Robotics: What Robot Learning Can Borrow from the Sensorimotor Brain},
  pdfauthor={Ruihan Zhang, Davy Deng}
}

\renewcommand{\Affilfont}{\small}
\makeatletter
\renewcommand\AB@affilsepx{\qquad\protect\Affilfont}
\makeatother

\theoremstyle{definition}

\theoremstyle{remark}

\title{Neuro Inspired Robotics:\\[0.2em]\large What Robot Learning Can Borrow from the Sensorimotor Brain}

\author[ ]{Ruihan Zhang\thanks{\texttt{ruihanzhang2015@gmail.com}}}
\author[1]{Davy Deng\thanks{\texttt{davy@mit.edu}}}

\affil[1]{MIT}

\date{}

\begin{document}
\maketitle

\begin{abstract}
Robot policies still fail at everyday manipulation: an unfamiliar task, an object never held before, a scene that changes mid-motion, a skill that should improve with use. They fail despite pretraining on over a million trajectories, pooled from dozens of robots at great expense, while a child acquires all of these abilities within a few years of learning from a single body. This survey argues that the gap is one of design, and proposes designs for the next generation of robot models, drawn from how the nervous system solves the same problems. From the body we take compliance: the policy commands stiffness alongside position, so the controller yields where the task allows and presses where it must, which greatly lowers the demand for precise position output. From the visual system, where a fast pathway for action runs in parallel with the slow pathway for recognition, we take a dedicated fast vision system for motor behavior: it precomputes the quantities the policy needs and drives a reactive policy at camera rate. From the cerebellum we take a predictor beside the policy that forecasts the consequence of each command and recalibrates from its own error. From motor cortex we take a single model that composes action units from primitives, verifies each at a sensory boundary, practices the weakest unit with a hypothesis in mind, and reasons each attempt closer to the goal. Proven primitives are then consolidated from code into weights, so they run fluently and respond quickly the next time. Together these describe a robot that acts on its own initiative, the basis of deployment and recursive self-improvement, and we offer each proposal as a hypothesis to be tested.
\end{abstract}


\tableofcontents

\clearpage
\section{Introduction}
\label{sec:introduction}

A seven-year-old opens drawers, catches a flying toy, and plays the piano, and acquired every one of these skills from a single body over a single childhood. In contrast, robot policies trained on over a million trajectories pooled across dozens of robots still fail at tasks this simple (\cref{fig:intro_human_robot_behavior}) \citep{oneill2024openx,black2024pi0,kim2024openvla}. Scale alone cannot explain the gap, because the child never saw more than seven years of data. The difference is in how the nervous system is organized, which makes the child more dexterous, more intelligent, and able to keep improving. This survey asks how that sensorimotor skill develops and which parts of the design robotics can borrow. It is built around three questions:

\begin{figure}[htbp]
\centering
\begin{tikzpicture}
\node[anchor=south west,inner sep=0] (art) at (0,0) {\includegraphics[width=0.84\textwidth]{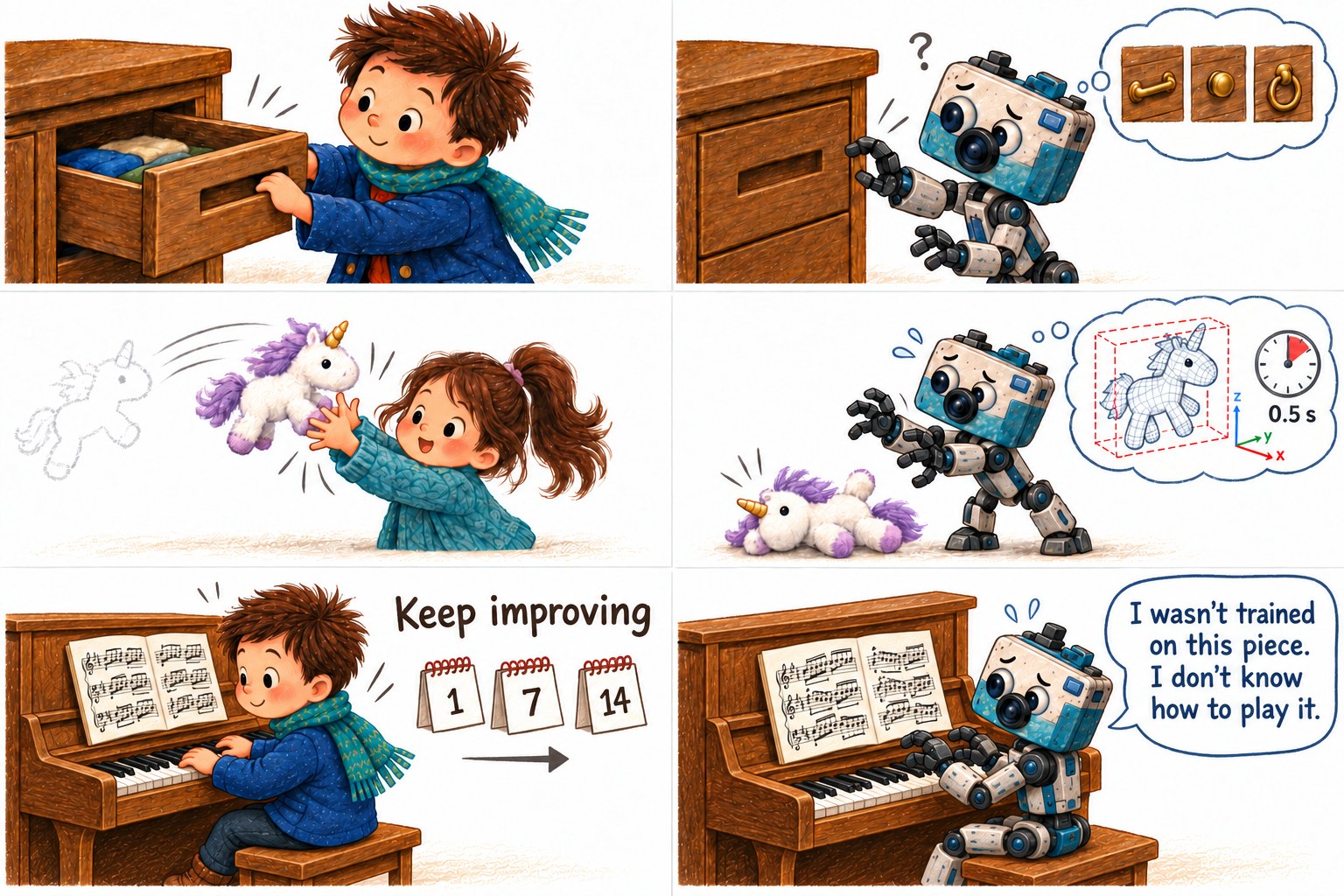}};
\begin{scope}[x={(art.south east)},y={(art.north west)}]
\foreach \panelx/\panely/\paneltext in {0.01/0.99/1a,0.51/0.99/1b,0.01/0.665/2a,0.51/0.665/2b,0.01/0.36/3a,0.51/0.36/3b} {
  \node[anchor=north west,font=\small\bfseries,fill=white,inner sep=2pt] at (\panelx,\panely) {\paneltext};
}
\node[anchor=south,align=center,font=\small\bfseries] at (0.25,1.02) {Human behavior};
\node[anchor=south,align=center,font=\small\bfseries] at (0.75,1.02) {Common limitations of\\current robot policies};
\node[anchor=east,align=right,font=\footnotesize\bfseries] at (-0.025,0.84) {Unfamiliar\\task};
\node[anchor=east,align=right,font=\footnotesize\bfseries] at (-0.025,0.51) {Sudden\\change};
\node[anchor=east,align=right,font=\footnotesize\bfseries] at (-0.025,0.18) {Keep\\improving};
\end{scope}
\end{tikzpicture}
\caption{\textbf{A child has three abilities that current robot policies lack.} Top: a boy opens an unfamiliar drawer on the first try, while a policy finds no training handle to copy. Middle: a girl catches a flying toy, while a robot is still reasoning about shape and coordinates when the toy lands. Bottom: the boy's piano playing improves with days of practice, while a robot never trained on the piano cannot learn it or improve over time, because its learning stopped the day it was deployed.}
\label{fig:intro_human_robot_behavior}
\end{figure}

\paragraph{How does a human solve a task it has never seen?} Never from scratch. A policy trained end to end has nothing to copy when the scene falls outside its training data (\cref{fig:intro_human_robot_behavior}). The brain instead uses reasoning to split the task into a sequence of action units and composes each from a small repertoire of primitives such as reach, orient, and grasp, each a verifiable goal and an action plan, so an unfamiliar drawer becomes reach, grasp, and pull on the first try. Each unit ends at a sensory event, the hand closing on the handle, and the brain checks for it before releasing the next. When the check fails, it reasons to the cause and revises the plan (\cref{sec:planner_commander}).

\paragraph{Why does a human act so fast?} Robotics debates how to pair a large reasoning model, powerful but slow, with a small action policy, fast but shallow, so the system is quick, capable, and still fits on the GPU a robot can carry. The brain escapes the trade-off by keeping deliberation off the path from sensing to action. A second visual stream, largely separate from the one that recognizes objects, guides the next movement (\cref{sec:vision}). Its commands can be coarse because the body is compliant: the controller only decides where motion must be exact and where external forces can carry it (\cref{sec:action}). Touch corrects a slipping grip by reflex within a hundred milliseconds, before vision registers the slip (\cref{sec:touch}). And on the leading account of the cerebellum, a predictor forecasts each command's consequence before the sensors report it, so the response is ready when the world catches up (\cref{sec:cerebellum_world_model}).

\paragraph{How does a human keep self-improving?} A robot policy is trained once before deployment, and naive fine-tuning risks erasing what it knew \citep{mccloskey1989catastrophic,kirkpatrick2017ewc}. The brain keeps learning after deployment, on two timescales. Within minutes the predictor recalibrates to a new tool or a heavier door (\cref{sec:cerebellum_world_model}). Over weeks, practice concentrates on the step that still fails, dopamine scores each attempt and keeps what worked, and consolidation during rest and sleep turns the result into an action primitive that runs with little deliberation (\cref{sec:planner_commander}). Once consolidated, a skill resists interference from the next, so learning adds without erasing.

The survey follows the flow of information through a motor system. It begins with the modalities that carry it, action as output (\cref{sec:action}) and touch and vision as inputs (\cref{sec:touch,sec:vision}), and where brain and robot differ on each. Two models connect them. One predicts the consequence of a command: the cerebellum in the brain, the world model in robotics (\cref{sec:cerebellum_world_model}). The other decides what to do: the motor hierarchy from prefrontal to motor cortex in the brain, vision-language models (VLMs) and vision-language-action policies (VLAs) in robotics (\cref{sec:planner_commander}). Each chapter reviews where robotics stands, how the brain solves the same problem, and what to build. Evolution has run this experiment for hundreds of millions of years, and we hope its principles point toward robots that are intelligent, swift, and able to improve themselves.

\clearpage
\section{Action}
\label{sec:action}

\begin{figure}[H]
  \centering
  \includegraphics[width=0.5\linewidth]{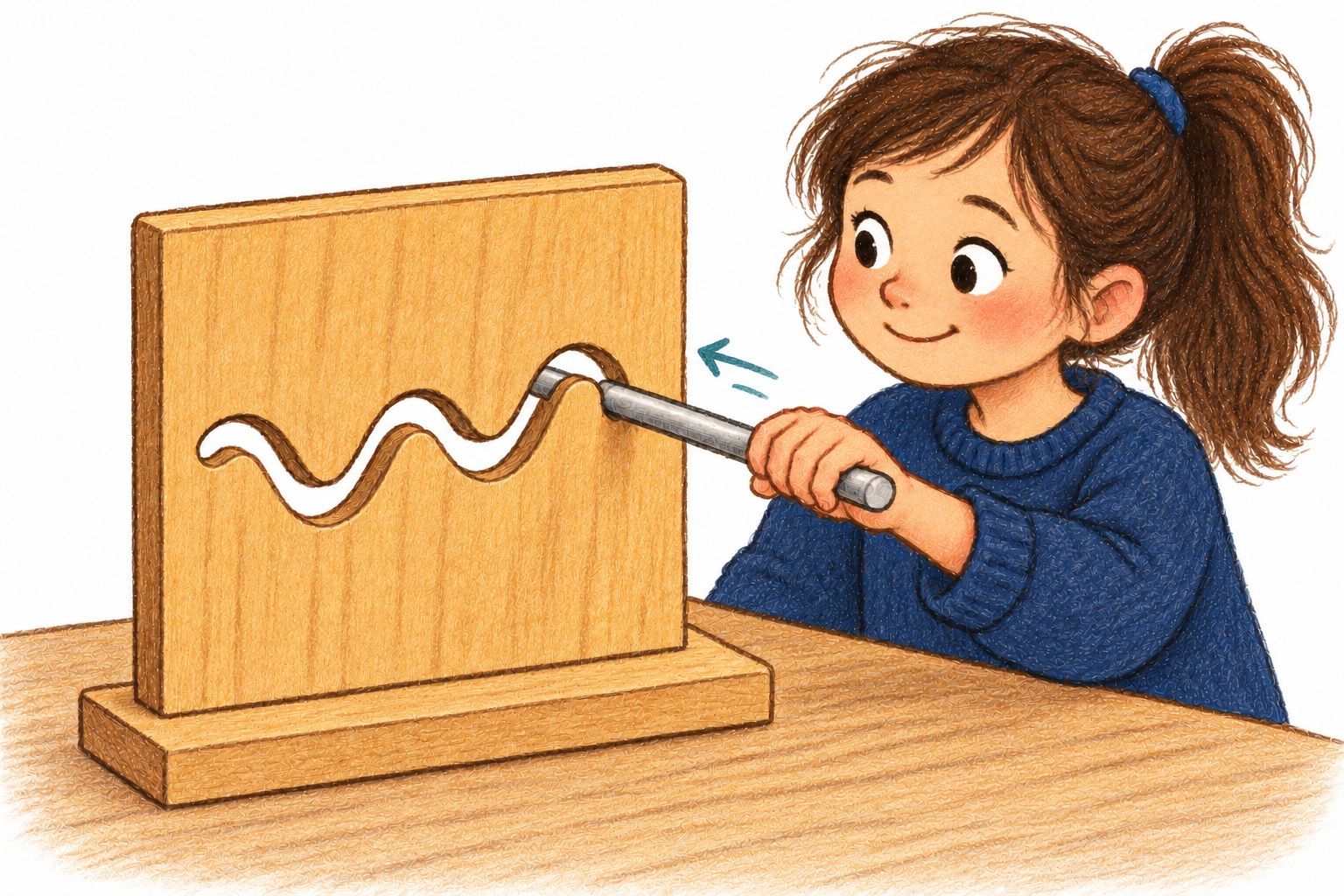}
  \caption{\textbf{The slot walls supply the rod's trajectory.} The hand pushes along the slot and yields across it, so the walls shape the path and the controller has far less of the curve to compute.}
  \label{fig:slot}
\end{figure}

In biology, action leaves the nervous system as spike trains on motor axons; in a robot, it leaves the policy as a vector of numbers. What the limb then does is a transformation of that signal by what lies downstream: muscle, tendon, and joint on one side, servo loops and motors on the other. The deepest difference is what the signal names. Robots command \emph{position}, where the limb should end up \citep{bronars2026tune}; the nervous system's final command is \emph{effort}, how hard each muscle pulls \citep{zajac1989muscle,enoka2015neuromechanics}. The brain also sets stiffness per task, deciding what must be precise and what can be left to the environment, whereas in a position-controlled robot the stiffness is set once by the servo gains, outside the policy's reach \citep{burdet2001central,franklin2008cns}. So a robot must compute every detail of a trajectory, while a human lets the environment supply part of it (\cref{fig:slot}). We first survey how robotics handles action (\cref{subsec:action_robotics}), then how biology does it (\cref{subsec:action_biology}), and finally draw the inspirations (\cref{subsec:action_inspiration}).

\subsection{Robotics}
\label{subsec:action_robotics}

Action, what a policy emits, is the boundary between the learned policy and the hand-tuned controllers that drive the motors. In the robotics literature the quantity that crosses it is nearly always a position, handed to a fixed servo stack that turns it into force (\cref{subsubsec:action_policy_position_servo_force}). The remaining choices concern only how the position is delivered: as a single target or a chunk (\cref{subsubsec:action_chunking}), and as a discrete token or a continuous number (\cref{subsubsec:action_representation}).

\subsubsection{The policy emits a position; the servo stack supplies the force}
\label{subsubsec:action_policy_position_servo_force}

\paragraph{The policy issues a position.} In nearly all imitation-learning and VLA work the policy outputs a position: joint angles, as in ACT and $\pi_0$ \citep{zhao2023learning,black2024pi0}, or an end-effector pose, as in Diffusion Policy and OpenVLA \citep{chi2023diffusion,kim2024openvla}. It runs at 5--50\,Hz \citep{chi2023diffusion,black2024pi0,kim2025finetuning}, hands each target to the servo stack and moves on to the next observation; how the target becomes torque is neither observed by the policy nor represented in its output.

\paragraph{The servo stack turns position into force.} The conversion is a cascade of nested feedback loops, each output the setpoint of the next, faster loop \citep{siciliano2009robotics,hutter2016anymal}. The policy's position target enters a joint PD servo, at up to 1\,kHz, that turns position and velocity error into torque via fixed gains \citep{franka2023fci}. That torque drives a torque loop, which sets the motor current the driver regulates at a still higher rate. This has two costs. A target is correct only under the gains it was trained with, so a policy transfers poorly across actuators unless their response is modeled or randomized \citep{hwangbo2019learning,peng2018simtoreal}. Modeling fits the real actuator's command-to-torque response and places it in the simulator; randomizing varies gains, latency, friction, and motor strength from one training episode to the next, so the policy cannot rely on any single response and must infer the one it faces from proprioceptive feedback. And the servo cannot tell intended contact from an obstacle: it pushes with a torque that grows with the error until something yields or a limit trips. Contact-rich behavior must therefore be restored beneath the policy, through force sensing \citep{liu2025factr}, compliant hardware \citep{pratt1995series}, or torque limits \citep{haddadin2017robot}.

\subsubsection{Compliance can be added beneath the policy, in hardware, torque, or target}
\label{subsubsec:action_compliance_control}
Compliance is the inverse of stiffness, the displacement yielded per unit of applied force, and a compliance controller regulates that relation rather than position alone. The robot then behaves like a chosen spring and damper about its target, so a push displaces it by a predictable amount and a contact error produces a bounded force \citep{hogan1985impedance,siciliano2009robotics}. Three schemes supply the spring, and they differ in where it lives.

\paragraph{Passive compliance puts the spring in the hardware.} The remote-center-compliance wrist arranges elastic elements so that a peg touching the chamfer of a hole is steered into alignment by the contact forces themselves, with no sensing or computation \citep{whitney1982quasi}. Series elastic actuators place a spring between motor and load, so that motor position becomes a force command through the spring's deflection \citep{pratt1995series}. Passive compliance acts at the speed of physics, but its stiffness is fixed once, at design time.

\paragraph{Impedance control puts the spring in the torque command.} The controller measures the joint's displacement and velocity from its target, computes the force a virtual spring and damper would exert there, and commands the motors to produce it \citep{hogan1985impedance}. Stiffness and damping become numbers the controller can change between tasks or within a movement. The price is that the actuators must deliver a commanded torque, through backdrivable transmissions, series elastic actuators, or joint torque sensing, as in the KUKA iiwa and Franka \citep{wensing2017proprioceptive,albuschaffer2007dlr,franka2023fci}. Conventional industrial arms do not: friction and reflected inertia in their high-ratio gearboxes make motor current a poor estimate of external force \citep{haddadin2017robot}.

\paragraph{Admittance control puts the spring in the position target.} On a stiff position-servoed arm the relation runs the other way. A wrist force sensor measures the contact, the controller computes the displacement a virtual spring would suffer under that force, and it shifts the position target by that amount, so the arm yields even though the servo beneath it never outputs a torque \citep{siciliano2009robotics}. This is how compliance is retrofitted to a position stack, and it is the form the correction of \cref{subsubsec:action_force_in_contact} takes.

\paragraph{Only a minority of policies emit something other than position.} Compliance control is well established beneath the policy, yet few policies emit anything but position. The exceptions are scattered: a few reinforcement-learned locomotion policies output joint torques directly \citep{chen2023learning}, a few manipulation policies output the force to apply along contact directions \citep{li2026forcevla2}, and a few emit an impedance target, a position together with the stiffness at which to hold it \citep{hogan1985impedance,hou2025adaptive}. 

\subsubsection{Action chunking decouples control rate from inference rate, at the cost of blindness mid-chunk and seams between chunks}
\label{subsubsec:action_chunking}

\paragraph{The policy is blind for the executed part of each chunk.} Action chunking has the policy predict a chunk of $H$ future targets from one observation rather than one target per step \citep{zhao2023learning,chi2023diffusion}. Fewer decisions per episode means less compounding error, and the control rate is no longer capped by inference speed: a VLA forward pass takes 50--300\,ms while joint servos expect targets at 50\,Hz or more, so a chunk of 0.5--2\,s at 10--50\,Hz lets the robot execute one chunk while the next is computed \citep{kim2024openvla,black2024pi0}. The cost is that a mid-chunk disturbance is answered only by the servo stack of \cref{subsubsec:action_policy_position_servo_force}, which holds the commanded positions against it whether or not that is appropriate.

\paragraph{Consecutive chunks must be joined.} Each chunk is generated from an observation that is stale by the inference time, and nothing forces it to agree with the chunk still executing, so the boundary can produce a discontinuity. Temporal ensembling averages the overlapping predictions of successive chunks \citep{zhao2023learning}; bidirectional decoding samples several candidates and keeps the one most consistent with the executed prefix \citep{liu2024bidirectional}; real-time chunking freezes the actions that will execute during inference and generates the rest conditioned on them, with a soft mask so that later actions stay free to respond to the new observation \citep{black2025realtime}.

\subsubsection{Token or number, absolute or delta, the representation changes how the position is encoded}
\label{subsubsec:action_representation}

\paragraph{Discrete heads inherit the language prior; continuous heads give precision and multimodality.} Binning each action dimension into a fixed vocabulary lets a pretrained VLM emit actions by next-token prediction \citep{brohan2023rt2,kim2024openvla}, but a chunk binned step by step is mostly redundant, since successive samples of a smooth trajectory are nearly equal; FAST compresses chunks with a discrete cosine transform before byte-pair encoding \citep{pertsch2025fast}. Continuous heads instead regress the numbers directly: $L_1$ regression is cheapest but collapses several valid actions onto a single compromise value \citep{chi2023diffusion}, and diffusion or flow matching recovers the full distribution over several denoising steps \citep{ho2020ddpm,chi2023diffusion,black2024pi0}. $\pi_0$ and $\pi_0$-FAST put the two heads on a shared backbone and isolate exactly this trade-off \citep{black2024pi0,pertsch2025fast}.

\paragraph{Delta actions pool across embodiments; absolute targets stop error compounding.} Delta end-effector actions \citep{brohan2022rt1,octo2024,kim2024openvla} are relative to the current pose, hence invariant to initial configuration and small enough to tokenize well, which is why cross-embodiment datasets favor them \citep{oneill2024openx}. But each delta is applied to wherever the arm actually is, so a lag or slip shifts every later target with it \citep{chi2023diffusion}. Absolute targets, joint angles in ACT and $\pi_0$ or poses in Diffusion Policy \citep{zhao2023learning,black2024pi0,chi2023diffusion}, live in a fixed frame, so each error is corrected at the next step rather than carried forward \citep{chi2023diffusion}. But that frame is tied to one body: joint angles to one arm's kinematics, poses to one base placement.

\subsection{Biology}
\label{subsec:action_biology}

We define action in biology as the output of the motor neurons of the spinal cord, the final common path through which every descending and reflex influence must pass \citep{sherrington1906integrative}. The brain's decisions arrive there as synaptic input, and what leaves is a train of action potentials that the muscle turns into movement \citep{kandel2013principles,zajac1989muscle}.

\subsubsection{The motor neuron integrates cortical, brainstem, and reflex drive into one command}
\label{subsubsec:action_motor_neuron_command}
Every alpha motor neuron receives a voluntary stream from cortex, a postural and locomotor stream from the brainstem, and a reflex stream from sensors in the muscle itself, and its output depends largely on their sum. The cortical stream is on the order of $10^6$ corticospinal axons \citep{lassek1939human,lassek1942human}, most ending on spinal interneurons, that drive on the order of $10^5$ motor neurons \citep{tomlinson1977numbers}. The brainstem stream carries posture and the go and speed commands for locomotion; the rhythm itself comes from central pattern generators, in the spinal cord for walking and running and in the brainstem for breathing and chewing, which produce it on their own \citep{grillner1981control,shik1966control}. The reflex stream closes fast loops that never wait on the brain: the spinal stretch reflex returns a limb toward its load within roughly 30\,ms, with a cortical component following 20--30\,ms later \citep{matthews1991human}, and the grip reflex of \cref{subsubsec:action_grasp_slip_reflex} tightens the fingers on a slipping object. The action that leaves the nervous system is therefore the sum of the brain's decision, locally generated rhythm, and sensory feedback.

\subsubsection{The population firing rate of a muscle's motor neurons is translated into effort}
\label{subsubsec:action_population_rate_effort}
More drive onto a muscle's motor pool recruits more of its neurons and fires each faster \citep{henneman1965functional,milnerbrown1973orderly}. The muscle turns that firing into a contraction that is its smoothed copy \citep{partridge1965modifications}, so the command is an effort. That force rises with the muscle's stretch over much of its working range, falls with its shortening speed \citep{hill1938heat,gordon1966variation}, and acts through a lever arm that changes with joint angle \citep{murray1995variation}, so the same firing gives a different torque at every posture, outside neural control. The stretch dependence makes an active muscle a spring, and the limb settles wherever its pull and the world agree.

\subsubsection{Opposing muscle pairs set torque and stiffness separately}
\label{subsubsec:action_opposing_pairs_torque_stiffness}
\begin{figure}[H]
  \centering
  \includegraphics[width=\linewidth]{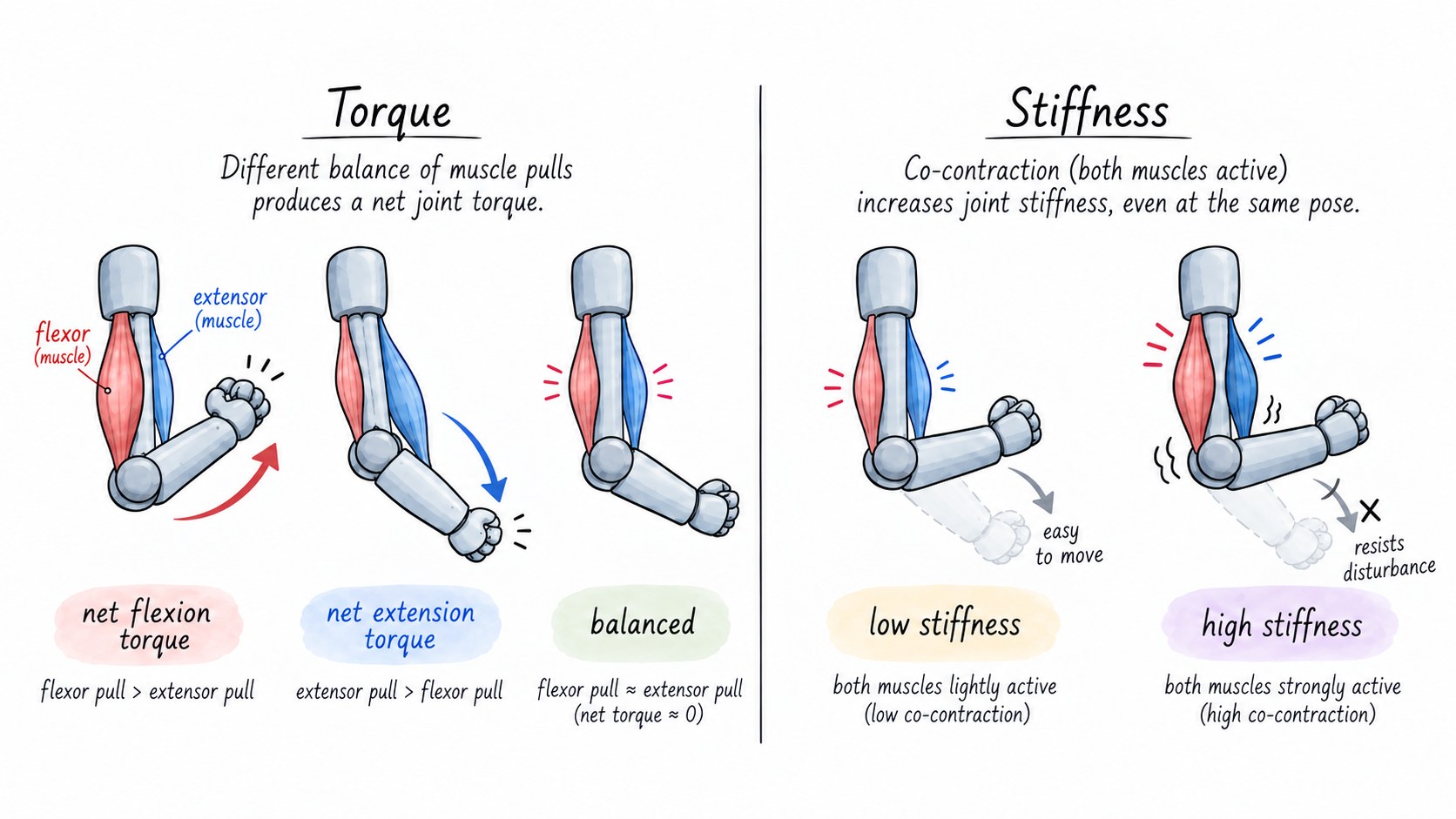}
  \caption{\textbf{Opposing muscles set torque and stiffness separately.} The difference of flexor and extensor pull sets the net torque, and their sum sets the stiffness at the same pose.}
  \label{fig:stiffness}
\end{figure}

Muscles come in opposing pairs, one pulling each way across a joint, so every joint is held between two springs. Two drives per joint thus set two quantities at once \citep{hogan1984adaptive}. Their difference is the net torque (\cref{fig:stiffness}, left): flexor pull above extensor pull flexes the elbow, the reverse extends it, and equal pulls leave it balanced. Their sum sets the stiffness (\cref{fig:stiffness}, right): co-contracting both muscles leaves the pose and the net torque unchanged, but the elbow that was easy to push aside under light co-contraction now resists the same push under strong co-contraction. Torque and stiffness are therefore two largely independent commands sent through one pair of muscles.

\subsubsection{Slip during a grasp is caught by a reflex that corrects grip force, with no position in the loop}
\label{subsubsec:action_grasp_slip_reflex}
During a grasp, grip force follows the motor pool's firing with a short delay \citep{johansson1984roles,partridge1965modifications}, and the whole task is specified in that quantity. The precision grip regulates the ratio of grip force to load force, holds it a fixed margin above the slip limit, and raises it $74 \pm 9$\,ms after a micro-slip \citep{johansson1987signals}, with no positional quantity anywhere in the specification. The slip itself is read from the skin's mechanoreceptors, whose classes and the signals they combine into a slip event are described in \cref{subsubsec:touch_mechanoreceptor_classes}. The correction is a reflex rather than a decision: a sudden pull on the held object triggers the same grip increase within about 70\,ms, before any voluntary response \citep{johansson1992pullingloads}, and anesthetizing the fingertips delays and weakens it, so the loop is closed mainly by those cutaneous afferents rather than by muscle sense \citep{johansson1992anesthesia}; \cref{sec:touch} takes up what the loop should look like in a robot.

\subsection{Inspiration}
\label{subsec:action_inspiration}
From the two subsections above we draw two lessons for the policy interface. The first is that compliance control matters because it lets the world supply parameters the policy would otherwise have to compute (\cref{subsubsec:action_stiffness_matters}). The second is our proposal for achieving it on the arms most policies run on, which accept only position targets: keep the position output and add a stiffness beneath it, with a touch sensor measuring the contact so the controller can take from it where to yield and how hard to press, moving the position target beneath the policy (\cref{subsubsec:action_force_in_contact}).

\subsubsection{Use compliance so the world supplies what the policy would otherwise compute}\label{subsubsec:action_stiffness_matters}
\begin{figure}[H]
  \centering
  \includegraphics[width=0.7\linewidth]{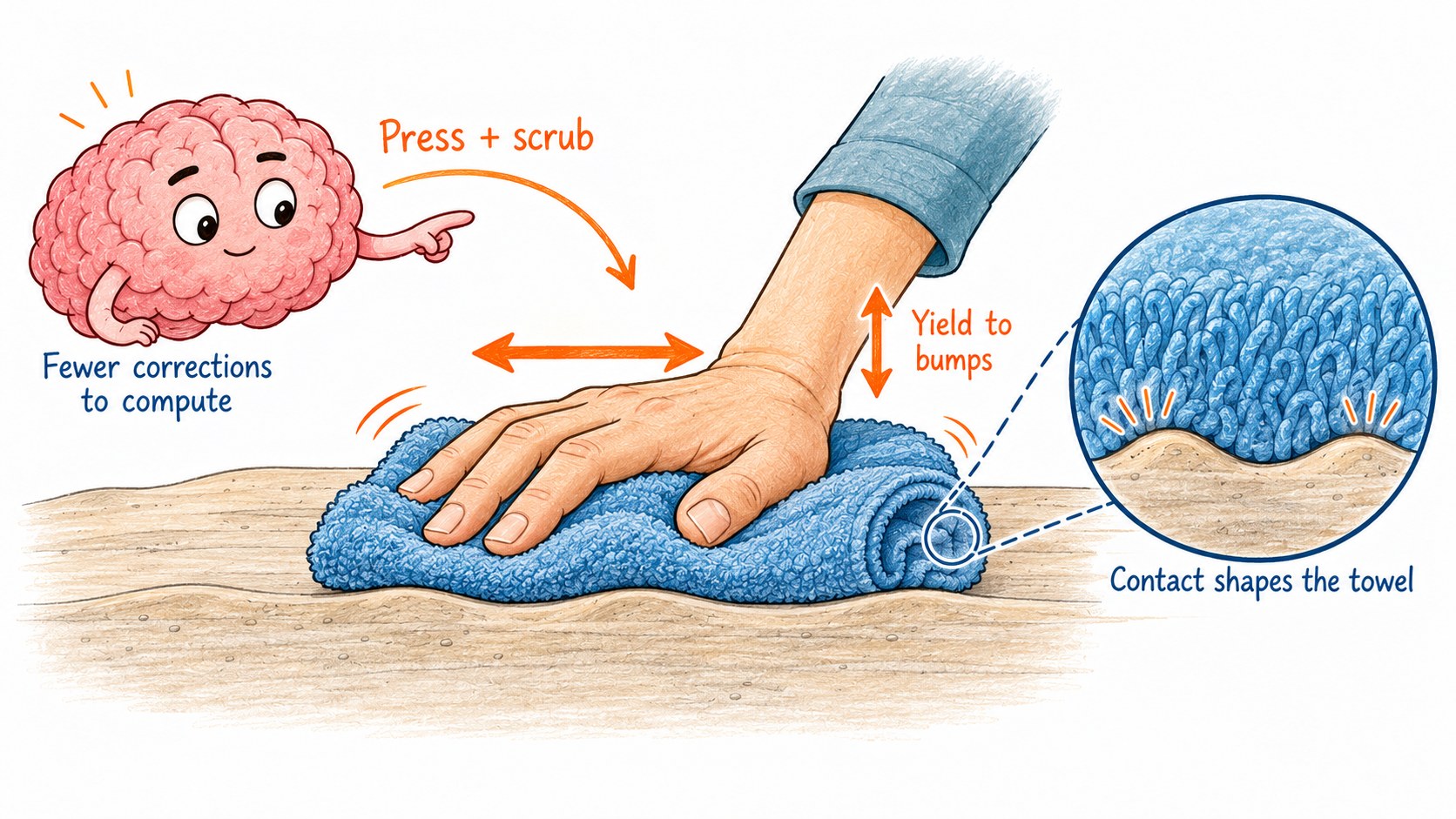}
  \caption{\textbf{Contact supplies the surface detail.} A yielding wrist and a soft towel follow the bumps, so the controller computes fewer corrections.}
  \label{fig:compliance_towel}
\end{figure}

Consider carrying a rod through a curved slot (\cref{fig:slot}). A robot with only position commands has to perceive the precise shape of the whole curve, and any error at contact becomes uncontrolled force through a stiff servo. A human does not track the curve. The hand pushes with high stiffness along the slot and low stiffness across it, and lets the board do the shaping: as long as the push is roughly along the slot, the wrist tolerates some wobble, and whatever the wall presents is fine, because the hand yields to it. The same holds when a yielding wrist scrubs a bumpy surface (\cref{fig:compliance_towel}). The controller's only job is to choose which axes to hold and which to yield; the curve of the slot and the height of each bump arrive through contact for free.

\subsubsection{On the position-controlled arms most policies run on, keep the position output, add a stiffness beneath it, and let touch set where to yield and how hard to press}\label{subsubsec:action_force_in_contact}

We propose that the policy, like the muscle pair of \cref{fig:stiffness}, command a stiffness alongside its target, and that the stiffness be realized beneath it: by impedance control where the joints deliver torque, and by admittance on the position-controlled arms most policies run on, where even Adaptive Compliance Policy had to recover its stiffness through a wrist force sensor \citep{hou2025adaptive}. We take the position-controlled arm as the main case and let touch set where to yield and how hard to press.

\paragraph{Touch tells the controller where to yield and how hard to press.} On a position-controlled arm we propose adding the stiffness as admittance (\cref{subsubsec:action_compliance_control}): a touch sensor measures the contact and the controller moves the position target in response. Against a hard wall, the direction of the sensed force is the direction to yield in, and the rest of the commanded motion goes through unchanged. Against a soft object, the magnitude decides when to stop closing. This rule can be written as code by a language model, as in Code as Policies \citep{liang2023code}, or it can be learned. OmniTacTune \citep{yu2026omnitactune} keeps the visual policy frozen and adds a small residual network, trained by reinforcement on the real robot, that reads the tactile signal and corrects the end-effector target while each chunk executes, with the residual scaled to 5--15\% of the base action. It raises success on contact-rich tasks from 5--40\% to 85--100\% within 40--80 minutes of training, on an arm that never outputs a force.

\clearpage
\section{Touch}
\label{sec:touch}

\begin{figure}[H]
\centering
\includegraphics[width=0.5\textwidth]{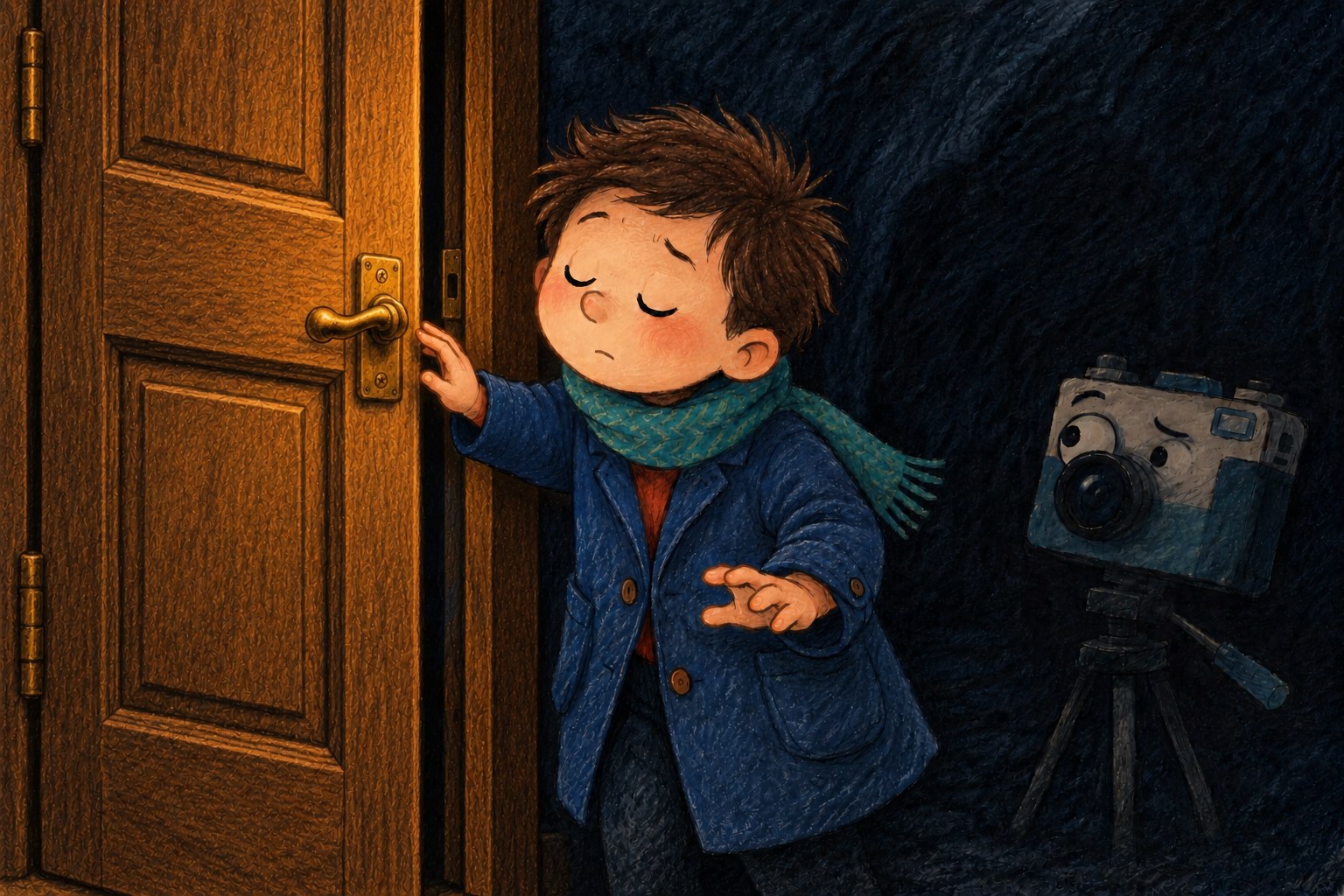}
\caption{\textbf{Touch alone finds the handle and opens the door.} With eyes closed the hand feels its way to the handle, turns it and pushes the door open; in the dark a camera has nothing to offer.}
\label{fig:touch_sketch}
\end{figure}

A human can find a door handle in the dark and open the door without a single glance (\cref{fig:touch_sketch}). Most current robot policies cannot, because they rely on vision \citep{black2024pi0,geminirobotics2025,luo2025tactile}. One might argue that, with the lights on, vision could cover what touch does, but the human evidence says otherwise: with anesthetized fingertips and the hand in plain view, subjects still pick up some objects, yet they apply excessive, poorly scaled forces and drop others \citep{johansson1992anesthesia,monzee2003,augurelle2003}.

Beyond empirical evidence, a first-principles view says the same. Touch supplies the force in Newton's second law, $F = ma$; vision sees only displacement, velocity, and acceleration, the right-hand side, and can infer force from them only indirectly. A robot that senses only motion sees the effect of every interaction and must guess its cause, and that cause is what physical intelligence must reason about. This section examines how humans use touch and what robotics can learn from it.

\subsection{Robotics}
\label{subsec:touch_robotics}

Tactile sensing has recently attracted growing attention as an input modality for robot learning \citep{luo2025tactile}. We first discuss sensors and then how the policy consumes their signal.

\subsubsection{No sensor family matches human skin on resolution, speed, and force at once}
\label{subsubsec:touch_robotic_sensors}

Every robotic touch sensor is a physical filter. The transduction principle fixes what the skin can resolve in space, how fast it responds, and which components of force it reports, and no current family scores well on all three \citep{dahiya2010tactile,luo2025tactile,yamaguchi2019recent}. \Cref{tab:touch_sensors} shows the pattern: camera-based pads resolve tens of microns, piezoelectric films catch kilohertz transients but are blind to steady pressure, and the cheapest arrays read normal force only \citep{dahiya2010tactile,howe1993dynamic,tenzer2014takktile}.

Durability is a further limit. Touch sensors sit on the part of the robot built to collide with the world, yet the soft elastomer that makes a sensor sensitive wears out long before the electronics beneath it \citep{lambeta2020digit,bhirangi2021reskin}.

\begin{table}[H]
\centering
\footnotesize
\begin{tabular}{@{}P{2.1cm}P{2.9cm}P{1.7cm}P{1.6cm}P{2.7cm}P{2.8cm}@{}}
\toprule
\textbf{Sensor type} & \textbf{How it works} & \textbf{Spatial res.} & \textbf{Speed} & \textbf{Strengths} & \textbf{Weaknesses} \\
\midrule
Piezoresistive \citep{dahiya2010tactile,huang2024vitac} & Pressure squeezes a conductive polymer and lowers its resistance & mm-scale & tens of Hz & Cheap, thin, flexible; scales to large arrays & Hysteresis; reading drifts with load history; normal force only \\
Capacitive \citep{dahiya2010tactile,yousef2011tactile} & Pressure narrows the gap between two electrodes (touchscreen principle) & mm-scale & $\sim$100\,Hz & Sensitive to light contact; mature technology & Picks up electrical interference; mostly normal force \\
Piezoelectric \citep{howe1993dynamic,dahiya2010tactile} & A polymer film emits a voltage when strained & mm-scale & kHz & Captures vibration, texture, and the onset of slip & Responds only to \emph{change}; blind to steady pressure \\
Barometric \citep{tenzer2014takktile} & Air-pressure chips cast inside rubber, one chip per taxel & mm-scale & $>$100\,Hz & Robust, inexpensive, factory-calibrated & Coarse; normal force only \\
Magnetic \citep{bhirangi2021reskin,tomo2018uskin} & Magnetized elastomer over magnetometer chips; deformation shifts the field they read & mm-scale & hundreds of Hz & Three-axis force including shear; thin; replaceable skin & Hysteresis; signal drifts as the elastomer ages, so each new skin needs recalibration \\
Multimodal fluid-filled fingertip \citep{fishel2012biotac,wettels2008biomimetic} & Fluid-filled fingertip with impedance electrodes, a pressure transducer and a thermistor & tens of sensing points per fingertip & $\sim$100\,Hz, with a kHz vibration channel & Pressure, vibration, and heat flow together, the closest to a human fingertip & Expensive; signals need learned decoding \\
Vision-based \citep{yuan2017gelsight,lambeta2020digit,ward2018tactip} & A small camera films an illuminated gel or pin array deforming under contact & tens to hundreds of \textmu{}m & camera rate, 30--60\,Hz & Spatial resolution finer than human skin; full contact geometry, shear from marker motion, slip onset & Bulky; slow; needs image processing; gel wears out \\
\bottomrule
\end{tabular}
\caption{No single sensor family combines fine resolution, kilohertz speed, and shear. Resolution and speed are order-of-magnitude values typical of each family, and the cited works are representative examples. A \emph{taxel} (``tactile pixel'') is one sensing point on the skin.}
\label{tab:touch_sensors}
\end{table}

\paragraph{Touch rarely extends beyond the fingertips.} Most sensorized hands carry touch only at the fingertips, as caps or gripper pads, and leave the phalanges and palm bare \citep{lambeta2020digit,luo2025tactile}. Even at the tip, the sensor covers the pad and not the sides of the fingers that steer chopsticks and scissors \citep{dahiya2010tactile,yamaguchi2019recent}. Whole-hand skins remain rare: the scalable ones read normal force only \citep{sundaram2019stag}, and shear-capable coverage of palm and fingers exists in a few research hands \citep{tomo2018uskin,zhao2025ftac}.

\subsubsection{Policies now take touch as an input, but data and control rate hold it back}
\label{subsubsec:touch_robotics_models}

\paragraph{Touch enters the policy as an extra image or as learned tokens.} The design choice is how the tactile signal is represented. Because vision-based sensors emit images, the simplest approach treats the tactile image as an additional camera view and reuses the vision encoder \citep{hao2025tla}. Touch-specific pretraining improves on this borrowed encoder, either through self-supervised learning on tactile video or through contrastive alignment of touch with vision and language \citep{higuera2024sparsh,yang2024unitouch}. Other policies learn tactile tokens of their own: a point cloud from piezoresistive pads \citep{huang2024vitac}, a single token from the force history \citep{huang2025tactilevla}, latent tokens from a dedicated tactile corpus \citep{n0vtla2026}, or a few tokens compressing the tactile history \citep{wang2026tacmamba}.

\paragraph{Cost, standardization, scale, and speed still hold touch back.} Tactile sensors are expensive and wear quickly, so most demonstrations are collected without them. Each sensor family emits a different signal, an image, a three-axis force, or a pressure array, so data from different platforms cannot be pooled without a learned shared latent space \citep{feng2025anytouch,park2026tactx}. Tactile corpora are therefore orders of magnitude smaller than video \citep{ye2026tacgen,n0vtla2026}. Finally, policies run at camera rate, 10--60\,Hz, and miss the hundreds-of-hertz transients that signal events such as slip (\cref{tab:touch_sensors}).

\subsection{Biology}
\label{subsec:touch_biology}

\subsubsection{Four receptor classes encode touch in spike timing}
\label{subsubsec:touch_mechanoreceptor_classes}

\needspace{16\baselineskip}
\begin{figure}[H]
\centering
\includegraphics[width=0.62\textwidth]{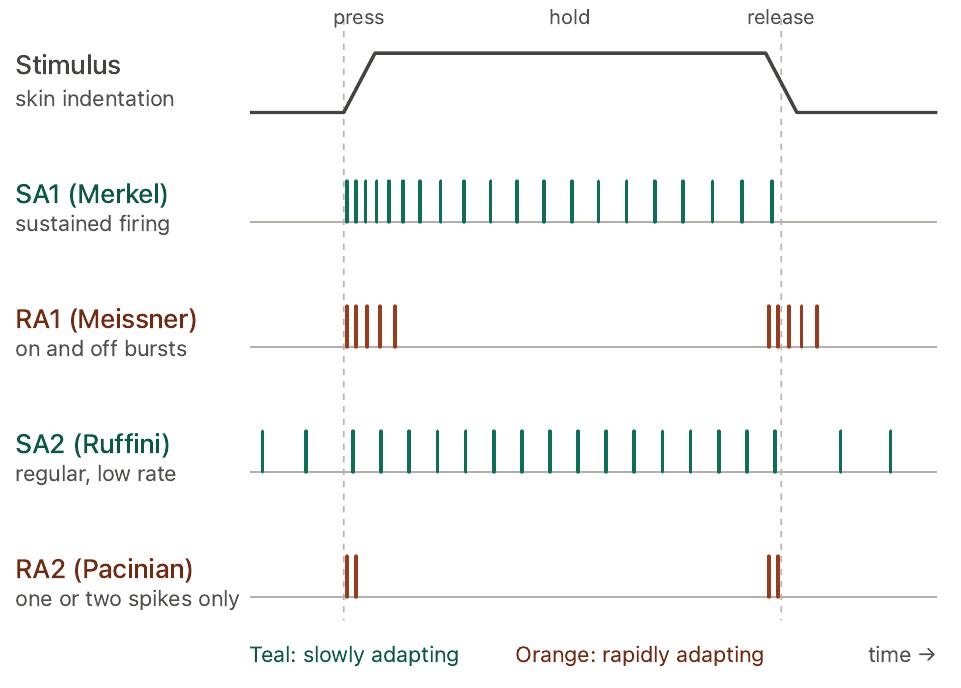}
\caption{Slowly adapting (SA) afferents report steady pressure and rapidly adapting (RA) ones report change, so the four classes split what touch carries. The traces are schematic responses to a ramp-and-hold skin indentation: SA afferents sustain firing during the hold, RA afferents respond transiently during pressing and release, and SA2 afferents also fire at a low rate before contact, driven by resting skin stretch. RA afferents can also fire during ongoing motion or vibration. Spike counts and timing are illustrative rather than measured; responses depend on the stimulus and afferent \citep{johnson2001roles,abraira2013sensory}.}
\label{fig:mechanoreceptor_firing}
\end{figure}

\paragraph{A mechanoreceptor reports only through spike timing.}
Touch starts in the skin. Buried at different depths are the endings of sensory nerve fibers, most ending in a specialized structure of tissue: the mechanoreceptors \citep{johansson1979tactile,johnson2001roles,abraira2013sensory}. When the capsule deforms, the nerve ending fires brief, stereotyped electrical pulses called spikes. A spike cannot be made bigger to say ``more''; a stronger press makes the ending fire sooner and more often, so everything a fiber reports is carried in when its spikes come and how often \citep{adrian1926impulses,johnson2001roles} (\cref{fig:mechanoreceptor_firing}).

\paragraph{Tactile percepts arise from combining the four receptor classes.}
A receptor's wrapping and depth set two properties: how large a patch of skin it listens to, and whether it reports steady contact (slowly adapting, SA) or only change (rapidly adapting, RA). The four combinations give the four classes in \cref{tab:mechanoreceptors}: SA1 reports sustained pressure and fine shape, RA1 the onset of movement, SA2 skin stretch and force direction, and RA2 high-frequency vibration. None of them reports a tactile feeling on its own. Texture, weight, and slip each arise from combining the four channels \citep{johansson2009coding}. When a held glass begins to slide, RA1 afferents fire the most reliable burst at the onset of the slip, with RA2 and SA1 afferents contributing, while the grip force itself has not changed \citep{johansson1987signals}: a burst of change with no change in squeeze signals slip.

\begin{table}[H]
\centering
\footnotesize
\begin{tabular}{@{}P{2.6cm}P{2.8cm}P{2.4cm}P{2.4cm}P{3.4cm}@{}}
\toprule
\textbf{Receptor} & \textbf{Where it sits} & \textbf{Skin patch it hears} & \textbf{Steady or change?} & \textbf{Everyday role} \\
\midrule
SA1 (Merkel ending) & Just under the surface; densest at the fingertip & Tiny (2--4\,mm), sharply bounded & Keeps firing under steady pressure & Reading fine shape and texture; edges and curvature, the Braille channel \\
RA1 (Meissner corpuscle) & Just under the surface, in the hairless skin of the hand & Small (3--5\,mm) & Fires only when contact changes & Catching micro-slips and low-frequency flutter, the trigger for tightening your grip \\
SA2 (Ruffini ending) & Deeper in the skin; sparse & Large, blurry borders & Keeps firing & Sensing skin stretch, finger posture, and which way a force is pulling \\
RA2 (Pacinian corpuscle) & Deepest of all; sparse, roughly 350 per finger & Very large, a whole finger or the hand & Fires only at sharp changes & Feeling high-frequency vibration through a held tool, pen on paper \\
\bottomrule
\end{tabular}
\caption{Four afferent classes split touch by receptive-field size and adaptation \protect\citep{johansson1979tactile,vallbo1984properties,johnson2001roles}. The pairing of SA2 afferents with Ruffini endings is disputed, since the human fingertip holds few such endings \protect\citep{pare2003paucity}. \Cref{fig:mechanoreceptor_firing} uses the same codes.}
\label{tab:mechanoreceptors}
\end{table}

\subsubsection{Successive contacts build an object representation beyond geometry}

\paragraph{S1 estimates the surface under one fingertip.}
Discriminative-touch signals from a fingertip reach the primary somatosensory cortex (S1) through the dorsal column--medial lemniscus pathway (\cref{fig:touch_pathway}) \citep{kandel2013principles}. The input is dense, roughly 17{,}000 tactile afferents for the glabrous skin of the hand \citep{johansson1979tactile,corniani2020tactile}, and S1 turns it into a first estimate of the surface under each fingertip: area 2 integrates cutaneous input with finger posture to recover the curvature and orientation of the surface in contact \citep{kandel2013principles}.

\paragraph{Parietal cortex assembles the whole object from many contacts.}
From S1 the signal passes to the secondary somatosensory cortex (S2) and on to posterior parietal cortex, where processing moves from individual skin contacts to the object. Parietal cortex relates each contact to hand and arm posture, placing it in space, and so combines touches made by different fingers at different times into an estimate of the object's three-dimensional form \citep{kandel2013principles}.

\begin{figure}[H]
\centering
\includegraphics[width=0.8\textwidth]{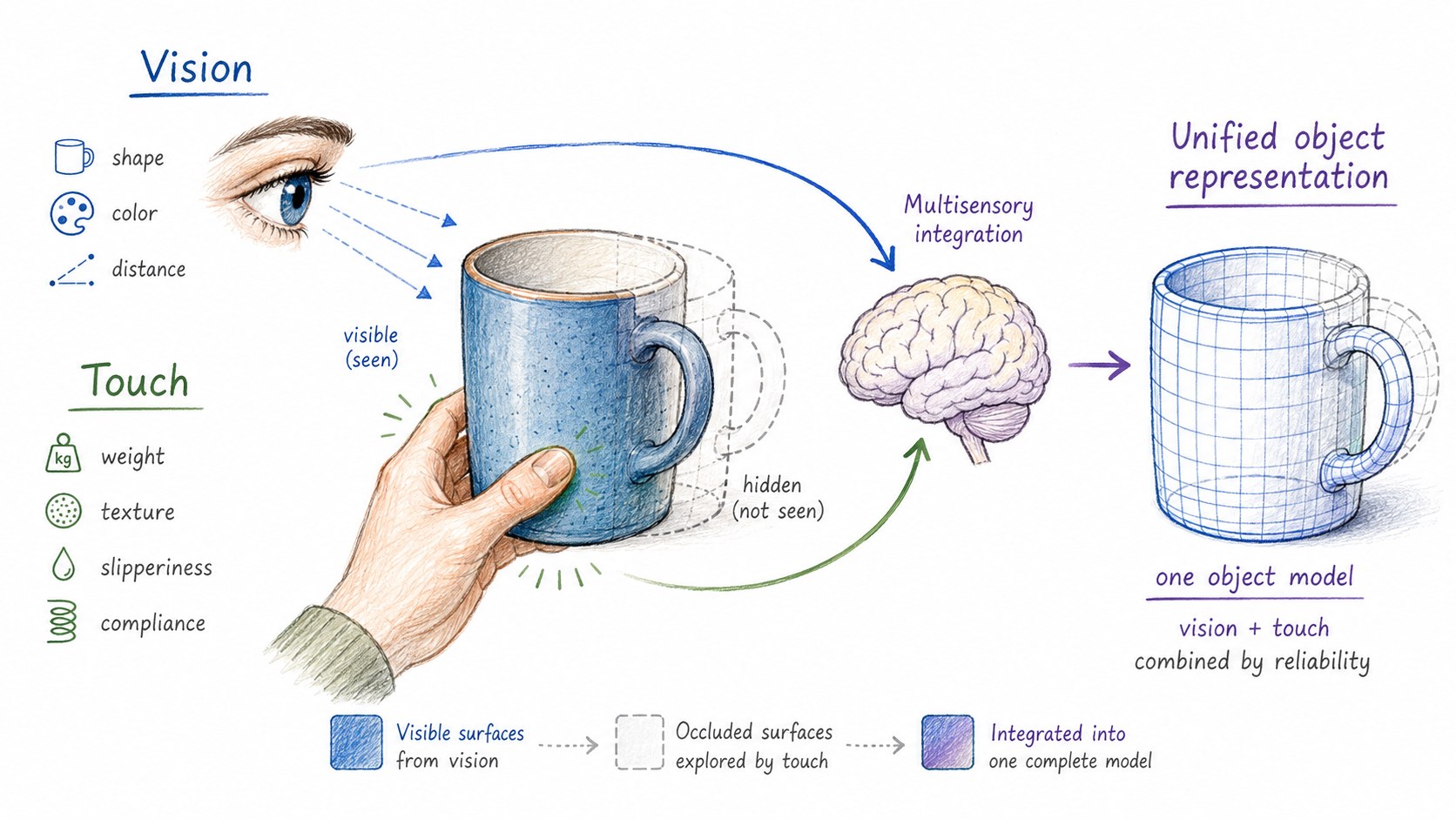}
\caption{Vision and touch build a single object model, and each sense fills in what the other cannot. Vision covers the visible surfaces, while touch supplies weight, texture, slipperiness, and compliance, and explores the occluded back.}
\label{fig:object_representation_sketch}
\end{figure}

\paragraph{Vision and touch converge into one object representation.}
Tactile and visual shape processing converge in regions such as the lateral occipital tactile-visual area (LOtv), which responds to objects explored by touch as well as those seen \citep{amedi2001lotv}. The two estimates are weighted by their reliability: clear vision dominates, and touch gains weight as visual information becomes uncertain \citep{ernst2002humans}. The result is one object representation that combines visible geometry with properties only contact reveals, such as weight, texture, and compliance (\cref{fig:object_representation_sketch}).

\subsubsection{Tactile reflexes correct grip within about \texorpdfstring{75\,ms}{75 ms}, faster than any visual correction}
\label{subsubsec:touch_motor_control}

\begin{figure}[H]
\centering
\includegraphics[width=0.54\textwidth]{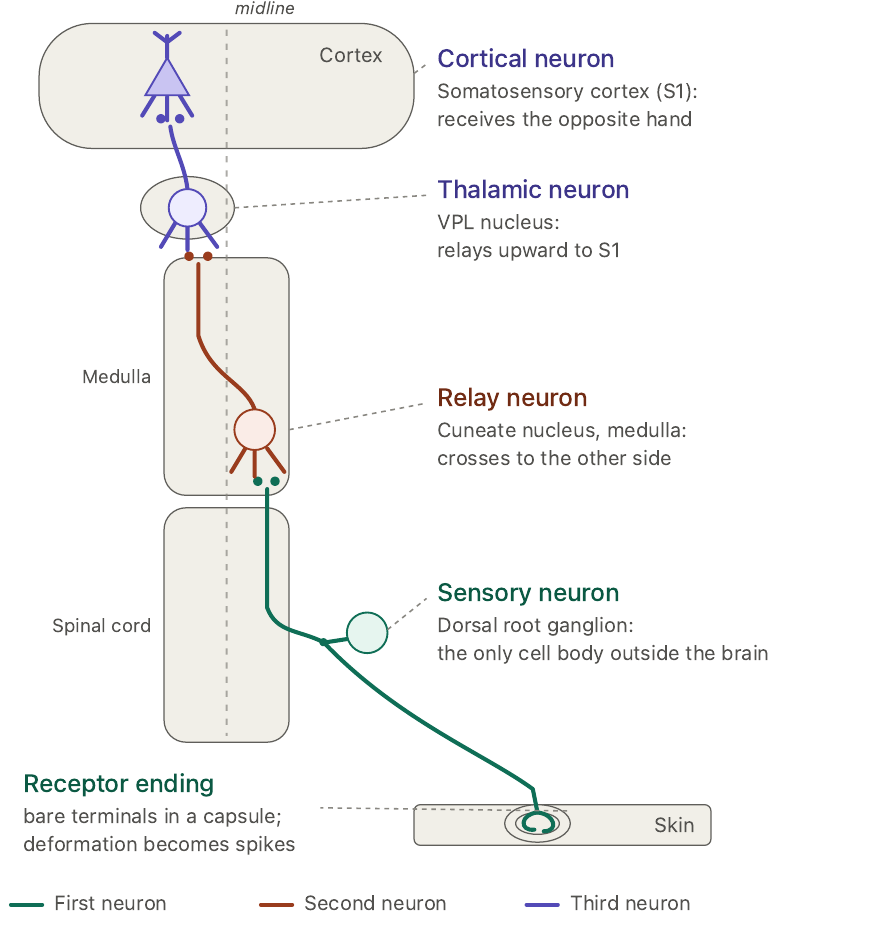}
\caption{Three neurons carry a fingertip signal to primary somatosensory cortex (S1), a chain short enough to leave time for a tactile correction within about 75\,ms. The first-order neuron ascends on the same side through the spinal dorsal column to the cuneate nucleus of the medulla; the second-order neuron crosses there and ascends to the ventral posterolateral (VPL) thalamus; the third-order neuron projects to S1 on the side opposite the stimulated hand. Each line ends in a dot at its synapse, and the legend colors mark the three cell bodies. Other branches and circuits are omitted \citep{kandel2013principles}.}
\label{fig:touch_pathway}
\end{figure}

Touch does not have to wait for a full object representation before it acts, because tactile signals reach the motor system along two pathways of different speed. The reflex pathway runs through the spinal cord and brainstem, or through the chain of \cref{fig:touch_pathway} to S1 and the neighboring motor cortex: when a held object begins to slip, grip force rises $74 \pm 9$\,ms later, before the slip reaches awareness \citep{johansson1987signals,johansson1988rapidload,johansson2009coding}. The deliberate pathway passes through conscious perception and planning and takes roughly 150\,ms or more, the range of voluntary reaction times \citep{johansson2009coding}. The reflex pathway is what vision cannot match, since even the fastest visual corrections take roughly 100\,ms or more (\cref{subsubsec:vision_relies_on_predictions}).

\subsubsection{The brain predicts touch and acts on the discrepancy}\label{subsubsec:touch_predictive_delta_adjustment}

Why can you not tickle yourself? Because the brain predicts how your own action will feel: a copy of every motor command is used to forecast the touch it should produce \citep{blakemore1998tickle,wolpert1995internal,flanagan2006control}. The difference between predicted and actual touch is the control signal: either the prediction was off or the execution fell short. It serves two purposes at once. It triggers a correction of the action in progress \citep{johansson1984roles}, and it updates the predictor, a role attributed to the cerebellum \citep{wolpert1998internal}, so the next lift starts with nearly correct grip forces. We return to the cerebellum, and to what it shares with a robot world model, in \cref{subsec:cerebellum_world_model_biology}.

\subsubsection{Descending control suppresses predictable and irrelevant tactile channels}
\label{subsubsec:touch_suppress_rest}

\begin{figure}[H]
\centering
\includegraphics[width=0.7\textwidth]{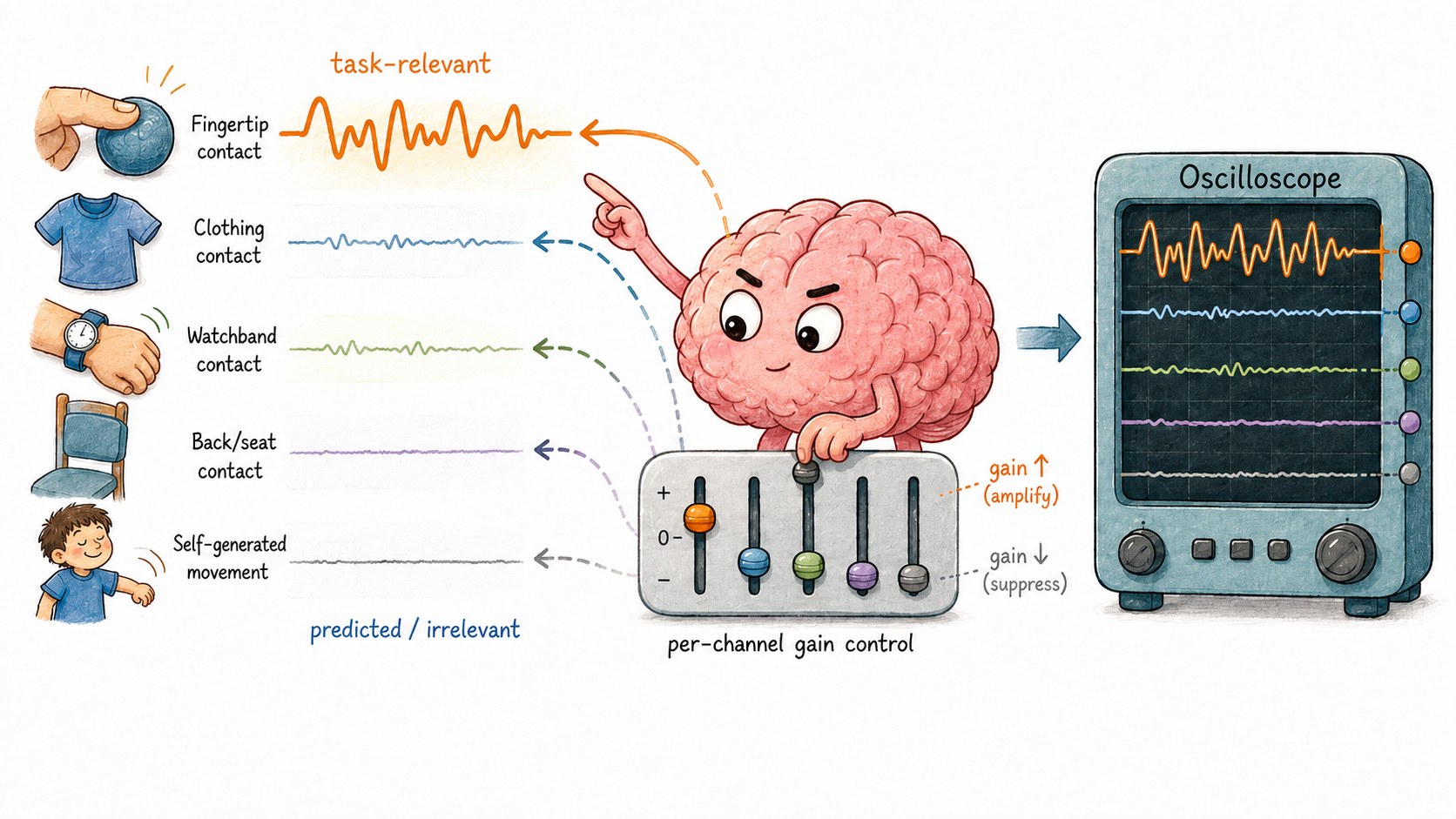}
\caption{Descending signals from cortex turn down predicted or irrelevant tactile channels and turn up the task-relevant one, so the fingertip signal arrives at high contrast.}
\label{fig:oscilloscope_sketch}
\end{figure}

Prediction and attention together decide what the tactile system transmits. The skin streams signals constantly (clothes, watchband, chair), yet by a textbook estimate, of roughly $10^6$ bits/s entering the nerves only about five bits/s reach perception \citep{zimmermann1986sensory}. The selection is top-down: descending signals set the gain at each relay, turning down channels that prediction marks as expected, such as touch from your own movement, or that attention marks as irrelevant, and turning up the relevant one \citep{chapman1994active,tiriac2016gating,wimmer2015}. Filtering this early spares downstream spikes, which are costly \citep{attwell2001energy,lennie2003cost}, and delivers the attended contact to cortex at high signal-to-noise (\cref{fig:oscilloscope_sketch}).

\subsection{Inspiration}
\label{subsec:touch_inspiration}

Touch should measure the contact force that compliance control needs (\cref{subsubsec:touch_enabled_compliance_control}), reveal the physical properties of objects (\cref{subsubsec:touch_tactile_object_representation}), and gather evidence actively with small probes (\cref{subsubsec:touch_active_probing}).

\subsubsection{Represent touch as a force field first, and let it drive compliance control}\label{subsubsec:touch_enabled_compliance_control}

We propose that touch enter the model as a force field: a contact force per point on the skin, with normal and shear components. Most sensor families can be calibrated to this form, at least approximately \citep{yuan2017gelsight,bhirangi2021reskin}, so data from different sensors pool directly and one embedding serves them all, where cross-sensor encoders must first learn a shared latent space \citep{feng2025anytouch,park2026tactx}. The force field keeps only force and discards texture, vibration, and temperature. Raw signals keep all of these and also work \citep{higuera2024sparsh,n0vtla2026}, but each sensor then needs its own tokenizer and enough data to train it.

The model uses that representation in two ways, both through the contact force in strength and direction. Against a fragile or soft object, the strength decides when to stop closing; against a hard surface, the direction marks the axis to yield along, the compliance rule of \cref{subsubsec:action_force_in_contact}. Either form of the model's action plan can apply it (\cref{subsubsec:planner_commander_commander_action}): as code, the model writes the rule, yield along the sensed normal and stop at a force threshold \citep{liang2023code}; as a reactive policy head, the head takes the tactile tokens as input and corrects the trajectory at the control rate.

\subsubsection{Train the model to predict the tactile consequence of each command}
\label{subsubsec:touch_tactile_object_representation}

Given past observations and a new command, the model should predict the sensory consequence \citep{higuera2026vtwm,liu2026taco,tacforesight2026}: if past touch suggested high friction, how far the object slides under a given push; if it suggested a soft sponge, how much the sponge deforms under a given grip. This forces the model to infer the properties that geometry \citep{park2019deepsdf,kerbl2023gaussian} leaves out and contact measures directly: mass, friction, compliance, and fragility, on which grip force and the risk of slip or breakage depend \citep{ye2026tacgen,lederman1987hand}. We train on exactly such data: the robot's own executed actions, command as question and tactile outcome as answer, free of annotation, through the force-field calibration of \cref{subsubsec:touch_enabled_compliance_control} so one model serves every sensor. 

The same prediction pays off at run time. Once the real touch arrives, the model compares it with what it expected and acts on the difference, as the brain does (\cref{subsubsec:touch_predictive_delta_adjustment}). A grip that slips or a surface that gives more than predicted shows up as a mismatch at the first contact, and the sign and size of that mismatch say which action compensates: tighten the grip, ease the press.

\subsubsection{Probe actively to test hypotheses about properties that cannot be seen}
\label{subsubsec:touch_active_probing}
\begin{figure}[H]
\centering
\includegraphics[width=0.75\textwidth]{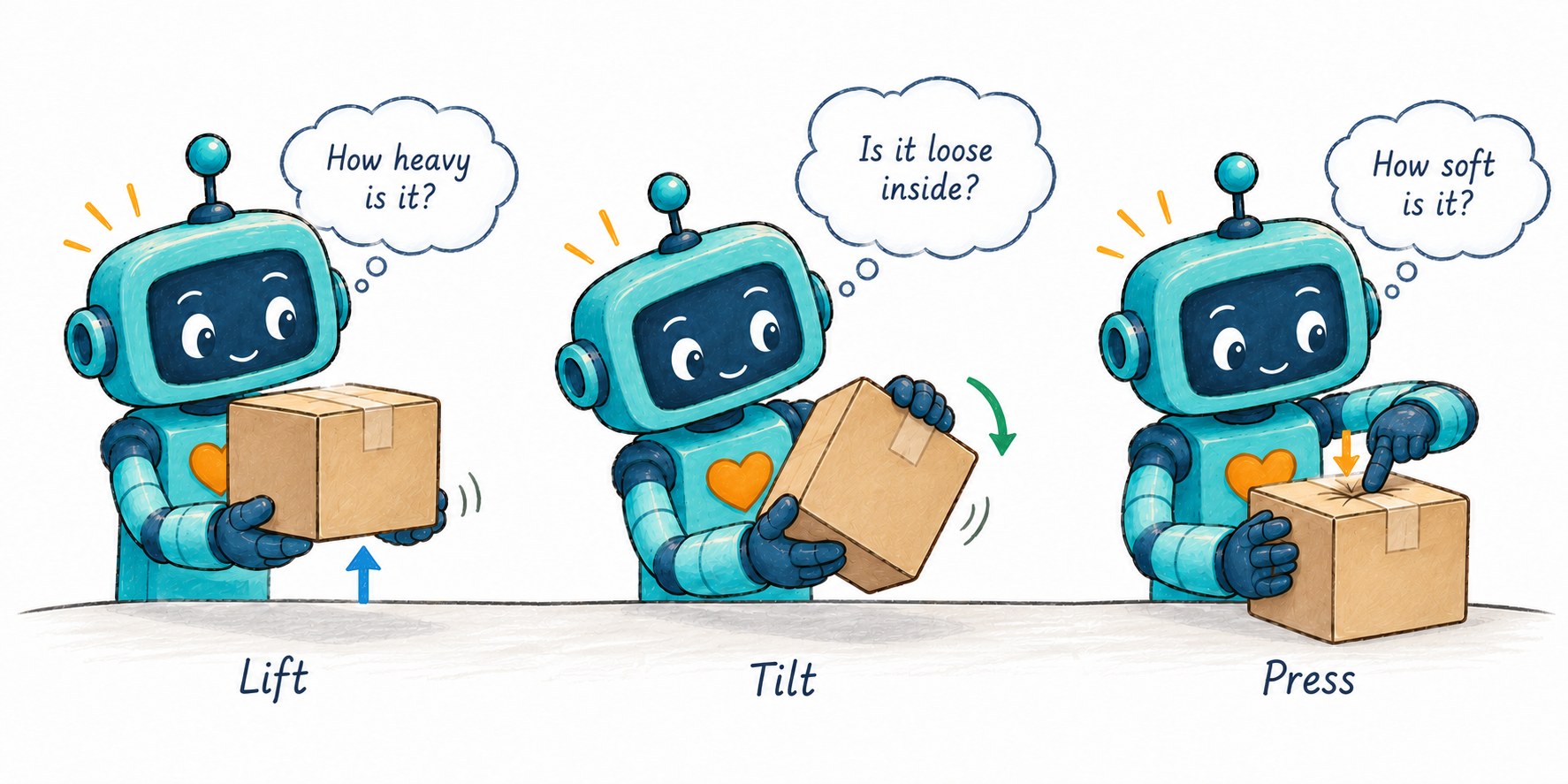}
\caption{When an object's properties cannot be seen, the robot should probe before it commits. Lifting, tilting, and pressing within safe bounds each test a hypothesis the camera cannot settle.}
\label{fig:haptic_probing}
\end{figure}

When the property that matters cannot be seen, the robot should create the evidence before committing: lift the object a few millimeters to estimate load, tilt it to feel whether anything shifts inside, or press gently to measure stiffness, and let the outcome set the following strategy (\cref{fig:haptic_probing}) \citep{bohg2017interactive,lederman1987hand}. Probes should be uncertainty-driven, choosing the contact where competing hypotheses predict different outcomes \citep{mackay1992information,fishel2012bayesian,gordon2025active}. The same probes are the cheapest self-improvement data: each is an action with a measured consequence, so it trains the model's expectation of what its actions will do, data that passive demonstrations rarely contain (\cref{subsubsec:cerebellum_world_model_predictor_beside_commander}) \citep{gibson1962active,pinto2016curious}.

\clearpage
\section{Vision}
\label{sec:vision}

\begin{figure}[H]
\centering
\includegraphics[width=0.7\textwidth]{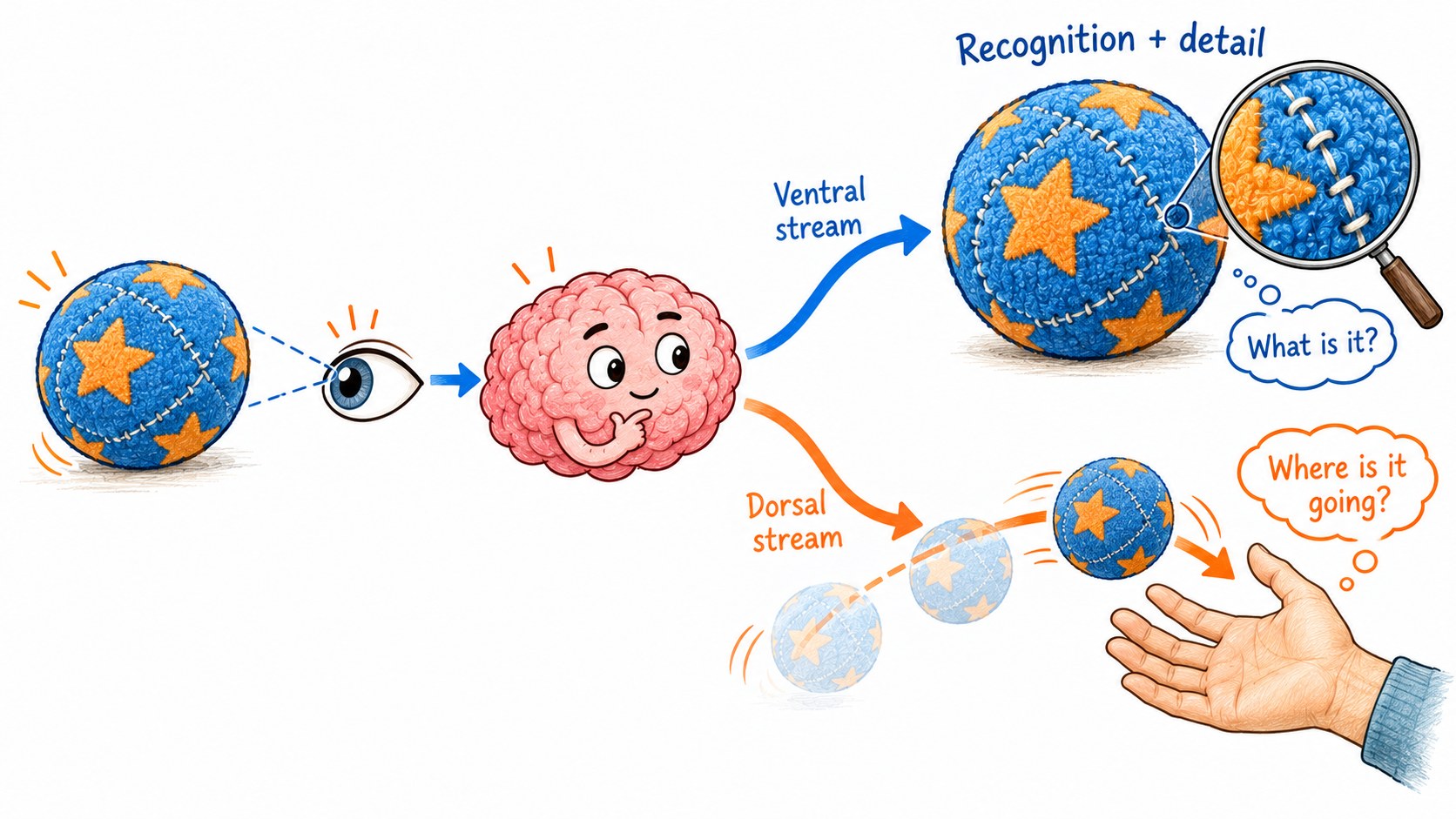}
\caption{\textbf{The same ball is read twice, by the two visual streams in the brain.} The ventral stream resolves its pattern and identity; the dorsal stream resolves its trajectory and guides the catch.}
\label{fig:vision_two_streams_sketch}
\end{figure}

Robotics borrows its vision encoders from computer vision, where they were optimized for object recognition and for image or video generation. An agent acting on an object does not need every detail of it at every instant; it needs where the object is, how it moves, and how to take hold of it. Biology runs two parallel ``vision encoders'' (\cref{fig:vision_two_streams_sketch}): a slow one for fine detail and recognition, and a fast one dedicated to guiding motor behavior. Current encoders resemble the first, and the second has no counterpart. We first survey what current encoders were built for (\cref{subsec:vision_robotics}), then turn to how the brain builds a second visual system for motor behavior (\cref{subsec:vision_biology}), and finally draw the algorithmic inspirations (\cref{subsec:vision_inspirations}).

\subsection{Robotics}
\label{subsec:vision_robotics}

We review the training objective for current vision encoders (\cref{subsubsec:vision_encoder_training_objectives}), how the backbone behind them reshapes their tokens (\cref{subsubsec:vision_backbone_robotics}), and the limits that follow (\cref{subsubsec:vision_information_robotics}).

\subsubsection{The pretraining objective decides what the tokens keep}
\label{subsubsec:vision_encoder_training_objectives}

Almost all tokenizers are ViTs \citep{dosovitskiy2021vit} pretrained at scale and then frozen or lightly fine-tuned; what distinguishes them is the pretraining objective, which leaves a durable fingerprint on the representation. Nearly every recipe asks for one of two things: predict \emph{the image or video itself}, in pixels or in an embedding of it, or align it with \emph{a text description}.

\paragraph{Self-supervised encoders predict the input, in pixel space or in latent space.} The recipes differ mainly in where the prediction target lives. Masked pixel reconstruction puts it in pixel space: the model repaints masked patches directly \citep{he2022mae,xie2022simmim,bao2022beit,tong2022videomae}. Masked latent prediction instead forecasts an exponential-moving-average teacher's \emph{embeddings} of the masked regions \citep{assran2023ijepa}; on video this recovers motion and temporal structure without paying for pixel fidelity \citep{bardes2024vjepa,assran2025vjepa2}. View-invariant self-distillation targets the teacher's embedding of a \emph{different augmented view} of the same image, with centering and sharpening to prevent collapse \citep{caron2021dino,oquab2024dinov2,simeoni2025dinov3}; correspondence, segmentation, and depth emerge in frozen probes.

\paragraph{Caption alignment buys semantics but never asks for geometry.} Image--text contrastive learning predicts which caption belongs to which image, pulling matched pairs together in a joint embedding space \citep{radford2021clip,zhai2023siglip,tschannen2025siglip2}. Because language alignment is the objective itself, these encoders became the vision towers of VLMs and, through them, of most VLAs. The cost is inherited from the captions: they rarely mention exact pose, contact, or metric layout, so nothing in the objective pressures the encoder to represent geometry.

\subsubsection{The VLM or video world-model backbone acts as an extended vision encoder}
\label{subsubsec:vision_backbone_robotics}

The backbone that consumes the tokens decides what the action head finally reads. In a VLA, image tokens enter the language model through a projector; the action head then reads its hidden states or key-value cache \citep{black2024pi0,bjorck2025groot}, or the language model emits the action tokens itself \citep{kim2024openvla}, so vision arrives fused with the instruction and the backbone's semantic knowledge. World Action Models replace the language model with a video-diffusion backbone, so the action head reads features shaped by predicting scene change (\cref{subsubsec:cerebellum_world_model_world_action_models}) \citep{hu2024vpp,dreamzero2026,dyna2_2026}. Smaller imitation policies feed encoder features straight to the action head \citep{chi2023diffusion,zhao2023learning,dinov3dp2025}. In every case, recognition, language, or generation shaped the representation, and none of them is control.

\subsubsection{Current vision stacks are slow, blind to velocity, and miss fine detail}
\label{subsubsec:vision_information_robotics}

\paragraph{Policies act on one stale observation for up to seconds.} Cameras deliver 30--60 frames per second. Autoregressive VLAs typically infer at a few hertz \citep{brohan2023rt2,kim2024openvla}, and even the fastest flow-matching VLAs, which infer in tens of milliseconds \citep{black2024pi0,ma2025realtimevla}, still commit half a second to two seconds of motion to one observation and execute it blind \citep{vlash2025}. Widening the window multiplies the tokens the backbone must process \citep{efficientvla2025survey,streampi2026}, so most VLAs act on a single stale frame \citep{chen2025historytracking,pi2026mem}.

\paragraph{Single frames carry no velocity.} Any event shorter than the hundreds of milliseconds a chunk executes between looks, such as the onset of a slip (\cref{subsec:touch_biology}), falls between frames, and a frame holds no motion to recover it from. Image-pretrained encoders represent task progression poorly even with temporal features added on top \citep{pvrfallshort2025}; \cref{subsubsec:vision_streaming_control_encoder} takes up the video and latent-prediction remedies.

\paragraph{Fine detail is lost.} Thin cables, small screws, and fine texture span a few pixels at the 224--384-pixel inputs these encoders take \citep{dosovitskiy2021vit,zhai2023siglip}. The limit is not the encoder: models that take higher resolution \citep{tschannen2025siglip2,simeoni2025dinov3,bai2025qwen25vl} or reconstruct the image \citep{he2022mae,oquab2024dinov2} keep such detail. The limit is the token budget: tokens grow with the square of the resolution, and a VLA multiplies them by its cameras and its time window, so it runs the encoder small and loses the detail the task needs. 

\subsection{Biology}
\label{subsec:vision_biology}

This section traces how primate vision splits into a stream for recognition and a stream for action, then follows the second.

\subsubsection{The retina compresses before transmitting, and recognition and action get separate streams}
\label{subsubsec:vision_retina_two_streams}

The retina funnels the output of roughly 100 million photoreceptors \citep{curcio1990human} into about one million ganglion-cell axons \citep{curcio1990ganglion}, a hundredfold compression forced at the sensor itself. From there the signal takes two routes. Most of it relays through the lateral geniculate nucleus of the thalamus to primary visual cortex (V1) \citep{kandel2013principles}, and by the time it arrives it is already features, not pixels: V1 neurons are selective for oriented edges, binocular disparity, and, crucially for a moving animal, motion direction and speed. A smaller fraction takes an older, subcortical route straight to the superior colliculus, which drives orienting movements of the eyes and head at very low latency and without awareness \citep{kandel2013principles,wurtz2008stability}.

\begin{figure}[H]
\centering
\includegraphics[width=0.75\textwidth]{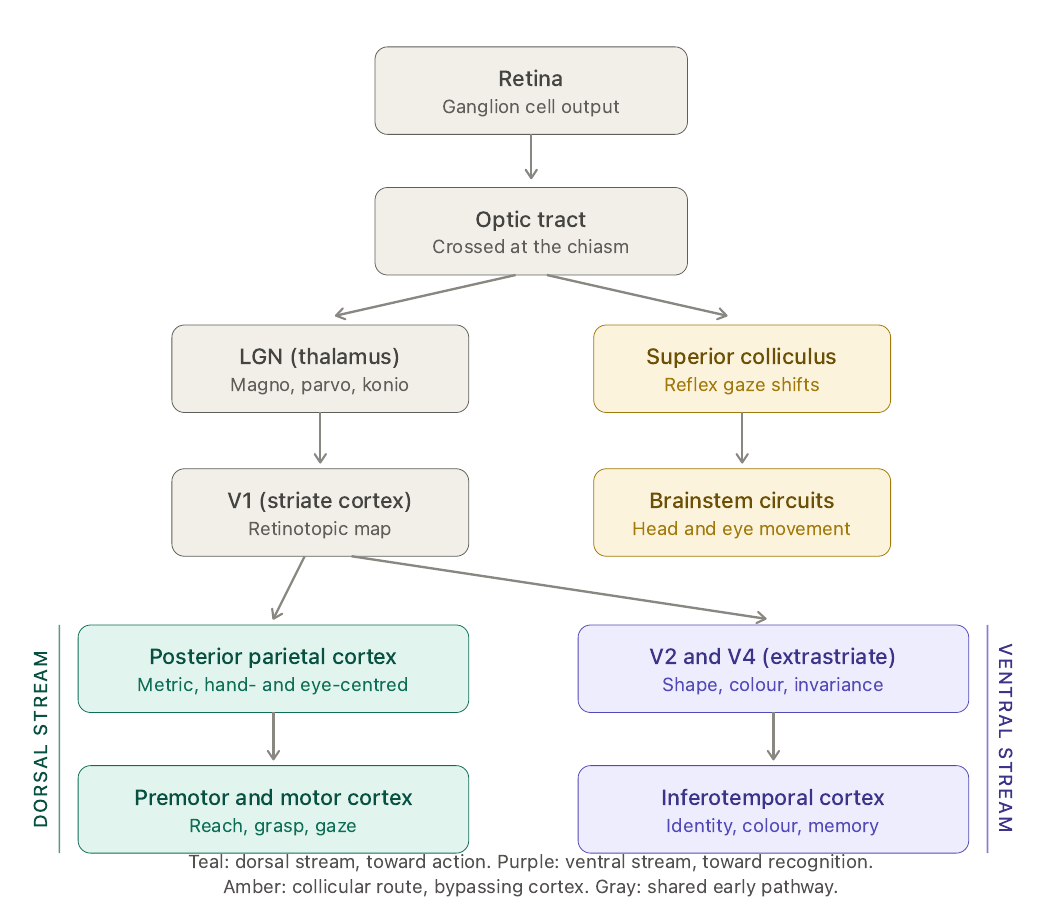}
\caption{Vision for action and vision for recognition separate after V1, and a collicular route bypasses cortex entirely. The geniculate route through V1 divides into a dorsal stream through posterior parietal cortex to premotor and motor areas (vision for action) and a ventral stream through V2 and V4 into inferotemporal cortex (vision for recognition); a separate collicular route drives fast orienting of eyes and head without cortex.}
\label{fig:two_streams}
\end{figure}

\paragraph{Recognition and control get separate streams with different codes.} Beyond V1, primate vision divides into two anatomically distinct streams (\cref{fig:two_streams}) \citep{ungerleider1982two,goodale1992separate,milner2008two}; a third, lateral pathway for social perception has been proposed \citep{pitcher2021third}. The \emph{ventral stream}, into inferotemporal cortex, serves recognition. Its codes are position-invariant and durable: ``a mug, near the laptop'' survives eye movements and can be stored in memory. The \emph{dorsal stream}, into posterior parietal cortex and on to premotor and motor areas, serves control. Its codes are metric and anchored to the hand or eye: ``handle 31\,cm out, 40\degree{} from the wrist'', recomputed continuously and short-lived. Picking up that mug needs both: the ventral stream supplies what it is and which part to take hold of, the dorsal stream where the handle is relative to the hand right now. How cleanly the two streams divide recognition from control is contested \citep{schenk2010independent}, but that debate is not discussed in this paper. The rest of this section follows the dorsal stream.

\subsubsection{The dorsal stream is modeled as a bank of parallel channels, one per class of action}
\label{subsubsec:vision_dorsal_stream_channels}

\begin{figure}[H]
\centering
\includegraphics[width=\textwidth]{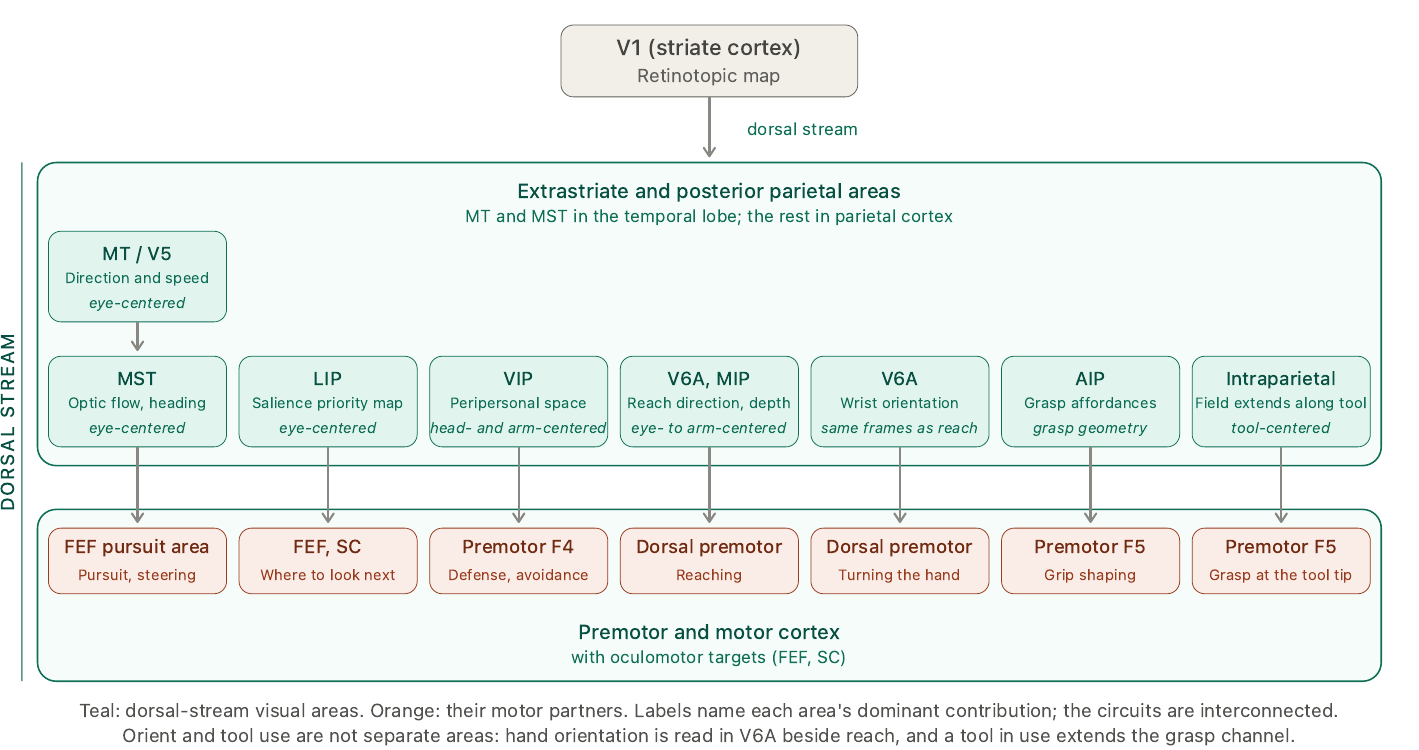}
\caption{In the prevailing model, the dorsal stream is a bank of parallel channels. The figure expands the dorsal branch of \cref{fig:two_streams} and draws the motion areas MT and MST, which that figure folds into the parietal box, as a separate stage. Each visual area estimates one quantity the body needs (where to look, what is moving, what is within reach, where to reach, how to turn and shape the hand, and what a tool in hand reaches) and sends it to a partner in premotor, motor, or oculomotor areas; the italic line in each visual box names the reference frame the channel works in (\cref{subsubsec:vision_coordinate_transforms}). Each label names an area's dominant contribution rather than its only one \citep{rizzolatti2003streams}.}
\label{fig:dorsal_loops}
\end{figure}

In the prevailing model, the dorsal stream is organized by action rather than by feature: each channel pairs a region in parietal cortex that estimates one aspect of the world with a partner in premotor, motor, or oculomotor areas that plans the matching movement (\cref{fig:dorsal_loops}) \citep{colby1999space,rizzolatti2003streams}. The tuning of each area below is well replicated; the division into channels is the model's interpretation of it.

\begin{itemize}
    \item \emph{Motion: what is moving, and how?} The middle temporal area (MT) measures the direction and speed of movement at each point in the image \citep{born2005mt}. Its neighbor MST combines these local measurements into whole-field patterns, the biological counterpart of optic flow: expansion, which signals approach and gives time-to-contact \citep{lee1976tau}; translation, which signals self-motion and gives heading; and rotation \citep{duffy1991mst}.
    \item \emph{Reach: where must the hand go?} Areas V6A and MIP encode the direction and depth of the reach target, and dorsal premotor cortex turns that estimate into an arm trajectory \citep{galletti2003role,batista1999reach}.
    \item \emph{Grasp: how must the hand be shaped?} Area AIP extracts the size, shape, and orientation of the graspable parts of an object, and its premotor partner F5 selects a matching grip type, such as a precision pinch or a whole-hand grasp \citep{murata2000aip,murata1997f5,schaffelhofer2016object}.
    \item \emph{Orient: how must the hand be turned?} Orientation for action appears to be computed separately from perception. Patient D.F., with ventral-stream damage, could not report the orientation of a slot, even by turning a card to match it, yet she posted the card through it, rotating her hand correctly during the reach \citep{goodale1991dissociation}. Neurons in V6A are tuned to wrist orientation during reach-to-grasp, so this computation sits between the reach and grasp channels \citep{fattori2009orientation}.
    \item \emph{Gaze: where to look next?} Area LIP keeps a priority map of the scene, combining what stands out with what the task needs, and the eyes move to its peak \citep{bisley2010priority}.
    \item \emph{Near-body space: what is close to the body?} Area VIP and its premotor partner F4 track objects in the space within arm's reach of the head and body, for avoidance and close-range reaching \citep{fogassi1996f4,graziano1997visuospatial}.
    \item \emph{Tool use: what is in the hand?} A tool in use appears to join the hand's channel: after a monkey uses a rake, parietal neurons that respond to touch on the hand and to objects seen near it extend their receptive fields along the tool \citep{iriki1996tool}. F5 grasp neurons follow the tool tip even with reverse pliers, which close when the hand opens \citep{umilta2008pliers}.
\end{itemize}

The bank of channels is a model built mostly from macaque recordings, lesions, and stimulation \citep{rizzolatti2003streams}, and how separate the channels are is debated. The circuits are densely interconnected, a reach area also codes wrist orientation \citep{fattori2009orientation}, and the dorsal and ventral streams exchange object information rather than working apart \citep{milner2008two,schenk2010independent}. Its transfer to humans is also incomplete: the same parietal maps have expanded more than the rest of visual cortex \citep{meyer2025expansion}, and a left parietal region responding to tool use has no counterpart found so far in monkeys \citep{peeters2009tool,kastner2017toolmaker}.

\subsubsection{Vision selects one object, holds it on the fovea, and tracks its motion}
\label{subsubsec:vision_attention_selection}

The dorsal channels do not process the whole scene. The fovea resolves fine detail over a degree or two \citep{curcio1990human}, so the eye samples the scene in three or four fixations a second, placed by the task \citep{yarbus1967eye,land2009looking}: in manipulation they land where the hand will make contact and move on before it arrives \citep{landhayhoe2001,johansson2001eyehand,hayhoe2017}. Selection precedes the eye movement: attention shifts to the target before the saccade \citep{deubel1996saccade}, and motion neurons in MT and MST respond to the selected object while nearby motion is suppressed \citep{treue1996attention,reynolds2009}, so the channels of \cref{subsubsec:vision_dorsal_stream_channels} report that object alone. Once selected, the object is tracked: smooth pursuit holds it on the fovea, so its motion is read from the eye command rather than from retinal slip \citep{lisberger2010pursuit,landmcleod2000}. 

A few other moving objects are tracked at the same time, but only as unlabeled positions: attention keeps an index on each one as it moves, enough to find it again or to notice that it is about to collide, while what it is remains unresolved until the fovea visits it \citep{pylyshyn1988tracking,meyerhoff2017mot,holcombe2023attending}.

\subsubsection{Each dorsal channel re-expresses visual information toward the frame its effector needs}
\label{subsubsec:vision_coordinate_transforms}

\begin{figure}[H]
\centering
\includegraphics[width=0.7\textwidth]{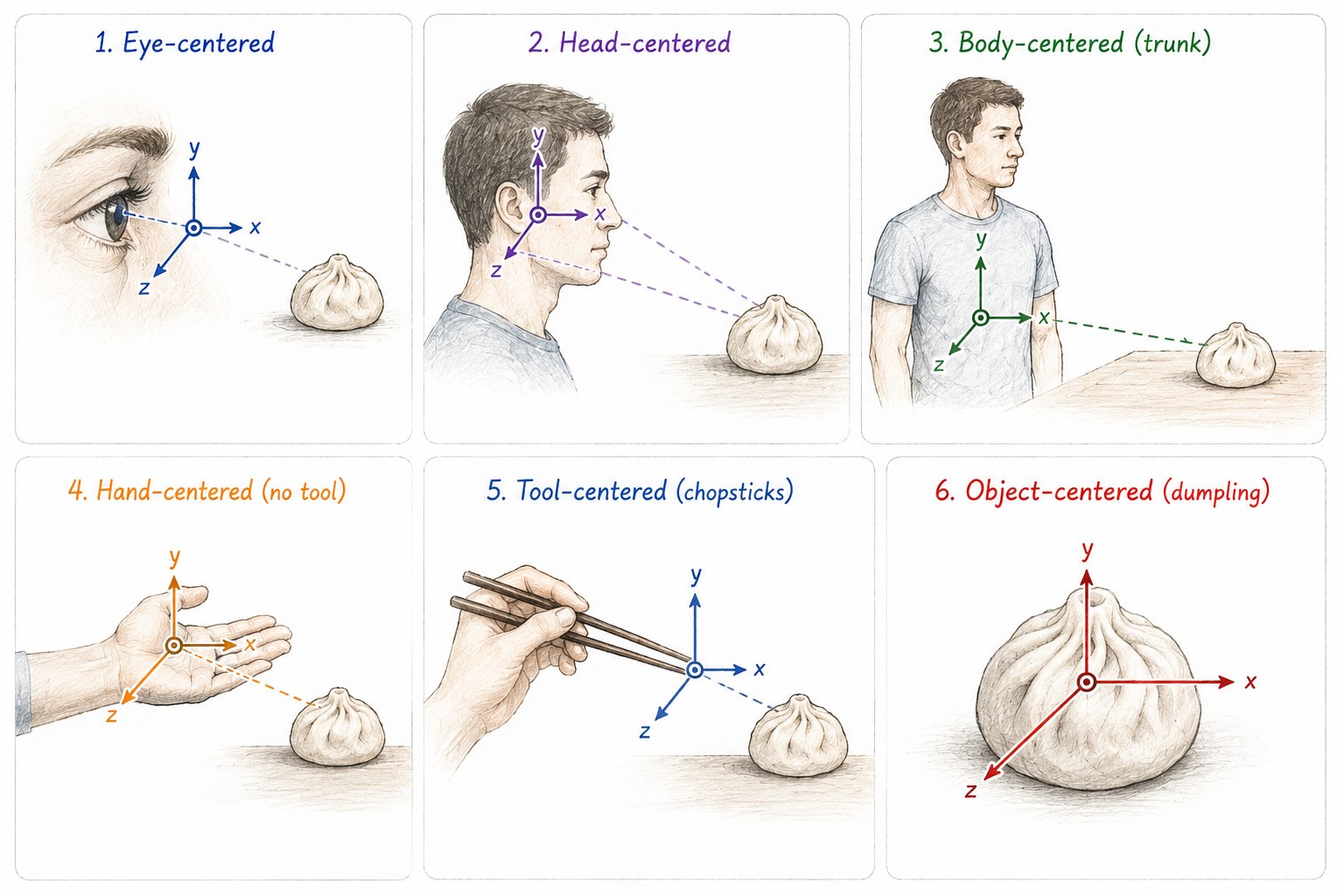}
\caption{A visually guided action must re-express the retinal report in the frame its effector needs. The same dumpling has different coordinates in each frame along the chain: eye, head, body, hand, tool, and object.}
\label{fig:coordinate_frames}
\end{figure}

\paragraph{The frame follows the effector: eye for gaze, head for near space, hand for reach and grasp.} The retina reports everything relative to the eye, but the eye moves in the head and the head on the body, and an action needs the target relative to the body part that will act on it (\cref{fig:coordinate_frames}). This re-referencing underlies every visually guided movement \citep{andersen2002intentional,crawford2011transformations}. Gaze needs no transform (\cref{fig:dorsal_loops}): an eye movement \emph{is} an eye-centered vector, so the motion and gaze channels stay eye-centered \citep{born2005mt,colby1999space}. Near-body space is coded relative to the head and limbs, so an approaching object is located against the face or arm it may hit \citep{duhamel1997headcentered,graziano1994coding}. Reach targets are coded in frames that range from eye-centered to hand-centered across parietal areas \citep{batista1999reach,galletti2003role}; grasp areas code the size, shape, and orientation of the graspable part, an object-centered description \citep{sakata1995parietal,murata2000aip}; and a tool in use extends the hand frame to its tip (\cref{subsubsec:vision_dorsal_stream_channels}). The assignment is flexible: human reach and grasp areas code the same target in a gaze-centered or a body-centered frame depending on whether it is seen or felt \citep{bernier2010flexible,leone2015flexible}, and they carry its direction and distance in several frames at once \citep{caceres2024direction}.

\paragraph{The transform is a population code built from gain fields.} Each channel computes its frame from retinal input using \emph{gain fields} \citep{andersen1985encoding}: a parietal neuron keeps its retinal receptive field, but eye or head position scales its response. Few single neurons encode the target cleanly in the new frame; the population forms a basis from which downstream areas read out any frame they need as a linear combination \citep{pouget1997basis}. The frames in \cref{fig:dorsal_loops} are therefore idealizations: measured receptive fields are commonly mixtures that shift only partway between eye and head coordinates \citep{avillac2005reference}.

\subsubsection{Vision is slow, so the brain acts on a prediction and corrects it later}
\label{subsubsec:vision_relies_on_predictions}

Vision is slow: recognizing what a scene contains takes about 150\,ms \citep{thorpe1996speed}, and a visual correction to a reach begins 100--150\,ms after the target moves \citep{prablanc1992automatic}, long enough for a reaching hand to travel several centimeters. The brain therefore predicts how the image will change and uses the delayed image only to correct the prediction. The image changes for three reasons, ordered here from easiest to predict to hardest.

\paragraph{The eye moved: an exact command.} Each eye jump shifts the whole image. A copy of the eye-movement command reaches the visual areas before the eye moves, and neurons there shift the patch of image they watch to where it will land \citep{duhamel1992updating,sommer2006thalamus}. The scene estimate is updated ahead of the movement, one leading account of why the world appears stable across eye movements \citep{wurtz2008stability}.

\paragraph{The body moved: a command plus limb dynamics.} During a reach, visual feedback about the hand lags on the order of 100--200\,ms, so a controller steering on feedback alone would always correct toward where the hand \emph{was} \citep{desmurget2000forward}. The cerebellum runs a forward model instead: a copy of the arm command predicts the limb's current state, and delayed vision corrects the prediction, the structure of a Kalman filter \citep{wolpert1995internal,miall1993cerebellum,kalman1960new}. The correction runs below awareness: a target displaced during an eye movement is never seen to jump, yet the hand corrects its path \citep{goodale1986large,prablanc1992automatic}.

\paragraph{The world moved: no command, only observation and priors.} A thrown ball's motion must be extrapolated from what is seen. \emph{Perceiving where a moving object is} belongs to the visual pathway. The retina already fires ahead of a moving stimulus \citep{berry1999anticipation}, and a moving object is perceived ahead of a flash at the same position, which one account attributes to continued extrapolation in cortex, although the mechanism remains debated \citep{nijhawan1994motion,turner2024extrapolation}. Learned physics fills the gaps: the cortex holds an internal model of gravity \citep{indovina2005gravity}. \emph{Timing an action on a moving object} brings in the cerebellum, which keeps tracking a target behind an occluder and predicts when it will reappear \citep{cerminara2009target,oreilly2008cerebellum}. Catching needs both: the grip is timed from the gravity model, so firmly that astronauts keep applying it in orbit \citep{mcintyre2001gravity}.

\subsection{Inspiration}
\label{subsec:vision_inspirations}

We make three proposals for robotic vision. First, add a fast encoder built for streaming motor control beside the one inherited from recognition, answering the speed and velocity limits of \cref{subsubsec:vision_information_robotics} (\cref{subsubsec:vision_streaming_control_encoder}). Second, let the model read the metric parameters and the decisions each action unit needs directly off one encoded frame (\cref{subsubsec:vision_direct_readout}). Third, move the camera to see better, which answers the detail limit and requires a policy that acts correctly from any viewpoint (\cref{subsubsec:vision_active_vision}).

\subsubsection{Add a fast streaming encoder for control beside the slow one for recognition}\label{subsubsec:vision_streaming_control_encoder}

\paragraph{Feed a camera-rate encoder straight into the reactive policy head.} One image encoder cannot supply detail and streaming rate at once (\cref{subsubsec:vision_information_robotics}). We should learn from the two visual streams of \cref{subsubsec:vision_retina_two_streams} and give control its own encoder. The recognition encoder stays as it is; its tokens carry identity, layout, and task context for planning and action-unit selection. The control encoder runs at camera rate and feeds a few short-lived motion tokens per frame straight into the reactive policy head, which corrects the action without waiting for a new detail frame. Both streams enter one model, which learns when a decision needs detail and when the latest motion suffices.

\paragraph{Distill the fast encoder from video models, then train it on control losses.} Much of the needed dynamics already exists in video foundation models \citep{assran2025vjepa2,agarwal2025cosmos}, whose pixel-generating variants spend their capacity on appearance a controller never uses \citep{tong2022videomae}, while latent prediction does not \citep{jepavla2026}. Policies that use optical flow \citep{xu2024im2flow2act} or point tracks \citep{wen2024atm,bharadhwaj2024track2act,chen2025historytracking} show that explicit motion helps manipulation. The fast encoder can be distilled from these sources and then trained with the policy on control losses, predicting the next action and the metric quantities of \cref{subsubsec:vision_direct_readout}, with no reconstruction term.

\subsubsection{Read the numbers and the decisions an action unit needs off one encoded frame}\label{subsubsec:vision_direct_readout}

\paragraph{Encourage the model to emit more of each unit's parameters directly from vision.} Code-as-policy systems send Cartesian pose targets, each a set of numbers read from the scene: a grasp point and pose, a slot angle, an insertion depth \citep{liang2023code,he2026astrabody,si2026pyrua}. Today those numbers come from code the model writes at run time, thresholding, segmenting, or line fitting \citep{kumar2026aor,fu2026capx}, or from a separate grasp or keypoint detector \citep{fang2023anygrasp,huang2024rekep}. The path works, but each step costs a round of generation and execution, and the thresholds break on the next object. We propose a complementary path, the direct readout of \cref{subsubsec:vision_dorsal_stream_channels}: given the unit, the model selects the features that matter, the slot's axis for orient or the graspable part for grasp, and emits their numbers in the body frame, as pointing and spatial VLMs do in 2D \citep{yuan2024robopoint,nasiriany2024pivot,chen2024spatialvlm} and spatially trained VLAs begin to in 3D \citep{qu2025spatialvla}. The control encoder of \cref{subsubsec:vision_streaming_control_encoder} should carry the same kind of numbers at camera rate: optical flow, time-to-contact, and the approach of the hand to a surface, the quantities of \cref{subsubsec:vision_dorsal_stream_channels}, so that the reactive head reads a metric state instead of inferring it from pixels. The code path supplies the training data, as \cref{subsubsec:planner_commander_consolidation_agent} describes, so measurement migrates into weights while the code remains as the test.

\paragraph{Answer the unit's decisions with parallel typed experts over one encoded frame.} A unit also needs decisions from the image, such as whether the handle is free. We propose typed experts inside the model, routed as in a mixture of experts \citep{fedus2022switch}: the frame is encoded once, each question goes to the expert of its type, and the experts run in parallel, the channels of \cref{subsubsec:vision_dorsal_stream_channels}. Each returns a probability, a choice, or a score that downstream code reads directly (\cref{subsubsec:planner_commander_planner_units,subsubsec:planner_commander_directional_judgment}), about ten times faster than generating text \citep{openai2026decisions}.

\subsubsection{Move the camera to see better, and train across viewpoints}\label{subsubsec:vision_active_vision}

\begin{figure}[H]
\centering
\includegraphics[width=0.68\textwidth]{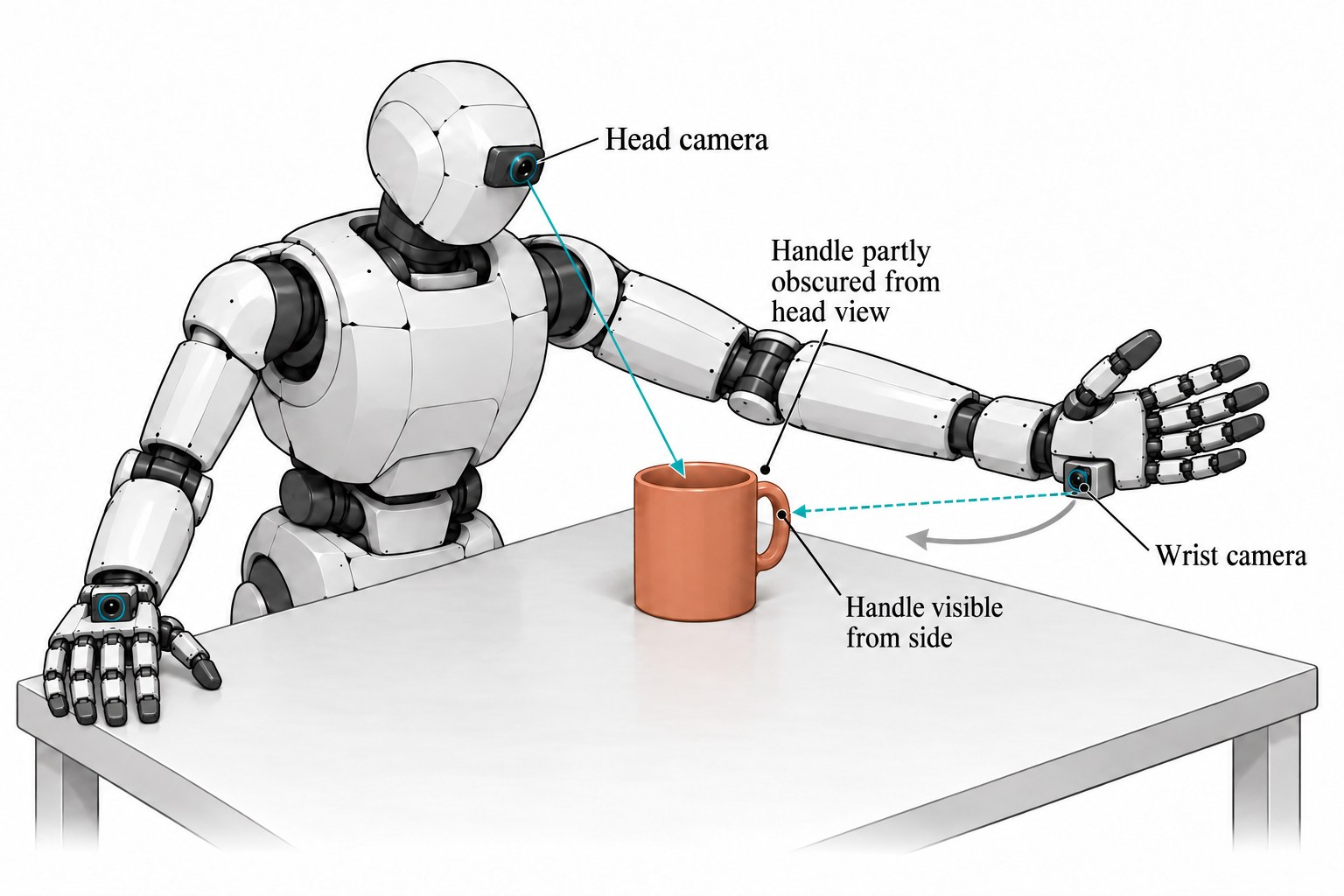}
\caption{A second viewpoint reveals what the head camera cannot see. The mug body partly hides its far-side handle from the head camera. Moving the wrist camera of the free hand to a side viewpoint reveals more of the handle before grasping. Teal arrows mark viewing directions; the curved gray arrow marks the wrist motion.}
\label{fig:multi_view}
\end{figure}

\paragraph{Move the camera to see better.} We propose that the policy treat its wrist or head camera as an effector as well as a sensor and learn when to move it \citep{bajcsy1988,aloimonos1988,chuang2024avaloha}, as the eye is moved in \cref{subsubsec:vision_attention_selection}. Moving closer puts more pixels on the target, and a new angle removes an occlusion that no processing could see through \citep{bajcsy2018}: when a grasp is ambiguous, the policy should choose a pose that reveals the handle or the occluded side before committing the arm (\cref{fig:multi_view}). Ambiguity has a measure: a grasp question of \cref{subsubsec:vision_direct_readout} answered with a probability near one half calls for a new view before any further reasoning. Moving closer also answers the detail limit (\cref{subsubsec:vision_information_robotics}) with no change to the encoder.

\paragraph{Give the model its camera pose, and train it to use it.} We propose that every visual input arrive with its intrinsics and extrinsics, and that the model's outputs be expressed in the body frame (\cref{subsubsec:vision_coordinate_transforms}). For the code path, the model should read the points the code needs as 3D coordinates in the body frame, using the calibration in its context; the code then runs unchanged from any view, and points gathered from several views fall into one scene (\cref{subsubsec:vision_direct_readout}). For direct action, the policy should condition on the same calibration \citep{jiang2026camera}, so that a new pose is a new measurement of the same scene rather than a new task; views registered in the body frame become one token set the model attends over, which is the fusion. We propose recording each episode from several calibrated cameras \citep{xie2026multiview}, moving the wrist camera during episodes, and including tasks that no single view can solve, so that the model is forced to combine views.

\clearpage
\section{Is the cerebellum the brain's world model?}
\label{sec:cerebellum_world_model}

\begin{figure}[H]
\centering
\includegraphics[width=0.7\textwidth]{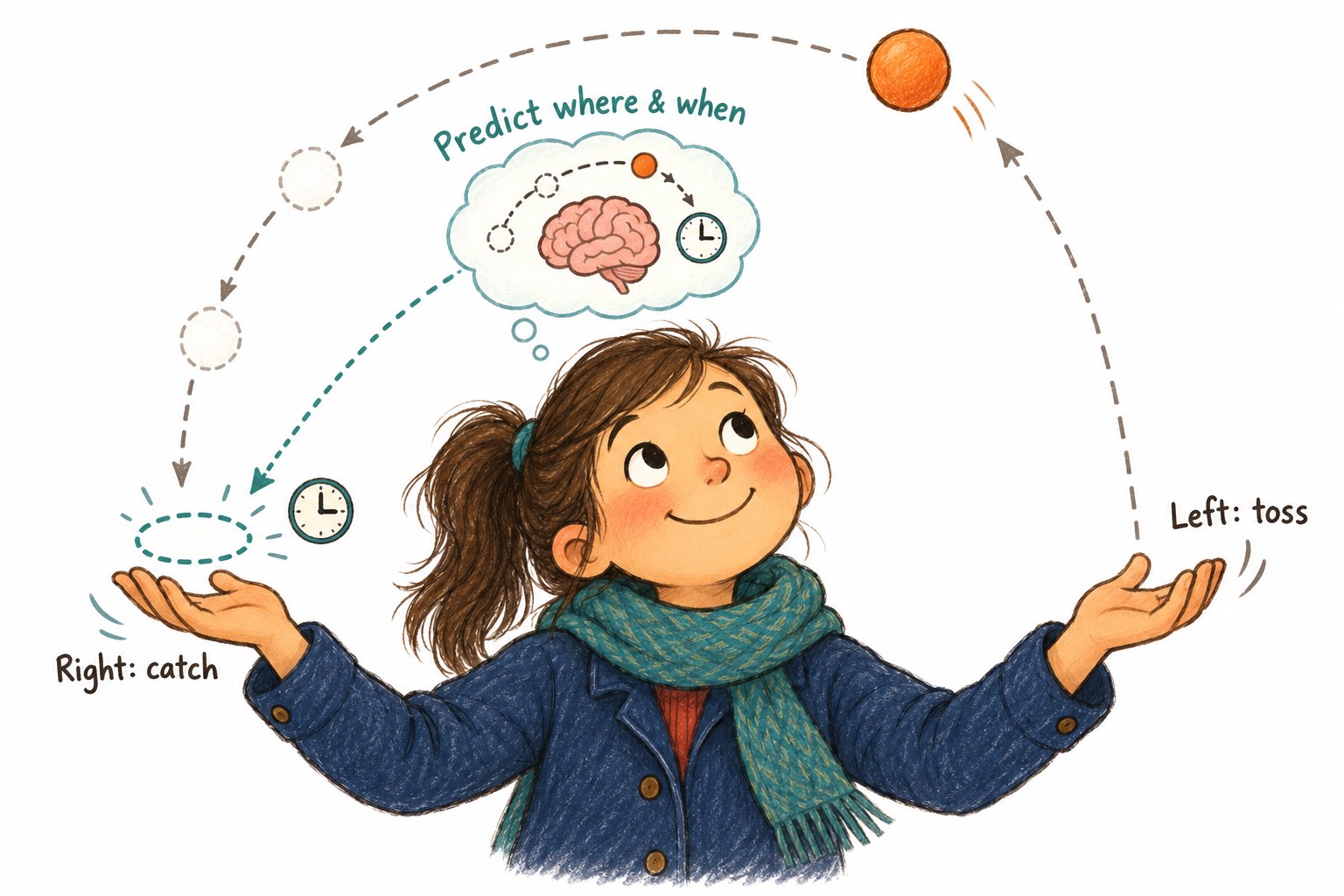}
\caption{\textbf{Having tossed the ball herself, she knows from the throw where and when it will land.} A copy of the tossing command is enough to forecast the ball's arrival, so the right hand is already in place while the ball is still in the air. The faint circles show predicted positions of the same ball.}
\label{fig:predict_action_consequence}
\end{figure}

``World model'' is robotics' name for a model that predicts how the future unfolds, most visibly as generated video. The brain has its predictor too: the cerebellum takes a copy of each outgoing motor command and predicts, in a compact sensory code, what its owner will feel and see (\cref{fig:predict_action_consequence}) \citep{wolpert1995internal,miall1993cerebellum,kawato1999internal}. Without it a person still acts, but coordination degrades badly \citep{bastian2006learning}. With it, the prediction arrives within tens of milliseconds and recalibrates itself from its own error. The chapter compares the predictor in robotics (\cref{subsec:cerebellum_world_model_robotics}) with the one in the brain (\cref{subsec:cerebellum_world_model_biology}), and proposes how a predictor beside the policy could let a robot adapt as quickly to its world (\cref{subsec:cerebellum_world_model_inspiration}).

\subsection{Robotics}
\label{subsec:cerebellum_world_model_robotics}

\subsubsection{World Action Models train the policy under a world-prediction loss}
\label{subsubsec:cerebellum_world_model_world_action_models}
World models learn environment dynamics at scale from internet video \citep{bruce2024genie,agarwal2025cosmos,assran2025vjepa2}. Robotics has traditionally trained the world model apart from the policy \citep{ha2018worldmodels,hafner2023dreamerv3}: the policy was a separate small network reading the world model's latent, so the prediction loss never shaped the policy's own weights. World Action Models (WAMs) instead train one model on the joint distribution over future states and actions, with the action as one output \citep{wang2026wam}. The policy then inherits the prediction objective as a dense signal on the shared backbone, which encourages it to encode task-relevant temporal structure \citep{lightwam2026}.

\begin{figure}[H]
\centering
\includegraphics[width=0.85\textwidth]{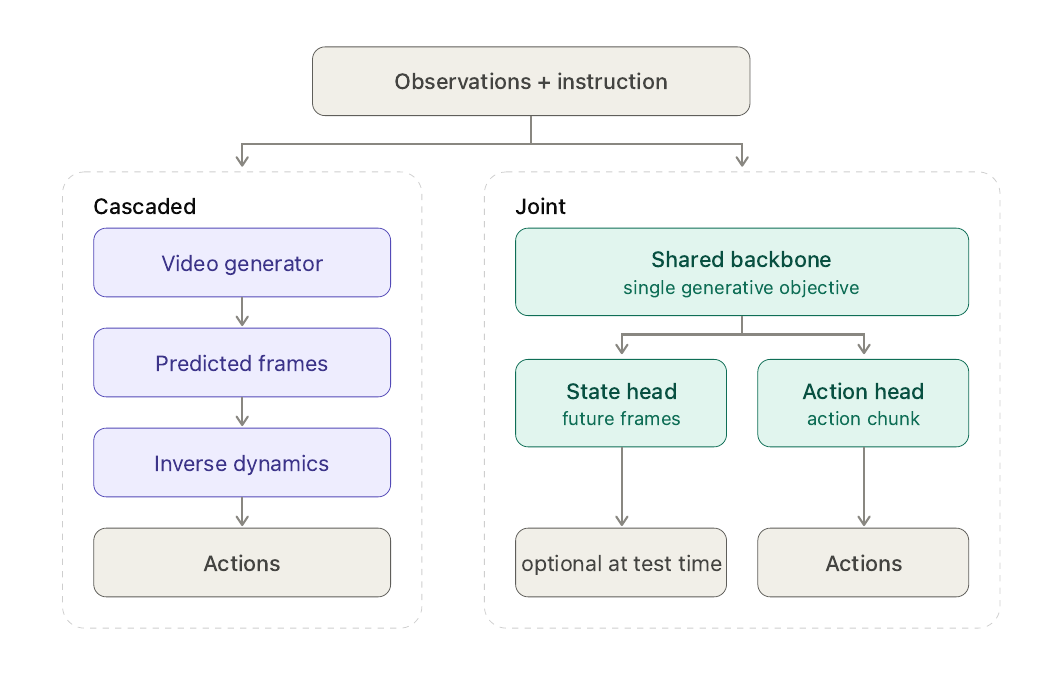}
\caption{Joint World Action Models let the prediction loss train the action head, while cascaded ones only pass predicted frames forward. In the cascaded design a video generator synthesizes predicted frames and a separate inverse-dynamics module reads actions off them. In the joint design a shared backbone trained under one generative objective feeds a state head that generates future frames, which some variants skip at test time, and an action head that emits the executed chunk.}
\label{fig:wam_architecture}
\end{figure}

\paragraph{How prediction is coupled to action decides what reaches the policy.} In the \emph{cascaded} family (\cref{fig:wam_architecture}), a generative model imagines how the scene will evolve and an inverse-dynamics model reads off the actions between predicted frames \citep{du2023unipi,wang2026wam}. The generator needs no action labels, but rollout errors reach the actions uncorrected and action quality never trains the generator. The field has therefore moved to \emph{joint} designs, where one network outputs both future frames and actions \citep{wang2026wam}. The discrete route predicts image, text, and action tokens in one autoregressive sequence, so an early action error propagates \citep{cen2025worldvla}. The continuous route denoises video and actions together, from one joint latent with decoupled heads \citep{li2025uva} or with a separate noise level per modality \citep{zhu2025uwm}, so the same network trains on video without action labels; latent actions inferred from optical flow \citep{bi2025motus} and co-training on action-free video, human hands included \citep{dreamzero2026}, add supervision from any embodiment.

\paragraph{The video branch that supplies the prior is also the source of the open problems.} Latency comes from imagining before acting rather than from model size. Backbones span 0.4--14B parameters \citep{lightwam2026,dreamzero2026}, overlapping mainstream VLAs, yet WAMs that imagine before acting take 0.3--1.2\,s per chunk, and up to 5\,s for the largest video model, against 60--200\,ms for VLAs; skipping generation at test time brings Fast-WAM to 190\,ms, on par with mid-size VLAs \citep{wamrobustness2026,fastwam2026}. Evaluation is unsettled: emerging benchmarks score visual fidelity, physical commonsense, and action plausibility \citep{wang2026wam}, but across the leading video generators physical understanding is poor and uncorrelated with visual realism, so a model can generate well while predicting dynamics that would fail on a robot \citep{motamed2026physics}. The same branch also bounds what a WAM can predict: existing WAMs predict almost exclusively in the visual modality, so contact-rich settings where the relevant state is occluded or only tactile remain underexplored \citep{vtwam2026}.

\subsubsection{Test-time adaptation stores recent experience in a latent, in temporary weights, or in the context}
\label{subsubsec:cerebellum_world_model_adaptation_today}

Test-time adaptation changes how a policy behaves during deployment without a new training run, because the robot's condition, the objects it handles, and the environment all drift. Three stores are in use, none specific to world models, and they differ in capacity and in cost.

\paragraph{A latent compresses recent experience into one vector.} A network summarizes recent experience into a vector that conditions the policy: Rapid Motor Adaptation regresses an ``extrinsics'' latent from proprioceptive history in one forward pass \citep{kumar2021rma,son2026tam,astrom1995adaptive}. The latent is instantaneous and sample-efficient but bounded by what it can represent.

\paragraph{Temporary weights hold more than a latent can.} Parameters are written by gradient descent during deployment and discarded afterwards. RoboTTT updates test-time-training layers inside the policy on 8{,}000 timesteps of history \citep{jiang2026robottt,sun2024ttt}; AdaJEPA takes one self-supervised step on a world model's final layers per replanning cycle \citep{wang2026adajepa}; test-time training of the VLA follows the same pattern \citep{ji2026vane,liu2026ttrl}. Updates stay few and small, and nothing is written back, which separates this from continual learning.

\paragraph{The context window holds the raw history.} The raw history itself conditions the policy: a sequence model reads recent observations, actions, and outcomes and changes its behavior within the episode. People can supply that history as a few demonstrations \citep{fu2024icrt,vosylius2025instant}, or the robot can gather it, so that a legged robot walks better after its early falls \citep{liu2025locoformer,laskin2022algorithmdistillation}. It is the most general store, but attention cost grows with the window and the experience is lost once it scrolls out.

\subsection{Biology}
\label{subsec:cerebellum_world_model_biology}

A WAM needs hundreds of milliseconds to seconds per prediction (\cref{subsubsec:cerebellum_world_model_world_action_models}); the cerebellum answers in tens of milliseconds. From the state of the body and the \emph{efference copy} of the outgoing command, it predicts what that command will produce: grip force tracks the load a lifting arm creates, and patients with cerebellar atrophy show impaired coupling and excess grip force \citep{nowak2002grip}. The cerebellum holds about four fifths of the brain's neurons \citep{herculanohouzel2009human}. What that investment buys is not settled; the forward-model account reads it as prediction \citep{wolpert1998internal,ito2008control}, and this survey adopts that reading.

\subsubsection{A three-synapse, very wide circuit predicts before the senses report}
\label{subsubsec:cerebellum_world_model_wide_shallow_predictor}

\begin{figure}[H]
\centering
\includegraphics[width=0.95\textwidth]{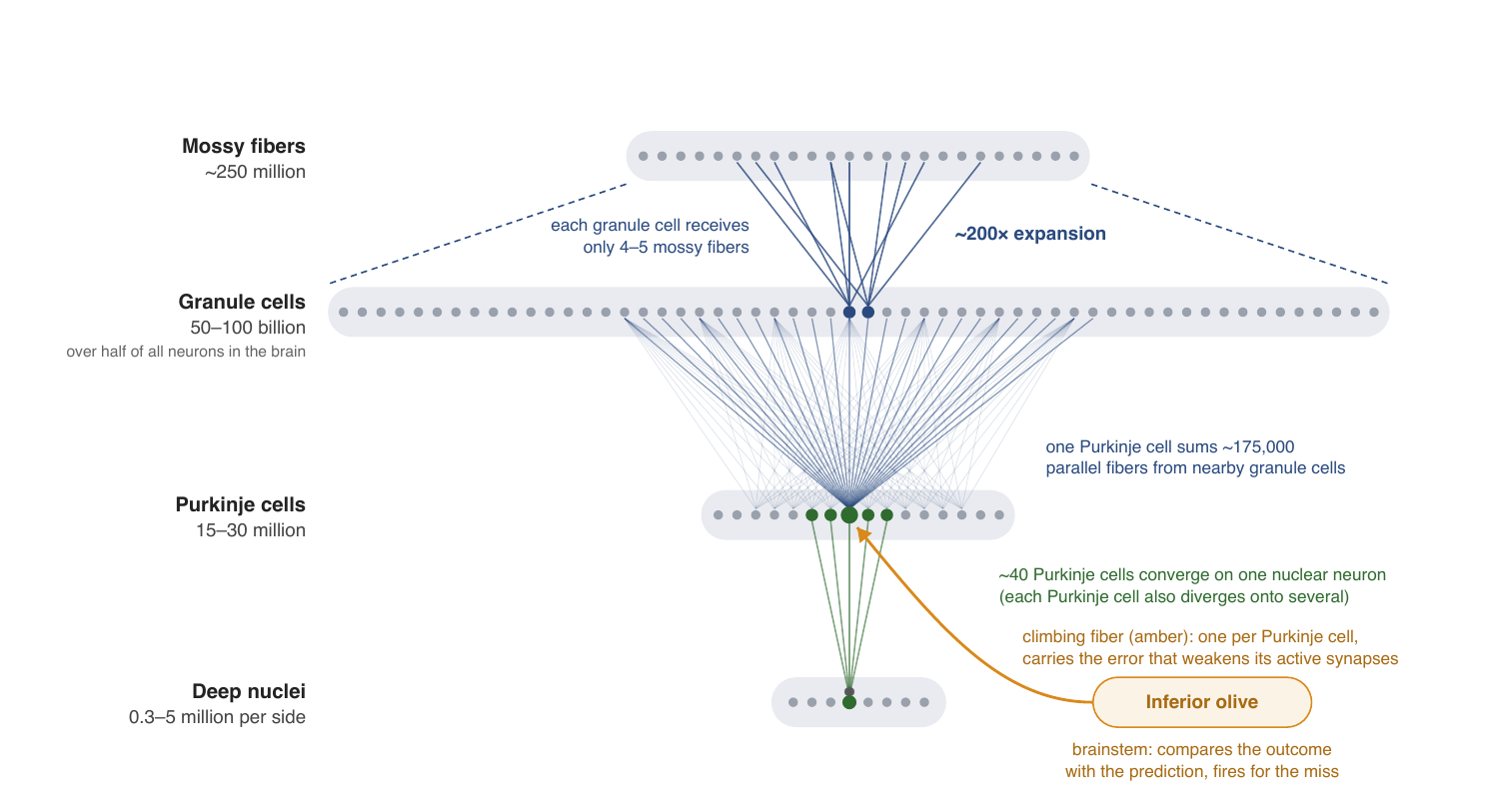}
\caption{The cerebellar circuit is three synapses deep, wide in the middle, and local along its length. The highlighted paths trace one local circuit: parallel fibers run only a few millimeters \citep{pichitpornchai1994parallel}, and about forty Purkinje cells converge on one nuclear neuron \citep{person2012purkinje}, so coupling between body parts is local by construction. The inferior olive (amber) returns the prediction error over one climbing fiber per Purkinje cell, which adjusts the parallel-fiber synapses that were active when the faulty prediction was made. Counts are rough human figures \citep{andersen1992quantitative,napper1988parallel,kawato2021fifty}.}
\label{fig:cerebellum_expansion}
\end{figure}

The cerebellar circuit is a feedforward network only three synapses deep, narrow at both ends and enormously wide in the middle (\cref{fig:cerebellum_expansion}). Input arrives over the \emph{mossy fibers}, which carry the efference copy and the current state of the body. The middle layer expands this input by about two hundred granule cells per mossy fiber \citep{albus1971theory,kawato2021fifty}, into 50--100 billion \emph{granule cells}, more than half of all neurons in the brain \citep{andersen1992quantitative,herculanohouzel2009human}, each sampling only 4--5 mossy fibers, so similar inputs are spread onto nearly disjoint sets of cells \citep{marr1969theory,albus1971theory}. The output layer contracts the code again: granule axons, the \emph{parallel fibers}, cross the dendrites of 15--30 million \emph{Purkinje cells} \citep{nairn1989purkinje,andersen1992quantitative}, each of which sums ${\sim}2 \times 10^{5}$ of them near-linearly \citep{napper1988parallel,walter2006linear}. On the forward-model account, the Purkinje population discharge is the prediction: the anticipated sensory consequence of the command just issued. Width costs nothing in time; only depth does. A whisker touch in mice changes Purkinje output within 6--8\,ms \citep{brown2024simplespike}, and the nuclear output drives motor cortex through a fast loop \citep{gao2018cortico}, well inside the 50--100\,ms that sensory feedback needs to alter a command through cortex \citep{pruszynski2012optimal}.

\subsubsection{Error signals recalibrate the cerebellum locally and fast}
\label{subsubsec:cerebellum_world_model_error_signal_readout_training}

\paragraph{The climbing fiber delivers the error to the synapses that made the prediction.} The brainstem reports the outcome of each command to the \emph{inferior olive} within tens of milliseconds of the outcome. In the modern reading of the Marr--Albus--Ito theory, inhibition from the nuclei subtracts the prediction, so the olive fires mainly for the miss \citep{medina2002,loyola2023olive}. The miss returns over the \emph{climbing fibers} (\cref{fig:cerebellum_expansion}), one per Purkinje cell, each wrapping its target's dendrites and firing it as one complex spike \citep{ohmae2015,silva2024climbing}. The spike drives one well-documented learning rule: the parallel-fiber synapses that produced the faulty prediction, those active one loop delay earlier (a delay that differs by region and is about 120\,ms in the flocculus \citep{suvrathan2016timing}), are weakened in proportion to the miss \citep{marr1969theory,albus1971theory,ito2001ltd}, or strengthened, depending on the microzone \citep{dezeeuw2021bidirectional}, so the corrected prediction appears when the same input recurs.

\paragraph{Wiring gives the error its direction, and the expansion keeps the correction local.} Purkinje cells are grouped into \emph{microzones}, strips of about a thousand cells that share one target in the \emph{deep nuclei} and so form one output channel \citep{apps2005foundations}. In the best-studied zones, each microzone moves the body along one axis and its climbing fibers report error along that same axis, so a proportional change reduces it; the alignment is laid down largely by developmental wiring \citep{apps2005foundations,sotelo2004olivocerebellar}. The granule-layer expansion of \cref{subsubsec:cerebellum_world_model_wide_shallow_predictor} then keeps the correction local, whether through sparse activity, as few as 1--2\% of cells active in any millisecond in the classical theories \citep{marr1969theory,kawato2021fifty}, or through expansion and decorrelation alone: the synapses active during one prediction overlap little with those of another, so adjusting them leaves most others intact.

\paragraph{One error produces a brief correction; repeated errors consolidate over days to weeks.} A single pairing depresses Purkinje firing on the next trial; repeated pairings stabilize it, and the trace shifts to the nuclei over about a week \citep{yang2014purkinje,shutoh2006memory,dezeeuw2021bidirectional}. Magnifying spectacles leave the vestibulo-ocular reflex short of the image, and retinal slip raises its gain within hours \citep{ito1982cerebellar}; reversing prisms invert its direction only over weeks \citep{gonshor1976prisms}. How cortex uses the recalibrated forecast is described in \cref{subsubsec:planner_commander_cerebellar_gate}.

\subsubsection{The prediction amends the action in flight or returns to cortex}
\label{subsubsec:cerebellum_world_model_ascending_output_cortex_routes}

\begin{figure}[H]
\centering
\includegraphics[width=0.95\textwidth]{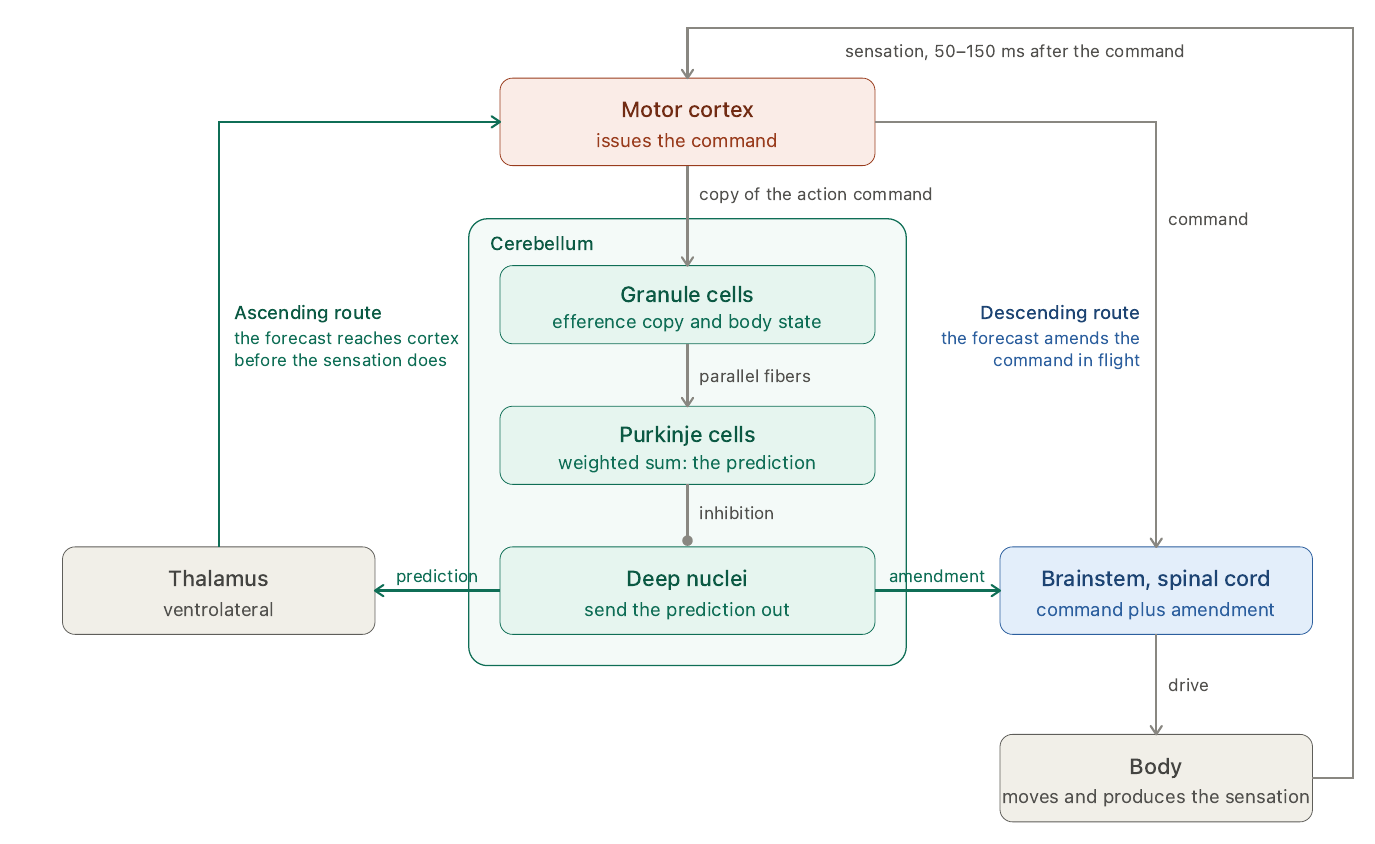}
\caption{The same nuclear prediction leaves by two routes. Ascending (left), it reaches cortex through the ventrolateral thalamus. Descending (right), it joins the command in flight at the brainstem and spinal cord. Both arrive within tens of milliseconds of the command, before sensory feedback could alter it through cortex.}
\label{fig:cerebellum_inference}
\end{figure}

The nuclear output prediction leaves by two routes (\cref{fig:cerebellum_inference}): down to the brainstem, to amend the command in flight, and up to the cortex that issued it.

\paragraph{Descending, the prediction amends the action in flight.} The nuclei project directly to the brainstem motor centers and the spinal cord \citep{fujita2020fastigial,sathyamurthy2020cerebellospinal}, so the prediction can alter a movement without passing through cortex. In a reach, nuclear activity scales with how hard the arm must brake, and driving it changes the reach in flight \citep{becker2019reach,low2018interposed}. The fingers are a partial exception: their muscles depend heavily on direct cortical projections, so much of the correction reaches them by the slower route through cortex \citep{lemon2008descending,riddle2009reticulospinal}.

\paragraph{Ascending, the prediction returns to the cortex that issued the command.} Through the ventrolateral thalamus the nuclei reach the motor and premotor areas in closed loops \citep{kelly2003loops,bostan2018basal}. This input drives motor cortex firing during a reach \citep{sauerbrei2020input}, and silencing the nuclei collapses preparatory activity \citep{gao2018cortico,chabrol2019preparatory}, which suggests cortex plans on where the limb will be. Somatosensory cortex attenuates the predicted touch, so self-touch feels weaker \citep{blakemore1998tickle,kilteni2020cerebellar}.

\paragraph{The cerebellum also sets timing and reaches the reward system.} A fixed-latency nuclear signal triggers a movement and sets its timing \citep{dacre2021initiation,kunimatsu2018timing}. The cerebellum is also wired into the reward system: its nuclei drive dopamine neurons \citep{carta2019cerebellar}, and its own circuit carries the expectation of reward alongside the sensory forecast \citep{wagner2017reward,kostadinov2019predictive}. A prediction therefore comes with a value attached, and the forecast feeds the system that scores practice (\cref{subsubsec:planner_commander_practice_habit}).

\subsection{Inspiration}
\label{subsec:cerebellum_world_model_inspiration}

The cerebellum predicts ahead of the senses and recalibrates from its own error. Our proposals follow in decreasing confidence. Run a self-labeling predictor beside the policy and let the policy consult it as a tool; once the predictor answers before the sensory report arrives, let the policy steer by the forecast in the loop. Make the predictor shallow, wide, and sparse, so that it is fast and an error-scaled step within the episode corrects only the forecast that was wrong. Tag every joint and sensor with a body-part embedding so the sensorimotor map transfers across embodiments.

\subsubsection{Consult the predictor as a tool when it is slow}
\label{subsubsec:cerebellum_world_model_predictor_beside_commander}

We propose a predictor that runs beside the policy, receives the state and a copy of each command, and forecasts what the sensors will report, so that the processing delay is covered. It is the easiest module to train: every executed action is labeled by its own outcome a moment later. Even when too slow for the control loop, as current WAMs are (\cref{subsubsec:cerebellum_world_model_world_action_models}), it makes the search for a correction cheap (\cref{subsubsec:planner_commander_directional_judgment}): a candidate correction is scored as a forecast before the robot moves \citep{pilworld2026,hansen2024tdmpc2}. If it predicts that the object will not lift under the planned grip force, the model raises the force first, and the gap between forecast and outcome calibrates its model of contact.

\subsubsection{Run the predictor in the loop when it is fast}
\label{subsubsec:cerebellum_world_model_predictor_in_loop}

A predictor fast enough to return before the real observation arrives changes what the model acts on: it steers by the forecast, so the delay between perception and inference drops out of every correction \citep{vlash2025,abadia2021cerebellar}. The main benefit is that prediction and policy stay separate. A wrong outcome scores each against its own target, so the world's dynamics are learned by the predictor alone and the policy's weights stay intact \citep{mccloskey1989catastrophic,kirkpatrick2017ewc}. This suits a world that changes faster than a strategy should: the way the task is handled stays fixed while the model of the physics underneath it adapts within the episode. No model today infers fast enough for this; it is a direction for the future, open only to a predictor built for speed (\cref{subsubsec:cerebellum_world_model_sparsity}).

\subsubsection{Make the predictor shallow, wide, and sparse, so it is fast and adapts at test time}
\label{subsubsec:cerebellum_world_model_sparsity}

We consider a predictor that is shallow, wide, and sparse. Shallow buys speed, since latency grows with depth while width runs in parallel; current world models are deep, and depth sets their latency. Wide keeps capacity: expanding the input into a code many times wider makes nonlinear relations linear, so a simple readout suffices \citep{cover1965geometrical,rahimi2007random}, and the expansion decorrelates overlapping inputs \citep{caycogajic2019pattern}, as the cerebellar granule layer is thought to do \citep{marr1969theory,albus1971theory}. Sparse makes test-time adaptation safe: in a dense model a gradient step moves shared features and shifts every forecast built on them \citep{mccloskey1989catastrophic}, which AdaJEPA mitigates by updating only the final layers \citep{wang2026adajepa}.

Sparsity is where a mixture of experts enters, in the delegated sense of \cref{subsubsec:planner_commander_end_to_end} rather than the load-balancing sense of today's models, whose experts rarely map onto nameable functions \citep{shazeer2017,fedus2022switch,mai2026meat}. Groups of units are delegated to particular functions, a payload regime or a body part, so that overlapping inputs select nearly disjoint groups. Test-time adaptation then touches only the units behind the wrong forecast: one step scaled by the error corrects the current regime and leaves every forecast with a disjoint support unchanged, so a heavier payload never disturbs an earlier calibration. The exact design, how deep, how wide, and how functions are assigned to units, is to be determined.

\subsubsection{Tag every joint and sensor with a body-part embedding so the map transfers}
\label{subsubsec:cerebellum_world_model_predictor_learning}

We should replace the arbitrary joint index in VLA, world model, and multimodal large language model (MLLM) backbones with a \emph{body-part embedding}. Cross-embodiment training today flattens each robot's state and action into one vector, padded to a common length \citep{oneill2024openx,octo2024,doshi2024crossformer}, so an element's index is all that identifies it, and the index says nothing about which body part it belongs to or which element of another robot it corresponds to. Instead, each joint, actuator, and sensory channel should carry a learned vector named by its role, thumb, index, wrist, and the like, so every token has semantic as well as positional meaning. This structure need not be learned from data: the robot's kinematics fix which actuator moves which sensor, as the cerebellum's developmental wiring does (\cref{subsubsec:cerebellum_world_model_error_signal_readout_training}).

Two gains follow. For cross-embodiment transfer, human and robot hands trained under shared finger names can pool their data; we expect grip-force statistics for a thumb to carry across morphologies, as morphology-aware tokens already transfer across bodies \citep{sferrazza2024bodytransformer,suzuki2026morphology}. For adaptation within one embodiment, the forecast error routes by body part: when an object is wider than expected, the error lands on the thumb and middle-finger channels and corrects only the units delegated to them (\cref{subsubsec:cerebellum_world_model_sparsity}). Per-embodiment stems approximate this without naming joints \citep{wang2024hpt,przystupa2025morphology}; the named tag makes the structure explicit inside one model.

\clearpage
\section{Planning and command: how the brain divides what a VLA does in one network}
\label{sec:planner_commander}

\begin{figure}[H]
\centering
\includegraphics[width=0.65\textwidth]{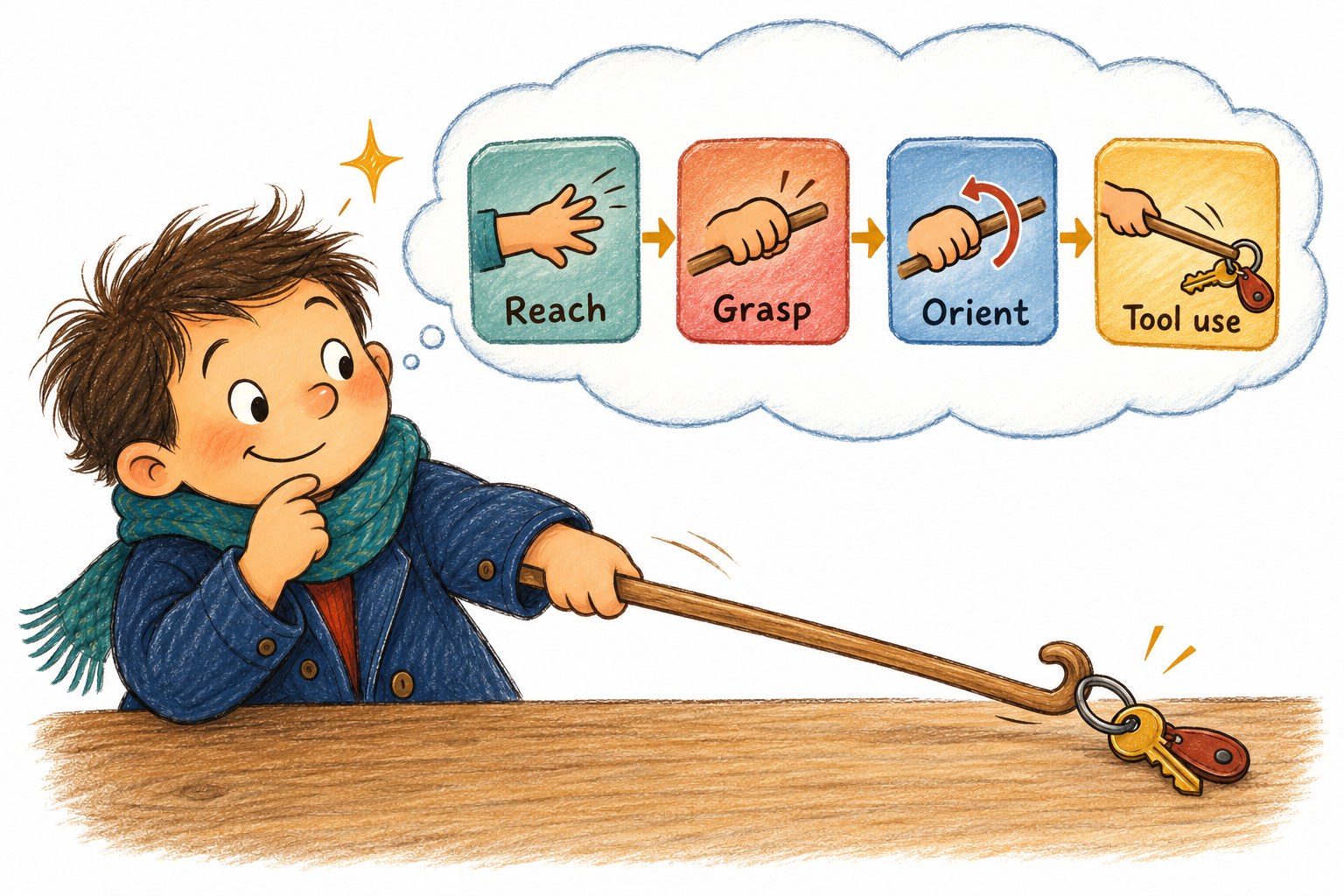}
\caption{\textbf{One model composes familiar primitives into a new solution.} Reach, grasp, orient, and tool use, none of them learned for this task, combine to retrieve an out-of-reach key, and the same model turns the plan into movement.}
\label{fig:primitive_composition_sketch}
\end{figure}

An \emph{action policy} is a model that takes in observations and acts on them. Robotics has three forms of it (\cref{subsec:planner_commander_robotics}): a VLA, one network that outputs actions; a slow planner that issues subgoals to a VLA; and, most recently, a multimodal large language model that writes the policy as code. Motor cortex serves the same function with a different structure: separate areas plan, choose, release, and verify each action, and practice trains them for life (\cref{subsec:planner_commander_biology}). We propose one model that composes action units from primitives (\cref{fig:primitive_composition_sketch}), turns each unit into motion, and keeps learning after deployment by moving each primitive from code to weights, from explicitly parameterized perception and action to parameters held implicitly in the weights (\cref{subsec:planner_commander_inspiration}).

\subsection{Robotics}
\label{subsec:planner_commander_robotics}

In this subsection we cover the VLA's architecture (\cref{subsubsec:planner_commander_vla_architecture}) and its training stages (\cref{subsubsec:planner_commander_vla_training}), the planner over an executor (\cref{subsubsec:planner_commander_planner_executor}), the multimodal model that writes the policy as code over tools (\cref{subsubsec:planner_commander_code_over_tools}), and the state of continual learning and recursive self-improvement (\cref{subsubsec:planner_commander_continual_learning}).

\subsubsection{A VLA is a backbone plus action head}
\label{subsubsec:planner_commander_vla_architecture}

\begin{figure}[H]
\centering
\includegraphics[width=0.95\textwidth]{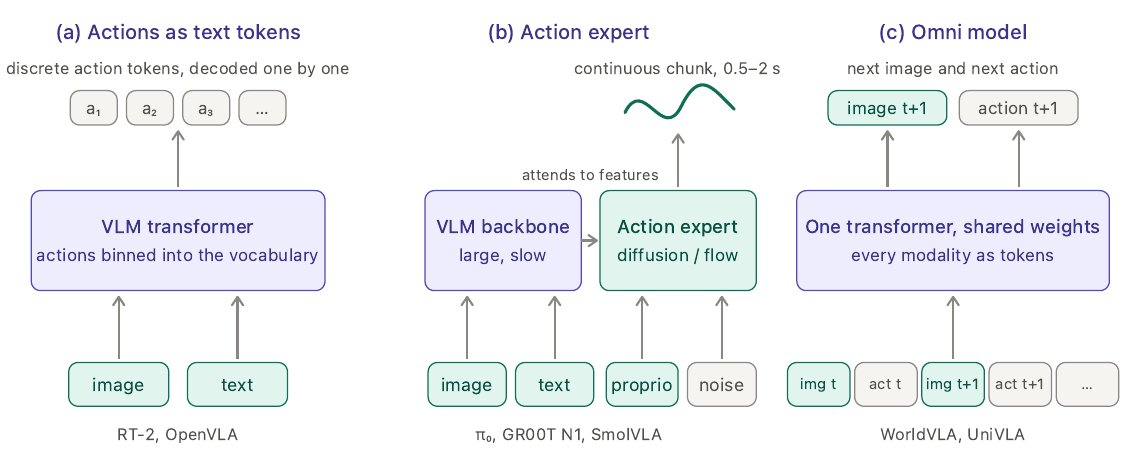}
\caption{Actions attach to a vision-language backbone in one of three ways. (a) The first generation bins each action dimension and decodes actions as text tokens, one at a time. (b) The current generation keeps the backbone and attaches a small action expert that attends to its features and denoises a noisy chunk into continuous actions by diffusion or flow matching; proprioception enters the expert, as drawn, in $\pi_0$ and GR00T N1, and the backbone in SmolVLA \protect\citep{black2024pi0,bjorck2025groot,shukor2025smolvla}. (c) An omni model has no head. Images, text, and actions are tokens in one sequence, so the same network predicts the next action and the next image. Purple: transformer with pretrained weights. Teal: images, text, proprioception, and the action expert. Gray: discrete action tokens and the noise the expert denoises.}
\label{fig:vla_architectures}
\end{figure}

A VLA is a vision-language model with actions attached. The backbone brings visual and language grounding from internet-scale pretraining, and behavior cloning on robot demonstrations adds the motor output \citep{brohan2023rt2,black2024pi0}. In every form the output is a chunk of future targets, typically 0.5--2\,s at 10--50\,Hz (\cref{subsubsec:action_chunking,subsubsec:action_representation}); the forms differ in how the chunk is produced (\cref{fig:vla_architectures}).

The first form treats actions as text. Each action dimension is binned and written onto the vocabulary's least-used tokens, and the backbone decodes the chunk one token at a time \citep{brohan2023rt2,kim2024openvla}. Taken to its limit, this gives an \emph{omni} model in which images, text, and actions share one token sequence and one set of weights \citep{cen2025worldvla,wang2025univla}. The costs follow from the same choice: binned high-rate actions form long, highly correlated token streams that train poorly, decoding one token at a time is slow, errors accumulate along the chunk, and with no separate head the whole backbone runs at the control rate \citep{pertsch2025fast,cen2025worldvla}.

The second form attaches a separate \emph{action expert}, a smaller transformer that attends to the backbone's features and emits the chunk as continuous values by diffusion or flow matching \citep{chi2023diffusion,lipman2023flow}; $\pi_0$, CogACT, GR00T N1, and SmolVLA all take this form \citep{black2024pi0,cogact2024,bjorck2025groot,shukor2025smolvla}. The head is generative mainly because demonstrations are multimodal: the same scene admits several valid trajectories, and a regression head would average them where sampling keeps one \citep{chi2023diffusion,zhao2023learning}.

\subsubsection{Three training stages shape the VLA, and none writes deployment experience back}
\label{subsubsec:planner_commander_vla_training}

\paragraph{Pretraining imitates demonstrations, and demonstrations are scarce.} Pretraining teaches the inherited vision-language backbone to act by behavior cloning on pooled multi-robot data, usually with a diffusion or flow-matching loss \citep{osa2018imitation,ho2020ddpm,lipman2023flow}; the loss rewards only fidelity to the demonstrator. Such data are scarce: the largest pooled datasets hold about a million episodes \citep{oneill2024openx,bu2025agibotworld}. Generalization grows with the variety of scenes, objects, and tasks, while extra demonstrations of the same scene and object stop helping after about fifty \citep{lin2024datascaling,lin2025diversity}. Human video is plentiful but records no robot actions, which must be inferred as latents between frames or from hand pose \citep{ye2024lapa,hoque2025egodex}.

\paragraph{Supervised fine-tuning adapts the policy to one robot and keeps the imitation objective.} Because pooled data mix embodiments, fine-tuning on tens to hundreds of target-robot demonstrations fixes the action space, camera placement, and control rate, through full updates or low-rank adapters \citep{kim2024openvla,black2024pi0,hu2022lora}. Imitation copies every demonstration with equal weight, so a smaller, consistent set often beats a larger one that mixes operators' strategies \citep{belkhale2023dataquality}. Recent systems add the policy's own rollouts, labeled by quality so the model conditions on poor data instead of copying it \citep{pi07_2026,agibot2026lwd}. The objective still rewards fidelity alone, so a policy that drifts off the demonstrated states compounds its errors: the data hold no recoveries \citep{ross2011reduction}.

\paragraph{Reinforcement learning is the only stage at which success enters the objective.} A reward scores the states the policy actually reaches, so its own failures become training signal. The experience can come from logged data reweighted by a learned value \citep{levine2020offline,kostrikov2022iql}, expert interventions at the moment of failure \citep{kelly2019hgdagger,luo2024hilserl}, a learned simulator \citep{li2025vlarft,wmpo2025}, or the real robot \citep{zhang2025reinflow,chen2025conrft,lei2026rl100}. At frontier scale the value verdict enters the context as a token \citep{recap2025}, the critic trains the flow head directly \citep{agibot2026lwd}, or a residual policy corrects a frozen VLA \citep{ankile2025resip,xiao2025pld}. In each case the policy surpasses its demonstrators within hours of real interaction on a single task \citep{luo2024hilserl,chen2025conrft,agibot2026lwd}. Yet this fixes only the failures that were collected; deployment brings new payloads, shifted dynamics, visual corruption, and wear, met with frozen weights.

\subsubsection{A planner decomposes the task into subgoals for a fast executor}
\label{subsubsec:planner_commander_planner_executor}

A policy fast enough for the control loop is small, and a small model plans poorly. A slow planner supplies what it lacks: in the System 1 and System 2 split \citep{kahneman2011thinking,openhelix2025,zheng2026dualsystem}, a 1--7B-parameter VLM at 7--10\,Hz interprets the task and a small transformer at 120--200\,Hz acts on it \citep{figureai2025helix,bjorck2025groot}. What passes between them trades legibility for bandwidth. Text is the most legible: the planner issues atomic commands \citep{shi2025hirobot} or writes the subtask before its own action chunk \citep{pi05_2025}, which generalizes better and can be corrected \citep{belkhale2024rth,zawalski2024ecot}, but carries little pose and no contact or timing. A subgoal image adds the goal layout, where each object and the hand should end up and what should be touching what, though still no force or timing: a generative model imagines the next scene and conditions the executor on it \citep{black2023susie,pi07_2026}. A latent embedding carries the most but is opaque and trained jointly with its executor \citep{lcb2024,figureai2025helix,chen2025fastinslow}.

\subsubsection{A multimodal large language model writes the policy as code}
\label{subsubsec:planner_commander_code_over_tools}

The third design is a general multimodal large language model with reasoning and coding ability, such as GPT-6 Astra \citep{openai2026astra}. It writes a program over a library of perception and motion routines, so the plan is explicit code \citep{liang2023code,huang2023voxposer,mu2024robocodex}; Cartesian control is the default, and a VLA is one option, called when the task is one it has been trained on \citep{zhang2026harnessvla,xu2026baton}. Because its competence comes from reasoning and coding beyond robot data, it generalizes beyond any demonstration set: zero-shot it ranks first among public policies at about 22\% success \citep{zhang2026unexpected,galbot2026astra}; handing familiar manipulation to a VLA roughly doubles that in some cases \citep{galbot2026astra}, and a harness lifts it further \citep{liu2026guava,zhang2026harnessvla,kim2026rhd}. 

The limit is time and reactivity. Each call costs tens of seconds of thought, so by the time the arm moves, the plan describes a scene that has already changed \citep{galbot2026astra}. And because that call is the only loop, nothing closes the gap in between: a slipping grasp waits for the next round of reasoning \citep{galbot2026astra,zhang2026unexpected}.

\subsubsection{Experience enters the model in context, as skills, or in the weights, and no loop yet runs without a human}
\label{subsubsec:planner_commander_continual_learning}

\paragraph{Past experience reaches the model by three routes.} A human can provide a demonstration to the context \citep{fu2024icrt,shah2025mimicdroid,pi07_2026}, or the robot can refer to its own history there \citep{liu2025locoformer,vandermerwe2025incontext}; both help, but the context has a size limit, so what scrolls out is lost (\cref{subsubsec:cerebellum_world_model_adaptation_today}). A coding agent can write what it learns into a playbook \citep{kim2026rhd} or a library of verified routines, where a record of past outcomes lifts real-robot progress several-fold \citep{zhang2026harnessvla,liu2026roboicl,singh2026cfam}. Rollouts and corrections can be post-trained into the weights offline \citep{pi07_2026,recap2025} or streamed from a fleet and redeployed within hours \citep{agibot2026lwd,agibot2026sop}, and a small replay fraction prevents most forgetting \citep{liu2026pretrained,utaustin2026forgetting}.

\begin{figure}[H]
  \centering
  \includegraphics[width=0.85\linewidth]{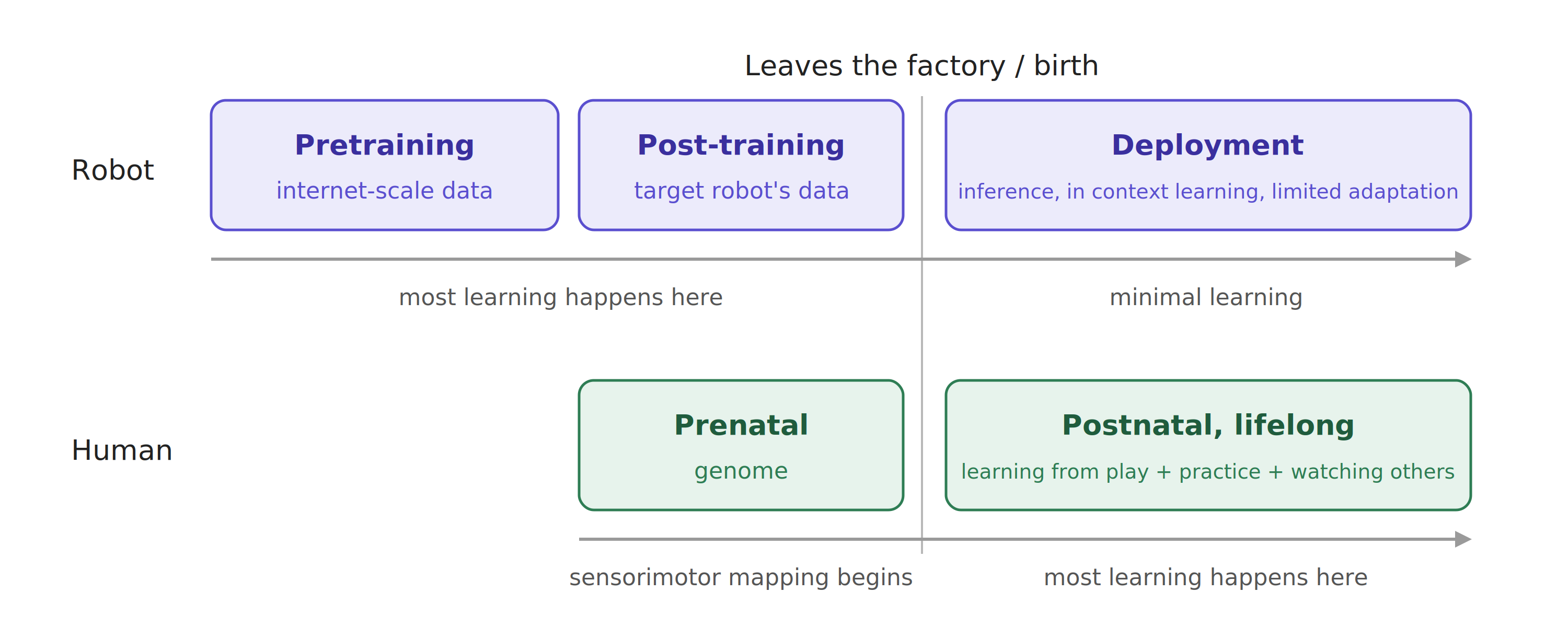}
  \caption{A robot learns mostly before it leaves the factory and afterwards executes the learned policy; a human begins with a genome rather than pretrained weights and does most of its learning after birth, through play, practice, and observation.}
  \label{fig:learning_timeline}
\end{figure}

\paragraph{Automating the loop is the open part.} The three routes are automated unevenly: agents already judge their own outcomes and write skills back, while choosing what to attempt next, deciding when a weight update is warranted, and verifying that an update is a real gain still fall to people. A robot that does all of these itself, as coding agents do for software, exists only in pieces \citep{ahn2024autort,ghasemipour2025selfimproving}. The latest attempts show what works. Attributing each repair to the skill that failed and validating it before reuse keeps improvements adding up across tasks \citep{wen2026roborsi,wang2026whatstops}. Matched controls that separate retained experience from repeated attempts make a success curve credible, and an independent evaluator of each update is the bar for recursive gain \citep{wang2026sear,zhou2026physicalrsi}; \cref{subsubsec:planner_commander_consolidation_agent} proposes the loop.

\subsection{Biology}
\label{subsec:planner_commander_biology}

Current evidence suggests that the cortical motor system issues action commands in four stages: a goal is decomposed into action units whose progress is monitored (\cref{subsubsec:planner_commander_goal_to_units}), each unit is probably composed from primitives that motor cortex already holds (\cref{subsubsec:planner_commander_primitives}), the cerebellar forecast may amend each unit in flight and time the next (\cref{subsubsec:planner_commander_cerebellar_gate}), and practice appears to convert the composition into a habit (\cref{subsubsec:planner_commander_practice_habit}).

\subsubsection{Prefrontal cortex sets the goal, SMA holds the action units, and ACC keeps the place}
\label{subsubsec:planner_commander_goal_to_units}

Prefrontal cortex holds the goal: its neurons carry a rule across a delay, independent of the stimulus and of the movement \citep{miller2001integrative,wallis2001rule}. It orders goals in a hierarchy and holds a sequence with each rank in its own subspace \citep{koechlin2003architecture,xie2022sequence}. The goal enters the motor areas through the supplementary motor area (SMA), which runs a sequence from memory where premotor cortex runs it from the senses \citep{mushiake1991premotor,nachev2008sma}.

\begin{figure}[H]
\centering
\includegraphics[width=0.6\textwidth]{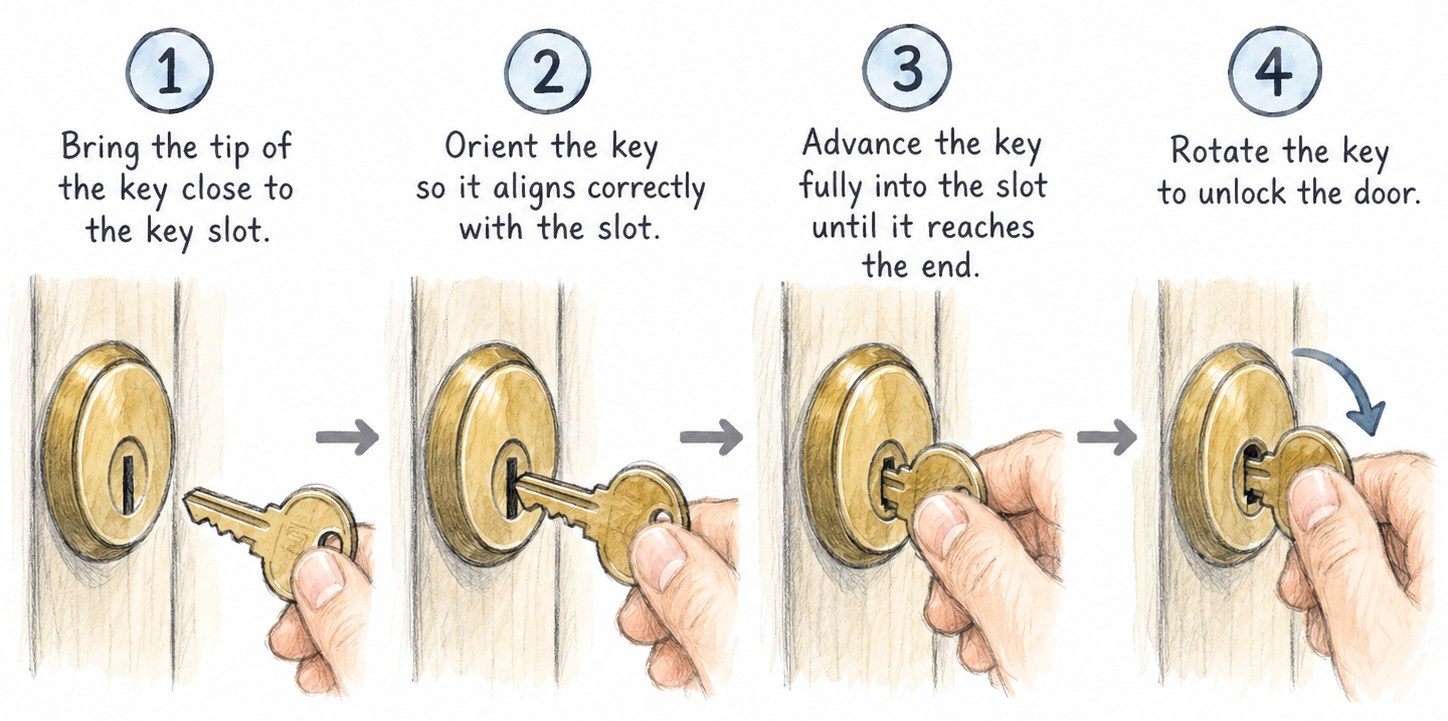}
\caption{Unlocking a door is four action units in a fixed order, reach, orient, insert, and rotate, and each ends at a sensory event: the tip reaches the slot, the key aligns, it stops at the end, the lock turns.}
\label{fig:key_sketch}
\end{figure}

SMA holds the action units and their order: a unit is represented as a whole, and each movement within it by the slot it fills \citep{tanji1994sma,shima2000sma,clower1998sequence}. People form such units without instruction as they learn a long sequence \citep{sakai2003chunking}, and pre-SMA, the front of the area, launches each unit and switches away from one in progress \citep{kennerley2004presma,isoda2007switching}. Anterior cingulate cortex (ACC) keeps the place: its neurons track how many units remain before reward, and a subgoal reached evokes a prediction error like a reward's \citep{shidara2002acc,ribasfernandes2011hrl}. Unlocking a door is such a sequence: reach, orient, insert, and rotate, each unit ending at a sensory event the next waits for (\cref{fig:key_sketch}).

\subsubsection{Motor cortex holds sensor-tuned primitives, and a competition among premotor candidates chooses which to compose}
\label{subsubsec:planner_commander_primitives}

\begin{figure}[H]
\centering
\includegraphics[width=\textwidth]{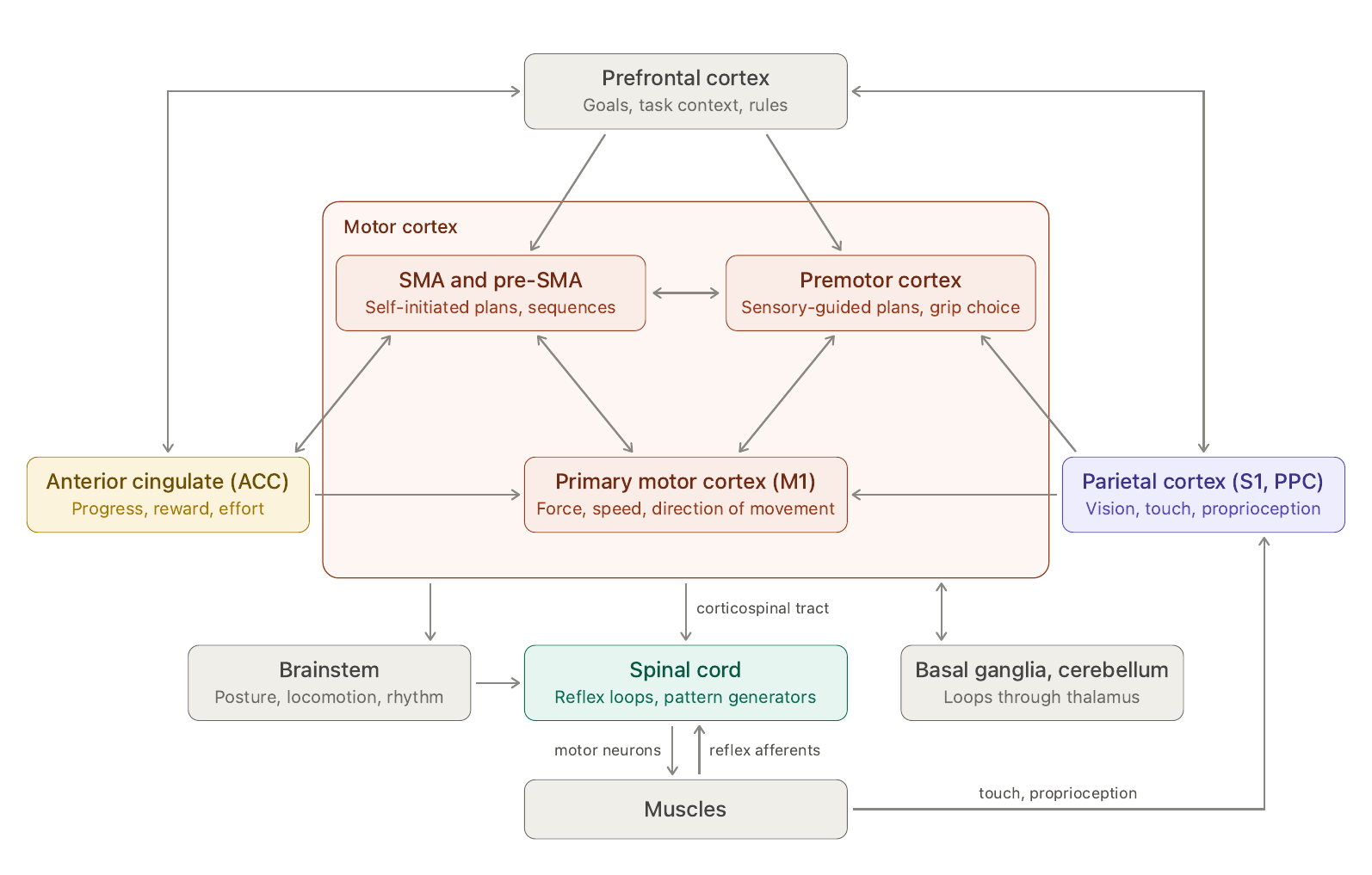}
\caption{Motor cortex is M1, premotor cortex, and SMA, reciprocally connected, with three cortical inputs: prefrontal goals enter through SMA and premotor cortex, the parietal state enters through premotor cortex and, from S1 and area 5, M1 directly, and ACC reaches SMA and M1 through the cingulate motor areas \protect\citep{strick2021motor}. M1 carries most of the corticospinal tract, and SMA, premotor, and cingulate areas each add a direct projection that bypasses it \protect\citep{dum1991corticospinal}. Motor cortex also drives the brainstem pathways and runs loops through the basal ganglia and cerebellum \protect\citep{lemon2008descending}; reflex afferents return to the cord and touch and proprioception return to parietal cortex. Only the principal pathways are drawn.}
\label{fig:motor_hierarchy}
\end{figure}

Motor cortex is primary motor cortex (M1), the last cortical stage before the spinal cord, plus a set of premotor areas that feed it \citep{strick2021motor}; we focus on premotor cortex and SMA (\cref{fig:motor_hierarchy}). Its repertoire is a set of primitives: stimulating premotor cortex or M1 in macaques for half a second evokes a whole action, the limb ending in the same posture wherever it started \citep{graziano2002complex,graziano2016ethological}. Each primitive has a sensory input already tuned for it: reach, orientation, and grasp run on separate parieto-premotor circuits \citep{galletti2003role,fattori2009orientation,murata2000aip}, and ventral premotor neurons fire at the sight of an object sized for the grip they command \citep{murata1997f5}.

Which action to take is decided by competition among candidates. Dorsal premotor cortex can hold several potential reaches at once, each scaled by its likelihood, until one grows and suppresses the rest \citep{cisek2005reaching,genkin2025choice}, and the prefrontal rule tilts the competition toward the one that serves the goal \citep{wallis2001rule,cisek2010choices}. M1 activity correlates with the force, direction, and speed of the movement in progress \citep{georgopoulos1986population}, though population dynamics rather than parameter coding may be the better description \citep{churchland2012}.

\subsubsection{The cerebellum's forecast amends the current unit and times the next}
\label{subsubsec:planner_commander_cerebellar_gate}

The cerebellum forecasts the consequence of each command and returns the forecast by two routes, descending to amend the command in flight and ascending to the cortex that issued it (\cref{subsubsec:cerebellum_world_model_ascending_output_cortex_routes}). For the action unit, the two routes do different work.

Descending, the forecast corrects the unit in progress without ending it. A small error is absorbed implicitly, without awareness \citep{taylor2014explicit,kim2018invariant}; a large one also prompts an explicit change of strategy at the boundary \citep{mcdougle2015explicit}.

\begin{figure}[H]
\centering
\includegraphics[width=0.85\textwidth]{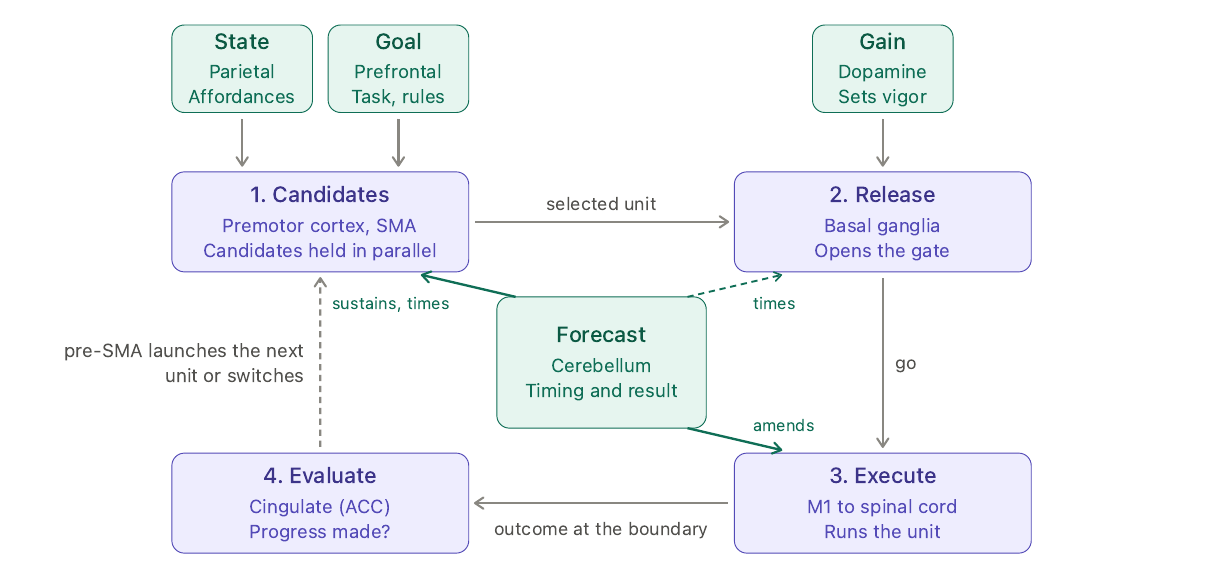}
\caption{One action unit runs as a loop; the arrows show the flow of control rather than anatomical pathways. Purple marks the four stages, run clockwise, and teal what each stage reads. Solid teal arrows have direct support: the cerebellar forecast amends the unit in flight \protect\citep{becker2019reach,low2018interposed} and sustains and times the cortical preparation of the next \protect\citep{gao2018cortico,chabrol2019preparatory}. Dashed arrows are hypotheses: whether the forecast also times the basal ganglia release is untested, and pre-SMA, which launches units and switches between them \protect\citep{kennerley2004presma,isoda2007switching}, is proposed here to do so on the ACC verdict. Assigning one area to each stage is a simplification; each area takes part in more than one.}
\label{fig:action_unit_lifecycle}
\end{figure}

Ascending, the forecast tells cortex where the current unit will end, and its gap from the goal is what cortex plans on: if the unit will fall short, cortex adjusts the next unit or replans the strategy before the shortfall is sensed, which is the forecast's main service to the planner. The next unit forms in an output-null subspace of premotor cortex and M1 where it cannot reach the muscles \citep{kaufman2014nullspace,churchland2024nullspace}, and cerebellar input sustains it and times its release (\cref{subsubsec:cerebellum_world_model_ascending_output_cortex_routes}). Release runs through a midbrain--thalamus--cortex trigger that the basal ganglia otherwise hold shut \citep{inagaki2022trigger,mink1996basal}; whether the cerebellar signal also reaches that gate is open (\cref{fig:action_unit_lifecycle}).

\subsubsection{Practice turns the composition into a habit, selected by dopamine and consolidated offline}
\label{subsubsec:planner_commander_practice_habit}

A new skill begins as a composition of primitives the repertoire already holds, acquired within a session where a pattern outside the repertoire needs days \citep{hikosaka1999parallel,sadtler2014constraints,oby2019newpatterns}. Practice turns the composition into a habit that runs largely below cortex. Striatal bursts come to bracket the whole sequence at its start and end \citep{jog1999building,graybiel1998chunking}, and the basal ganglia then set its detailed kinematics \citep{dhawale2021kinematics}. In rats, M1 is required to learn a lever-press skill but not to execute it once learned, unless the sequence must be reordered on the fly \citep{kawai2015motor,mizes2024flexibility}. In one account, control shifts from cerebellar to striatal circuits along the way \citep{doyon2003distinct}, and the co-contraction that stiffened the limb while its model was wrong is released \citep{thoroughman1999emg,shmuelof2012skill}.

Dopamine decides which synapses practice strengthens. Co-activity marks a corticostriatal synapse as a candidate, an eligibility trace \citep{gerstner2018eligibility}, and a dopamine burst or dip, the reward prediction error, supplies the verdict \citep{schultz1997neural,fremaux2016neuromodulated}. A spine enlarges only if dopamine arrives within about two seconds of its activation \citep{yagishita2014dopamine}, and corticostriatal synapses strengthen over the days of training \citep{yin2009dynamic,phillips2024striatal}. Practice concentrates where motor variability rises \citep{dhawale2019adaptive}.

Offline replay consolidates what dopamine chose. New spines form in M1 within hours of practice, and retention tracks how many survive \citep{xu2009rapid,yang2009stably}. A finger sequence runs about a fifth faster the next morning without further practice, and less so after an equal interval awake \citep{walker2002practice,kuriyama2004complexity}, though how much of this gain is specific to sleep is disputed \citep{pan2015sleep,gupta2022reactive}. During slow-wave sleep the ensembles active in the day's task reactivate, and the reactivation predicts the next day's gain and selects the synapses kept \citep{ramanathan2015sleep,gulati2017neural,yang2014sleep}.

\subsection{Inspiration}
\label{subsec:planner_commander_inspiration}

The proposals of this chapter combine into one agent. For a new task, a single multimodal model composes a solution from existing primitives, verifying each at its boundary (\cref{subsubsec:planner_commander_planner_units}), and runs each by Cartesian control or a reactive policy head (\cref{subsubsec:planner_commander_commander_action}); each primitive is a template the model parameterizes per instance (\cref{subsubsec:planner_commander_parameterized_primitives}). The solution then matures through two agents: a practice agent rehearses its weakest unit (\cref{subsubsec:planner_commander_practice_agent}), guided by judging which direction moves an attempt closer to success (\cref{subsubsec:planner_commander_directional_judgment}), and a consolidation agent prunes the strategy and graduates each primitive from explicit code into typed functions the model answers, then into the weights (\cref{subsubsec:planner_commander_consolidation_agent}). Beside the model, a predictor forecasts the consequence of each action (\cref{subsubsec:cerebellum_world_model_predictor_beside_commander,subsubsec:cerebellum_world_model_predictor_in_loop}).

\subsubsection{Let the model compose action units from reusable primitives and verify each at its boundary}
\label{subsubsec:planner_commander_planner_units}

Splitting a task into steps is largely solved: GPT-6 Astra already breaks opening an unfamiliar lock into orient, insert, and rotate \citep{openai2026astra,he2026astrabody}. What is missing is a library of reusable primitives, pick, place, orient, slide, insert, stored as skills or weights and shared across instances \citep{liang2023code,si2026pyrua,fu2026capx}. A new skill is then a code plan over them, as opening a lock is a plan over reach, orient, insert, and rotate (\cref{fig:key_sketch}), and where no primitive fits, the model writes a new one as code, since large models generalize to programs they have not seen \citep{liang2023code,si2026pyrua}. Each primitive holds two things. The first is a verifiable goal: a test the robot runs on its own sensors at the unit's boundary, and the next unit starts only once that test has fired \citep{wan2026nofreechecker,xu2026baton,chen2026astrarobomme}. The second is an action plan, code or a policy head that pursues the goal (\cref{subsubsec:planner_commander_commander_action}), written as a template whose parameters the model fills in for each instance (\cref{subsubsec:planner_commander_parameterized_primitives}). When a test fails, the model must reason to the cause, a slot too narrow, a second object dragging the part back, and revise the plan or adjust the force accordingly. Current multimodal models rarely manage this diagnosis, least of all where forces interact out of the camera's view \citep{navasardyan2026failbench,gu2025safe,li2026forcevla2}.

\subsubsection{Treat each primitive as a parameterized template}
\label{subsubsec:planner_commander_parameterized_primitives}

\begin{figure}[H]
\centering
\includegraphics[width=\textwidth]{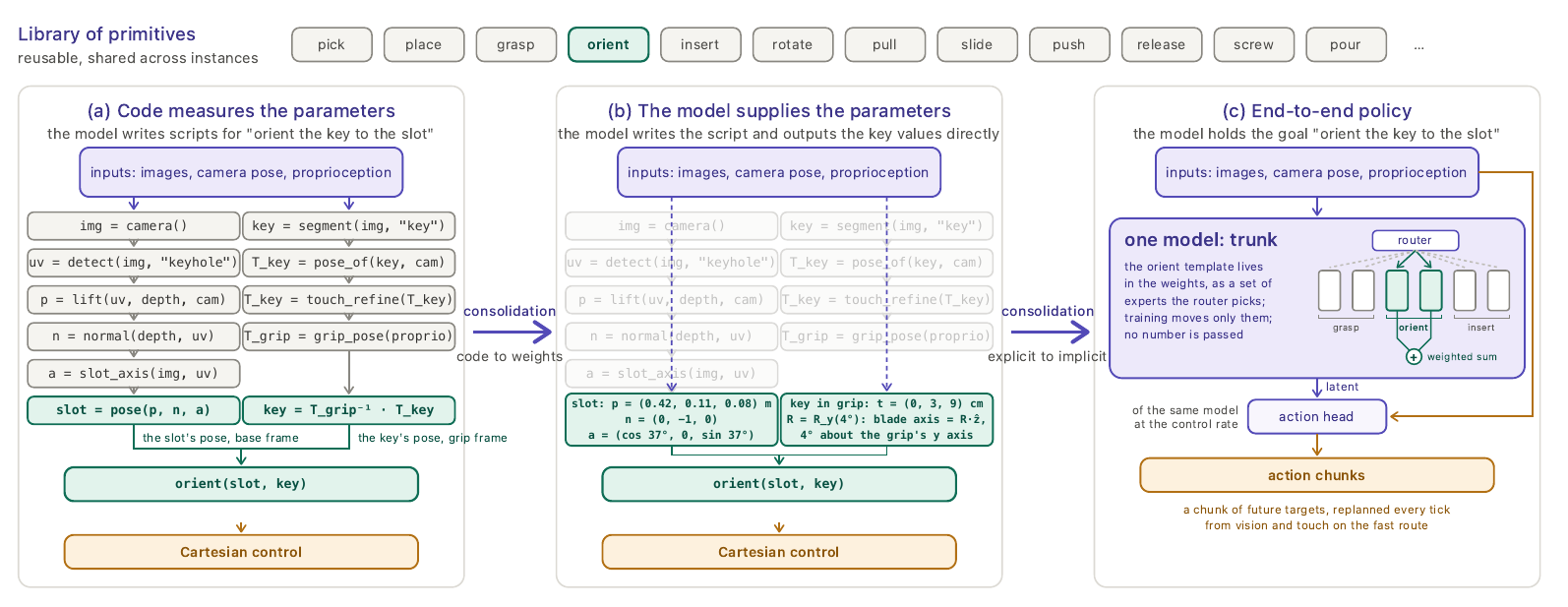}
\caption{Consolidation advances a primitive from a script over many functions to an end-to-end policy. (a) At first the model writes a perception pipeline that computes the slot's pose and the key's geometry in the hand before calling the orient template, which runs by Cartesian control. (b) Next the model outputs both values directly from the images and supplies them to the same template, so the explicit measuring code is no longer needed. (c) In the end the template lives in the orient experts of the model's mixture-of-experts trunk: the action head of the same model reads their latent, no number is passed, and the action stream reacts at the control rate.}
\label{fig:primitive_consolidation_stages}
\end{figure}

The plan of a primitive is generic; what changes from one instance to the next is its parameters: the grasp point and approach direction, the slot's angle, the insertion depth, the force at which to stop. A correct plan with a wrong angle fails as surely as a wrong plan, so supplying precise parameters is where multimodal capability matters most (\cref{subsubsec:vision_direct_readout}). Where the parameters come from is what changes as a primitive matures (\cref{fig:primitive_consolidation_stages}). At first the model writes code that measures them: segment the slot, fit a line, read off the angle. As the model sees more instances, it reads parameters off the image directly, one at a time, and the code shrinks to the measurements it cannot yet make reliably; the handover is gradual, with the code's answer serving as the check on the model's until the two agree often enough that the code is dropped. In the end the parameters are never named at all: the weights of the action head read them implicitly from the scene. The consolidation agent of \cref{subsubsec:planner_commander_consolidation_agent} advances each primitive along this path.

\subsubsection{Let the model run each unit by Cartesian control or by a reactive policy head}
\label{subsubsec:planner_commander_commander_action}

A primitive's action plan takes one of two forms. The direct form is code: the model writes Cartesian pose targets, interpolates them at the control rate, and leaves contact to the compliance layer of \cref{subsec:action_inspiration}. GPT-6 Astra does this today and succeeds where a task consists of reaches and placements \citep{galbot2026astra,he2026astrabody,si2026pyrua}. Between one boundary test and the next, though, the trajectory is blind: a cable that snags is noticed only when the unit ends, and code holds no prior about how objects behave.

The proposed form is a lightweight policy head that runs the plan against real-time feedback at the control rate, its weights holding that prior \citep{black2024pi0,bjorck2025groot,chen2025fastinslow}. The two forms compose: code sets a unit's milestone poses while the head generates the trajectory between them and reacts to contact \citep{zhang2026harnessvla,xu2026baton}. The head reads its instruction as a latent from the same model, already carrying pose, contact, and timing, so it needs no second language backbone \citep{lcb2024,figureai2025helix}. Reacting this fast needs an equally fast vision encoder \citep{reflexvla2026,unifs2026}, the streaming encoder of \cref{subsubsec:vision_streaming_control_encoder}.

\subsubsection{Add a consolidation agent that graduates proven primitives from code into weights}
\label{subsubsec:planner_commander_consolidation_agent}

Between sessions, an offline agent reviews the records of the boundary tests, answered by the typed experts of \cref{subsubsec:vision_direct_readout} that we call the judge, and keeps what paid off, as consolidation does in \cref{subsubsec:planner_commander_practice_habit}. It writes primitives, task sequences, grasp records, and failure explanations, and moves each proven primitive along the path of \cref{fig:primitive_consolidation_stages}: a cleaned script, then typed functions the model answers from the image, and in the end a policy with no code in the loop.

\paragraph{Consolidation first cleans the script, keeping the shortest sequence that is grounded and explained.}
\label{subsubsec:planner_commander_consolidation_clean}
The run-time code plan (\cref{subsubsec:planner_commander_planner_units}) carries whatever happened to work alongside what made it work. The agent replays it against the stored variations of the scene and keeps the shortest sequence that still passes; a retry loop survives only where the replay fails without it \citep{xu2026baton,luo2024hilserl}. Every kept unit states its precondition and boundary test, and every value traces to a measurement or a stored record of the object: the trial-found offset of \cref{fig:practice_agent_loop} stays only once it is explained by the measured slot and key. The script is then stored, keyed by goal and scene, for retrieval into context \citep{wang2023voyager,liu2026roboicl}.

\paragraph{Each measurement then graduates from code into a typed function the model answers directly.}
\label{subsubsec:planner_commander_consolidation_perception}
The stored script still finds the slot's angle with a program that thresholds the image and fits a line \citep{kumar2026aor,chen2026gap}. Each passed run pairs a scene with the number that worked, and training on these pairs lets the model answer the same call from the image, keeping the function's signature and dropping its body \citep{yuan2024robopoint,ha2023scaling,kim2026rhd}. Code and model answer the same inputs until they agree often enough that the code serves only as the test.

\paragraph{In the end the policy head runs the primitive end to end, and only the test stays explicit.}
\label{subsubsec:planner_commander_consolidation_end_to_end}
A practiced primitive runs with no code in the loop: the head receives ``screw the cap onto the bottle'' and reacts to streaming visual and tactile input from start to finish, as a VLA does today \citep{black2024pi0,reflexvla2026,chen2025fastinslow}. The angle once measured is carried implicitly in a shared latent, and a part that binds is answered without passing through a scalar \citep{lcb2024,yu2025forcevla,li2026forcevla2}. The boundary test does not graduate: as a person notices a cap cross-threading, the judge keeps asking its typed question at each boundary, because that verdict is what the loop trains on.

\subsubsection{Design the next architecture so that primitives are experts routed by the plan and share embeddings}
\label{subsubsec:planner_commander_end_to_end}

The current model is slow, and coding is slow \citep{liang2023code,galbot2026astra,zheng2026dualsystem}. The round trip of writing out text and reading it back in between the model's reasoning passes is wasteful. If everything passed in between were embeddings, turned into human language only when the robot needs our feedback, it would be much faster \citep{lcb2024,figureai2025helix}. Many animals do not speak a language and accomplish hard action tasks \citep{dhawale2017variability,umilta2008pliers}.

The logic that today lives in generated code, which unit runs next and what it receives, can instead be carried by routing between experts. An expert here is a small part of the large model that takes in an embedding and returns an embedding; the notion borrows from the mixture of experts, where today's mixtures route tokens between experts for load balance \citep{shazeer2017,fedus2022switch}. The difference is what an expert stands for. Here each primitive persists as a group of experts that behaves as a function call whose input and output are embeddings rather than numbers, and all groups read one shared encoding of the scene (\cref{subsubsec:vision_dorsal_stream_channels}). Orient activates the experts that align a part, insert those that regulate contact force, and a hand-off between primitives is a change of routing in place of a line of code \citep{mai2026meat,yu2025forcevla,hex2026}. More experts at fixed active parameters add room for new primitives at no cost per step.

The benefit is not only speed. Concepts no longer need to be re-encoded and decoded at every hand-off, which wastes much of today's compute and time \citep{zheng2026dualsystem,lcb2024}. An expert can also be trained from a small number of examples, since on well-trained input and output embeddings it is computationally on the level of a small multilayer perceptron \citep{yu2024moeadapters,modularmemory2026}. The routing itself needs learning: each expert would carry an address that is encoded, and the model learns the mapping from the plan to that address \citep{shazeer2017,mai2026meat}. The challenge is inspection. Once the output is no longer human language, it is hard for a person to inspect what training did, and a readout that decodes a latent to a number on demand stays for the judge. This is only our proposal, drawn with wide imagination, and we look forward to creative ways of developing the next generation of models for robotics.

\subsubsection{Add a practice agent that rehearses weak skills and tests what consolidation wrote}
\label{subsubsec:planner_commander_practice_agent}

\begin{figure}[H]
\centering
\includegraphics[width=\textwidth]{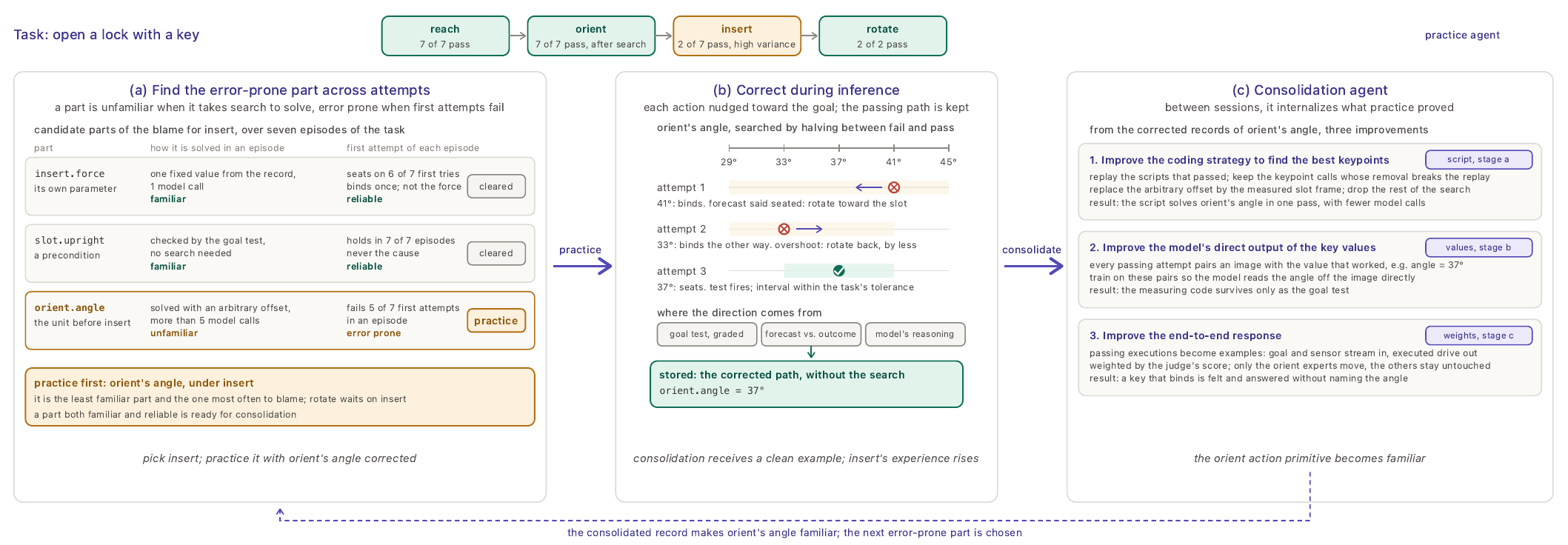}
\caption{The practice agent finds the error-prone part across attempts, corrects it during inference, and hands the corrected record to consolidation. (a) Over seven episodes of opening a lock, insert passes rarely, and the blame falls on orient's angle, which is solved with an arbitrary offset after several model calls and fails most first attempts; it is unfamiliar and error prone and is practiced first. (b) During practice the direction to move after each attempt comes from the graded goal test, the forecast against the outcome, or the model's reasoning; the interval halves until the goal test fires, and the angle that passed is stored. (c) Consolidation then improves the coding strategy that finds the keypoints, trains the model to output the values directly, and trains the orient experts of the trunk, so the orient primitive becomes familiar.}
\label{fig:practice_agent_loop}
\end{figure}

We propose an autonomous practice agent that picks the most valuable tasks, rehearses their weakest unit, and tests each record consolidation wrote by the trial that would refute it (\cref{fig:practice_agent_loop}).

\paragraph{It picks the most valuable tasks and practices their weakest unit.} A task is valuable when its reward is high and the gap between the robot's current success and the success it could reach is large: a task already solved earns nothing more, and one nobody rewards is not worth the practice. Within the chosen task, the model reasons over the records to decide which unit is to blame, since a failure that surfaces at one unit's boundary may have been caused by the unit before it; the blamed unit is the weakest, and practice goes there first. Three signs mark it: it fails most often, it demands the most deliberation before it passes, extra model calls and search for a parameter that should have been read off directly, and its parameters vary most from one attempt to the next. All three are signs of inexperience: an experienced unit chooses the same primitive and parameters across attempts in similar states and reaches the goal without extra search or reasoning, while an inexperienced one has few attempts behind it and is still being composed \citep{wu2014temporal,dhawale2017variability}, and inconsistency between successive action predictions within a rollout predicts failure at run time \citep{agia2024sentinel,seligmann2026vlafail}.

\paragraph{It treats each record as a hypothesis and runs the trial that would refute it.} What consolidation wrote is a set of claims that can be wrong: that a failure was due to the pose, that pouring does not depend on the cup's height. The practice agent schedules the one trial that tests the claim with everything else held, and a claim that fails is rewritten with the new evidence. Practice stops when the outcome variance inside it has fallen to the task's tolerance.

\subsubsection{Mastering a new skill rests on judging which direction moves it closer to success}
\label{subsubsec:planner_commander_directional_judgment}

A precise pose or force is rarely found on the first attempt, and it need not be: recursive self-improvement requires only that a direction of improvement be readable from a failed attempt, since whatever can be corrected can be improved without a human in the loop \citep{ghasemipour2025selfimproving,nguyen2026selfevolving,luo2024hilserl}.

Correction rests on one judgment: which of two outcomes lies closer to the goal. Where the gap is near linear in the adjustment, as a distance is in a pose, the model reads the correction off the gap, the plug stopped short so advance \citep{belkhale2024rth,zawalski2024ecot}, and climbs that gradient until the unit's test fires \citep{luo2024hilserl,tsay2024fundamental}. Where it is not, as in the body posture for a badminton stroke, no single joint tells which direction leads to a better outcome, so the agent samples more widely to map the landscape before choosing a direction \citep{dhawale2017variability,dhawale2019adaptive}.

Prediction is essential to the search, because it supplies the link between a change in the action and a change in the outcome. Forecasting the consequence of a candidate adjustment makes the gap between forecast and goal a near-linear function of the correction, so the direction to move is read off the forecast (\cref{subsubsec:cerebellum_world_model_predictor_beside_commander}) \citep{pilworld2026,hafner2025dreamer}. The final corrected trajectory is handed to the consolidation agent (\cref{subsubsec:planner_commander_consolidation_agent}).

\clearpage
\section{Conclusion}
\label{sec:closing}

How the next robot model should be built is an open debate: a single vision-language-action policy (VLA) or world action model (WAM) \citep{black2024pi0,pi07_2026}, code as policy \citep{openai2026astra,zhang2026harnessvla}, or a slow system 2 paired with a fast system 1 \citep{figureai2025helix,zheng2026dualsystem}. Whichever wins, humans remain far ahead of robots: we solve novel tasks, handle objects we have never held, react fast to a moving world, and keep improving for life, while policies trained on a million pooled trajectories do none of these reliably (\cref{sec:introduction}). The nervous system is a working design that already does all of this, so this survey asked what robotics should borrow from it.

The two architectures differ in where the work is done. The robot stack is a model plus servo control. The model is a VLA or WAM, often with a planner or coding agent above it, that reasons or predicts at hundreds of milliseconds to seconds per step; it takes a language instruction and camera frames and emits target positions (\cref{subsubsec:planner_commander_planner_executor,subsubsec:planner_commander_vla_architecture,subsubsec:planner_commander_code_over_tools}). Servo control tracks those positions with fixed gains. The nervous system has the same two levels, but each is richer. The upper level is a set of cooperating modules: cortex, thalamus, basal ganglia, and cerebellum, which divide goal decomposition, motion planning, gating, reward, and prediction among them. The lower level receives motor neuron firing rather than positions; muscles turn it into force, and opposing muscle pairs set stiffness as well as torque, so the body is compliant by construction. \Cref{tab:closing_summary} sets the two side by side, stage by stage, with what we propose.

\begin{table}[H]
\centering
\footnotesize
\begin{tabular}{@{}P{1.7cm}P{3.7cm}P{3.7cm}P{4.9cm}@{}}
\toprule
\textbf{Stage} & \textbf{Current robot stack} & \textbf{Nervous system} & \textbf{Proposal} \\
\midrule
Action (\cref{sec:action}) & A position target goes into fixed servo gains, and the same gains serve every contact. & The command is effort per muscle, and opposing pairs set torque and stiffness separately. & Emit position and stiffness together. Realize the stiffness beneath the policy by impedance or admittance, and let touch set where to yield and how hard to press. \\
Touch (\cref{sec:touch}) & Sensors cover only parts of the hand, and the signal differs per sensor family. & Four afferent classes cover the whole skin at high density and precision, measure both normal and shear force, and a grip reflex corrects slips. & Treat touch as a first-class input, and use it for compliance control. \\
Vision (\cref{sec:vision}) & One recognition encoder reads one stale frame with no velocity, and encoding is slow. & A ventral stream handles recognition, and a separate dorsal stream carries fast metric signals for guiding the hand. & Build a dedicated streaming vision encoder for control. \\
Predictor (\cref{sec:cerebellum_world_model}) & A video world model or WAM takes hundreds of milliseconds per step and adapts poorly at test time. & The cerebellum forecasts within tens of milliseconds, recalibrated with a few trials by its own error. & Train a self-labeling predictor beside the policy. Consult it as a tool.\\
Policy and learning (\cref{sec:planner_commander}) & A VLA copies demonstrations, a slow planner sits over a fast executor, code as policy is slow and blind between calls, and weights are frozen at deployment. & The goal is split into action units checked at their boundaries. Primitives live in motor cortex and combine to solve new tasks, and practice is scored by dopamine and consolidated offline into new motor skills. & One model composes parameterized primitives in code, verifies each at its boundary, and runs them through a reactive head. A practice agent rehearses the weakest unit, and a consolidation agent graduates each skill from code into weights. \\
\bottomrule
\end{tabular}
\caption{Discrepancies between the current robot stack and the nervous system, stage by stage, and the proposal each chapter derives from the gap.}
\label{tab:closing_summary}
\end{table}

Inspired by the neuroscience, here is our proposal for the next generation of robot systems.

\paragraph{One model composes the new task from verified primitives.} The model draws on a library of reusable primitives, reach, orient, insert, rotate, each a template with a verifiable goal: a test the robot runs on its own sensors at the unit's boundary. It breaks the task into action units and parameterizes each for this instance (\cref{subsubsec:planner_commander_planner_units,subsubsec:planner_commander_parameterized_primitives}). The parameters are what change between instances, the angle of this slot, the depth at which this key seats, and the model reads them off the encoded frame when it can and writes measuring code when it cannot. When a test passes, with touch or vision matching the expectation, the model moves to the next primitive. When a test fails, it reasons to the cause, the key is upside down, the lock is stiff, this is the wrong key, and revises the plan or adjusts the force. Where no primitive fits, the model writes a new one as code, since large models generalize to programs they have not seen.

\paragraph{Compliance and touch let the world supply what the policy would otherwise compute.} The insertion need not know the slot's exact pose. The policy emits a position and a stiffness, high along the insertion axis and low across it, and the stiffness is realized beneath the policy, by impedance where the joints deliver torque and by admittance on position-controlled arms (\cref{subsubsec:action_stiffness_matters,subsubsec:action_force_in_contact}). Touch supplies the rest, entering the model as a force field, a normal and shear force per skin point, which any sensor family can be calibrated to (\cref{subsubsec:touch_enabled_compliance_control}). On the key, its magnitude says when to stop closing, and a sudden change signals slip long before vision would.

\paragraph{A reactive policy head and a streaming encoder make the motion fast and continuous.} Code is blind between tests, and a plan executed pose by pose stops at every milestone. So each unit runs through a lightweight policy head inside the same model: code sets the milestone poses and passes down a latent, and the head turns that latent into actions at the control rate, adapting on the fly to small changes (\cref{subsubsec:planner_commander_commander_action}). Vision must keep up. The current encoder is built for recognition, so we propose a second one beside it that runs at camera rate and feeds the head a few short-lived tokens per frame carrying what a reaction needs: optical flow, motion direction, object speed (\cref{subsubsec:vision_streaming_control_encoder}).

\paragraph{A predictor beside the model forecasts the consequence of each action and trains itself.} The predictor receives a copy of every command and forecasts what the sensors will report, so a correction can start before the report arrives. Training is free: each executed action labels itself a moment later, and because the robot caused the outcome, the link it learns is causal (\cref{subsubsec:cerebellum_world_model_predictor_beside_commander}). How the policy uses it depends on speed. While the predictor is slow, the policy consults it as a tool and scores a candidate correction before moving. Once it answers before the sensory report, the policy steers by the forecast and the sensing delay drops out of every correction (\cref{subsubsec:cerebellum_world_model_predictor_in_loop}). A shallow, wide, and sparse architecture makes it fast and lets it recalibrate within the episode: a single wide layer projects the input into a very high-dimensional code and processes it in parallel groups of units, each delegated to one payload regime or body part, so an error-scaled step at test time corrects only the forecast that was wrong (\cref{subsubsec:cerebellum_world_model_sparsity}). The result is a robot that recalibrates to a stiffer lock or a heavier key within a single attempt, without touching the policy that decided to turn it.

\paragraph{Perception moves from explicit code into typed functions, then into the weights.} This is the proposal we rate highest. Large language models once handed arithmetic and programming to a calculator or an interpreter; today much of that computation runs inside the model. We propose the same path for perception. The slot's pose, the key's orientation, and how tightly the fingers should close start as quantities the model computes explicitly in code; with repetition they become typed functions the model calls; with enough practice they become weights that produce the action from the image directly. Facing an unfamiliar lock, the robot measures: it estimates the pose from the image, confirms it with the hand, and checks each step before the next. After consolidation it eyeballs: it glances at the lock and acts, as a person does with a familiar door. Eyeballing works because compliance and fast online correction absorb an estimate that is off by a few millimeters or degrees, so moderate precision is enough to finish the task. \Cref{sec:planner_commander} gives the details.

\paragraph{The robot should practice and sense on its own initiative.} We propose a practice agent that picks the most valuable task and, within it, the unit that fails most often, demands the most deliberation, and varies most between attempts, then rehearses it until its outcome variance falls within tolerance (\cref{subsubsec:planner_commander_practice_agent}). Rehearsal pays off only if the model can judge how far each outcome falls short of the goal, so we propose to strengthen that judgment: a model that rates each state can correct toward the goal during inference, and as long as each correction narrows the gap, the attempts converge (\cref{subsubsec:planner_commander_directional_judgment}). The same initiative extends to what the robot believes. Each record consolidation wrote is a hypothesis, and the agent schedules the trial that would refute it; each unseen property the plan depends on is a hypothesis too, tested before the robot commits: a small rotation of the key tells whether it is seated, a press gauges the lock's stiffness, a change of viewpoint exposes the slot (\cref{subsubsec:touch_active_probing,subsubsec:vision_active_vision}). A robot that takes the initiative and learns actively is the core of recursive self-improvement.

\paragraph{Limitations and outlook.} This is a position paper. The neuroscience it draws on is in part hypothesis, still debated in its own field, and the analogy has limits of its own: robots copy weights, pool records, and run many bodies at once, so some of the pressures that shaped the nervous system may not bear on them. Which of the proposals survives contact with real robots is an empirical question, and we offer each as a hypothesis to be tested. The experiments that decide between these designs are within reach, and the agent they point to is worth building. 

\bibliographystyle{plainnat}
\bibliography{references}

\end{document}